\documentclass[preprint,12pt]{elsarticle}

\usepackage{amssymb}
\usepackage{amsmath}
\usepackage{longtable}
\usepackage[acronym]{glossaries}
\usepackage[nolist,nohyperlinks]{acronym}
\usepackage{caption}
\usepackage{subcaption}
\usepackage{algpseudocode}
\usepackage{algorithm}
\usepackage[dvipsnames]{xcolor}
\usepackage{hyperref}
\usepackage{pdfpages}

\newcommand{\smalltext}[1]{{\tiny #1}}

\newacronym{wmh}{WMH}{white matter hyperintensities}
\newacronym{pvwmh}{PVWMH}{periventricular white matter hyperintensities}
\newacronym{dwmh}{DWMH}{deep white matter hyperintensities}
\newacronym{isl}{ISL}{ischaemic stroke lesions}
\newacronym{mri}{MRI}{magnetic resonance imaging}
\newacronym{ct}{CT}{computed tomography}
\newacronym{flair}{FLAIR}{fluid-attenuated inversion recovery}
\newacronym{dwi}{DWI}{diffusion-weighted imaging}
\newacronym{svd}{SVD}{small vessel disease}
\newacronym{ms}{MS}{multiple sclerosis}
\newacronym{fls}{$FLS$}{fully labelled subset}
\newacronym{pls}{$PLS$}{partially labelled subset}
\newacronym{ap}{AP}{average precision}
\newacronym{dsc}{DSC}{Dice similarity coefficient}
\newacronym{ddsc}{DSC$_\theta$}{distance-Dice similarity coefficient}
\newacronym{dpre}{PRE$_\theta$}{distance-precision}
\newacronym{drec}{REC$_\theta$}{distance-recall}
\newacronym{asd}{ASD}{average surface distance}
\newacronym{avd}{AVD}{absolute volume difference}
\newacronym{pre}{PRE}{precision}
\newacronym{rec}{REC}{recall}
\newacronym{spre}{S-PRE}{subject-level precision}
\newacronym{srec}{S-REC}{subject-level recall}
\newacronym{ldsc}{LDSC}{lesion-level Dice similarity coefficient}
\newacronym{lpre}{LPRE}{lesion-level precision}
\newacronym{lrec}{LREC}{lesion-level recall}
\newacronym{lcwa}{LCWA}{lesion count by weighted assignment}

\journal{Artificial Intelligence in Medicine}

\begin{document}

\begin{frontmatter}


\title{Comparative evaluation of training strategies using partially labelled datasets for segmentation of white matter hyperintensities and stroke lesions in FLAIR MRI}

\author[inst1,inst2]{Jesse Phitidis\corref{cor1}}
\ead{j.phitidis@ed.ac.uk}
\cortext[cor1]{Corresponding author}
\author[inst2,inst3]{Alison Q Smithard}
\author[inst1]{William N Whiteley}
\author[inst1,inst6]{Joanna M. Wardlaw}
\author[inst7]{Miguel O. Bernabeu}
\author[inst1,inst6]{Maria Vald\'{e}s Hern\'{a}ndez}

\affiliation[inst1]{organization={Centre for Clinical Brain Sciences, University of Edinburgh},%Department and Organization
            addressline={49 Little France Crescent}, 
            city={Edinburgh},
            postcode={EH164SB}, 
            country={United Kingdom}}

\affiliation[inst2]{organization={Canon Medical Research Europe},%Department and Organization
            addressline={Bonnington Bond, 2 Anderson Place}, 
            city={Edinburgh},
            postcode={EH65NP}, 
            country={United Kingdom}}

\affiliation[inst3]{organization={School of Engineering, University of Edinburgh},%Department and Organization
            addressline={Sanderson Building}, 
            city={Edinburgh},
            postcode={EH93FB}, 
            country={United Kingdom}}

\affiliation[inst6]{organization={UK Dementia Research Institute, Centre at The University of Edinburgh},%Department and Organization
            addressline={49 Little France Crescent}, 
            city={Edinburgh},
            postcode={EH164SB}, 
            country={United Kingdom}}

\affiliation[inst7]{organization={Usher Institute, University of Edinburgh},%Department and Organization
            addressline={NINE, 9 Little France Road}, 
            city={Edinburgh},
            postcode={EH164UX}, 
            country={United Kingdom}}

%% Abstract
\begin{abstract}
% background
White matter hyperintensities (WMH) and ischaemic stroke lesions (ISL) are key imaging biomarkers of cerebral small vessel disease (SVD) detectable on magnetic resonance imaging (MRI). The development of robust deep learning models to automatically segment and differentiate these pathologies remains challenging. Specifically, WMH and ISL frequently co-occur within the same subject and present as visually confounding hyperintensities on fluid-attenuated inversion recovery (FLAIR) sequences, complicating their accurate delineation.

% what we did
To address the scarcity of fully annotated cohorts, we systematically evaluated six accessible strategies for training a joint WMH and ISL segmentation model using partially labelled data. We aggregated privately held and publicly available datasets to curate a large-scale cohort of 2,052 MRI volumes, of which 1341 and 1152 volumes contained ground truth annotations for WMH and ISL, respectively.

% the results
Our analysis indicates that multiple strategies effectively leverage partially labelled data to enhance overall model performance, with pseudolabelling emerging as the most effective approach. This model exhibited a consistent WMH segmentation policy and successfully detected the majority of FLAIR-positive ISL. These findings demonstrate the viability of using partially labelled data to develop reliable automated segmentation tools, which can support ongoing SVD monitoring and high-throughput biomarker extraction for large-scale clinical research.
\end{abstract}

% \begin{highlights}
% \item Evaluated six strategies for partially labelled WMH and ISL segmentation.
% \item Large-scale validation utilising 2,052 FLAIR MRI volumes across 12 datasets.
% \item Pseudolabelling proved most effective for SVD biomarker extraction.
% \end{highlights}

%% Keywords

\end{frontmatter}

\begin{figure}[htbp]
    \centering
    \includegraphics[width=\textwidth]{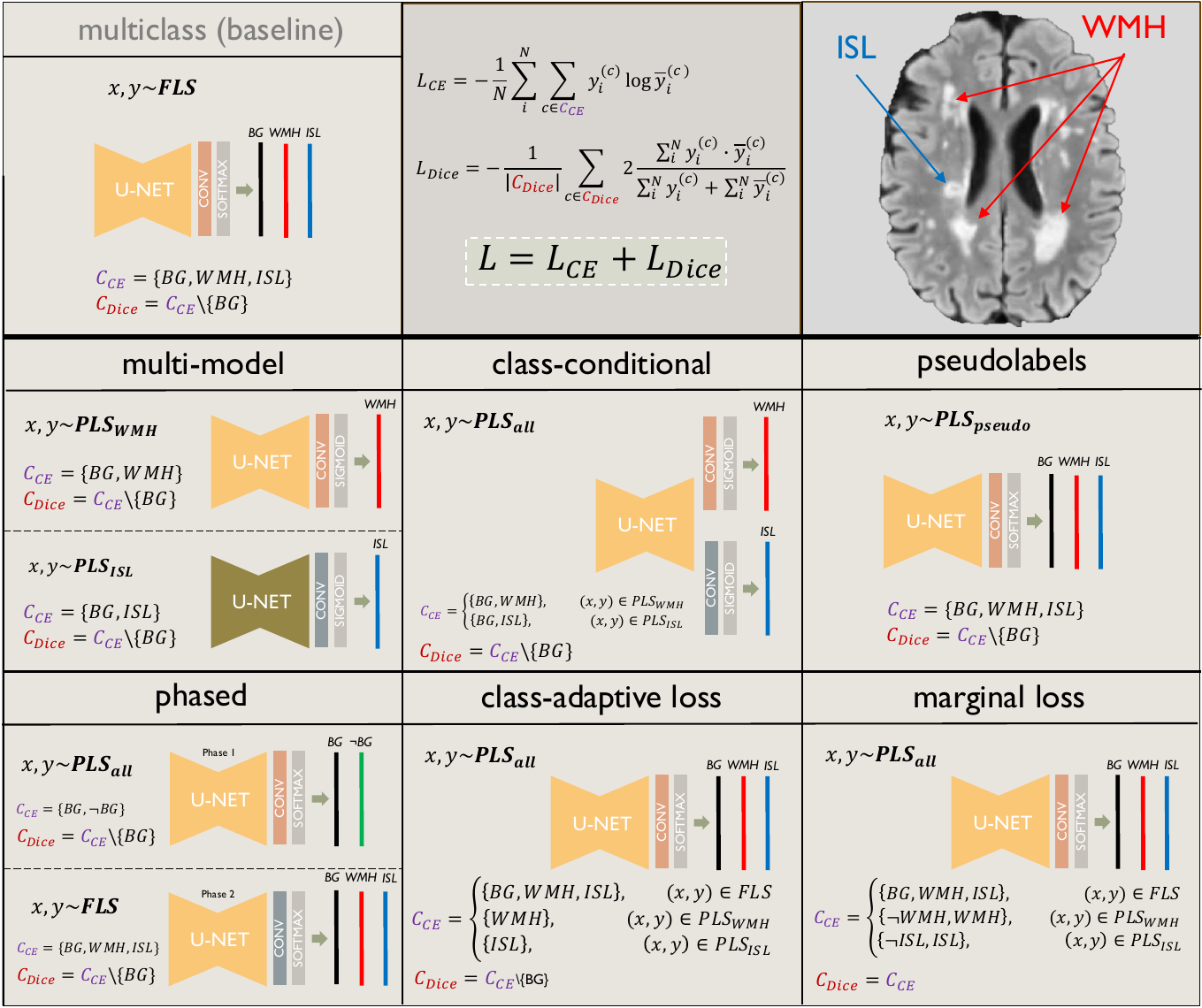}
    \caption{Schematic overview of the investigated training strategies for utilising partially labelled data. The \textbf{multiclass} baseline trains a standard multiclass model exclusively on the fully labelled subset (FLS). The \textbf{multi-model} approach trains separate binary networks for white matter hyperintensities (WMH) and ischaemic stroke lesions (ISL) using their respective partially labelled subsets (PLS). The \textbf{class-conditional} architecture features independent output heads for WMH and ISL; for samples with complete annotations, forward passes are executed for each head, and the average loss is computed. The \textbf{pseudolabels} approach trains a standard multiclass model by imputing missing annotations with model-generated pseudolabels. The \textbf{phased} approach pre-trains on the combined PLS by merging WMH and ISL into a single foreground class, before replacing the final classification layer to fine-tune on the FLS. The \textbf{class-adaptive loss} formulation computes the error exclusively across the available ground truth classes. Finally, the \textbf{marginal loss} strategy addresses missing labels by merging unannotated target classes into the background. BG: background class, $x$: input FLAIR scan, $y$: ground truth, $PLS_{\text{all}} = PLS_{\text{WMH}} \cup PLS_{\text{ISL}}$, and $FLS \subset PLS_{\text{all}}$.}
    \label{fig:fig1}
\end{figure}

\section{Introduction}

\subsection{Clinical motivation and challenges}

White matter hyperintensities (WMH) and \acrfull{isl} are neuroimaging markers of cerebral \acrfull{svd}, a major contributing factor to stroke and dementia \cite{debette2019clinical}. The clinical relevance of automating the segmentation of these lesions in \acrfull{flair} \acrfull{mri} is twofold: it facilitates the early diagnosis and monitoring of SVD, and it enables the large-scale extraction of imaging biomarkers necessary for clinical research and preventative trial design. By executing a single consistent segmentation policy across multiple datasets, computational approaches bypass the high variance and differing implicit biases introduced by relying on multiple human annotators.

A primary bottleneck in developing such automatic tools is the visually confounding nature of \acrshort{wmh} and \acrshort{isl}. Both pathologies usually present as hyperintense focal regions within the darker brain parenchyma on \acrshort{flair} sequences. Deep neural networks excel at isolating the type of nuanced features required to differentiate them, provided they are trained on datasets where both lesions are annotated. However, publicly available datasets meeting this annotation requirement do not currently exist, and privately held datasets remain small, limiting model generalisability. 

\subsection{Partially labelled segmentation}

To overcome the lack of fully annotated data, researchers must leverage partially labelled datasets, where images possess annotations for one class but lack ground truth for the other. Some approaches to this problem ignore class dependencies entirely, such as training independent binary models or training a single model using a shared backbone with a separate segmentation head for each class, or using the exact same model conditioned on the desired output class \cite{dmitriev2019learning,zhang2021dodnet,deng2022single}. While sufficient for semantically distinct objects, these independent formulations do not explicitly model the confounding spatial and visual relationships inherent to \acrshort{wmh} and \acrshort{isl}. Moreover, when these methods are applied to segmentation tasks like ours, where labels are mutually exclusive, an additional step is required to generate the final probability distribution over all classes.

A phased training approach was explored in \cite{llambias2024heterogeneous} to segment \acrshort{wmh}, \acrshort{isl}, and \acrfull{ms} lesions from partially labelled datasets (WMH challenge \cite{kuijf2019standardized}, ISLES 2022 \cite{hernandez2022isles}, MSSEG-1 \cite{commowick2021multiple} and MS08 \cite{styner20083d}). An initial training stage learns a foreground vs background task. Then, two different approaches to producing the final model are tested. The first approach involves reinitialising the final layer for multiclass segmentation, then fine-tuning the model using equally-weighted sampling of the different datasets. The second approach involves using equally-weighted dataset sampling to continue training on the initial binary foreground vs background task, followed by training a small classifier to predict the lesion class of each connected component that is predicted by the binary segmentation model, based on volume, surface area, centre of mass, orientation, and principle axis length. The results are not evaluated with respect to any alternative training strategies.

Other techniques for partially labelled segmentation include marginal loss formulations \cite{fang2020multi,shi2021marginal,fidon2021label}, which merge missing labels into the background class, and pseudolabelling approaches. Existing pseudolabelling approaches include PaNN \cite{zhou2019prior}, which is designed for multi-organ segmentation, not pathological segmentation, and PSSNet \cite{liu2024many}, which employs a highly bespoke architecture and training strategy, making it unlikely to become widely adopted. 

Consequently, a gap remains in the literature: the field lacks a systematic evaluation of how foundational, accessible training strategies handle the specific challenge of partially labelled, mutually exclusive, and visually confounding pathologies.

\subsection{Rationale and contributions}

Rather than introducing another highly complex or task-specific architecture, the rationale of this study is to systematically evaluate practical, easy-to-implement, and conceptually simple, training strategies for the uniquely challenging partially labelled \acrshort{wmh} and \acrshort{isl} segmentation task. 

To this end, we only consider methods based on fully supervised U-Net training. Acceptable modifications to the standard training paradigm include: training multiple models (or segmentation heads); training in multiple stages; or modifications to the loss function which do not necessitate a change in training recipe. We do not evaluate any methods relying on non-standard-U-Net architectures or training paradigms, such as self-supervised or adversarial learning.

The novelty of this work lies in our comprehensive comparison. 
By leveraging a large composite dataset, we establish a benchmark that guides future methodological choices for partially labelled \acrshort{wmh} and \acrshort{isl} segmentation.

\begin{enumerate}
    \item We address the literature gap by systematically comparing six techniques for handling partially labelled data, evaluating them on the clinically relevant and visually confounding task of simultaneous \acrshort{wmh} and \acrshort{isl} segmentation.
    \item We establish robust benchmarks by training and testing across 12 fully or partially labelled datasets, comprising over 2,000 MRI volumes, representing one of the most comprehensive evaluations of its kind.
    \item We perform an extensive qualitative analysis detailing the behaviour of our best model relative to variable ground truth generation policies, offering practical insights that go beyond purely quantitative evaluation.
\end{enumerate}

To additionally support further research on \acrshort{wmh} and \acrshort{isl} segmentation model analysis, in \ref{appendix A} we propose novel algorithms to perform robust region-of-interest (ROI)-based analysis, and we perform such analysis on our best model's predictions.

\section{Data}

\subsection{Mask availability and training splits}

\textbf{Table \ref{tab:data}} summarises the allocation of scans from 12 datasets into training, validation, and test sets, detailing the number of subjects per partition with available and non-empty ground truth masks. To ensure a balanced representation of lesion burden, the data was randomly partitioned into a 36:18:40 split (training, validation, and test, respectively), stratified within each dataset by \acrshort{wmh} and \acrshort{isl} volume tertiles. To assess out-of-distribution generalisation, the entirety of the ESS and LINCHPIN cohorts were reserved exclusively for the test set.

% Data splits table
\begin{table}[htbp]
\caption{Dataset splits. N is the total number of subjects, and WMH and ISL are the number of subjects with WMH and ISL ground truth masks, where X/Y represents X non-empty ground truth masks out of Y available ground truth masks. $FLS$, $PLS_{\text{WMH}}$, and $PLS_{\text{ISL}}$ are different fully labelled or partially labelled subsets of the same total pool of training data. Any subjects with WMH labels is in $PLS_{\text{WMH}}$, any subject with ISL labels is in $PLS_{\text{ISL}}$, and any subject with both WMH and ISL labels is in $FLS$ (and naturally also in $PLS_{\text{WMH}}$ and $PLS_{\text{ISL}}$). $FLS$: fully labelled subset, $PLS$: partially labelled subset, WMH: white matter hyperintensitiies, ISL: ischaemic stroke lesions.}
\label{tab:data}
\resizebox{\textwidth}{!}{%
\begin{tabular}{l|ccc|ccc|ccccccccc}
\hline
Dataset &
  \multicolumn{3}{c|}{Test} &
  \multicolumn{3}{c|}{Validation} &
  \multicolumn{9}{c}{Train} \\
 &
   &
   &
   &
   &
   &
   &
  \multicolumn{3}{c|}{$FLS$} &
  \multicolumn{3}{c|}{$PLS_{\text{WMH}}$} &
  \multicolumn{3}{c}{$PLS_{\text{ISL}}$} \\
 &
  N &
  WMH &
  ISL &
  N &
  WMH &
  ISL &
  N &
  WMH &
  \multicolumn{1}{c|}{ISL} &
  N &
  WMH &
  \multicolumn{1}{c|}{ISL} &
  N &
  WMH &
  ISL \\ \hline
MSS1 &
  32 &
  32/32 &
  31/32 &
  4 &
  4/4 &
  4/4 &
  28 &
  28/28 &
  \multicolumn{1}{c|}{28/28} &
  58 &
  58/58 &
  \multicolumn{1}{c|}{28/28} &
  28 &
  28/28 &
  28/28 \\
MSS2 &
  56 &
  56/56 &
  55/56 &
  6 &
  6/6 &
  5/6 &
  49 &
  49/49 &
  \multicolumn{1}{c|}{48/49} &
  50 &
  50/50 &
  \multicolumn{1}{c|}{48/49} &
  142 &
  49/49 &
  107/142 \\
MSS3 &
  114 &
  114/114 &
  104/114 &
  12 &
  12/12 &
  12/12 &
  101 &
  101/101 &
  \multicolumn{1}{c|}{91/101} &
  102 &
  102/102 &
  \multicolumn{1}{c|}{91/101} &
  101 &
  101/101 &
  91/101 \\
LBC1936 &
  333 &
  327/333 &
  47/333 &
  34 &
  33/34 &
  5/34 &
  297 &
  289/297 &
  \multicolumn{1}{c|}{43/297} &
  298 &
  290/298 &
  \multicolumn{1}{c|}{43/298} &
  300 &
  290/298 &
  43/300 \\
LBC1921 &
  22 &
  22/22 &
  0/0 &
  3 &
  3/3 &
  0/0 &
   &
   &
  \multicolumn{1}{c|}{} &
  18 &
  18/18 &
  \multicolumn{1}{c|}{0/0} &
   &
   &
   \\
WMH-ch &
  85 &
  85/85 &
  0/0 &
  9 &
  9/9 &
  0/0 &
   &
   &
  \multicolumn{1}{c|}{} &
  76 &
  76/76 &
  \multicolumn{1}{c|}{0/0} &
   &
   &
   \\
BRATS &
  17 &
  0/0 &
  0/17 &
  2 &
  0/0 &
  0/2 &
   &
   &
  \multicolumn{1}{c|}{} &
   &
   &
  \multicolumn{1}{c|}{} &
  14 &
  0/0 &
  0/14 \\
ISLES &
  65 &
  0/0 &
  64/65 &
  7 &
  0/0 &
  7/7 &
   &
   &
  \multicolumn{1}{c|}{} &
   &
   &
  \multicolumn{1}{c|}{} &
  57 &
  0/0 &
  55/57 \\
SOOP &
  192 &
  0/0 &
  189/192 &
  20 &
  0/0 &
  20/20 &
   &
   &
  \multicolumn{1}{c|}{} &
   &
   &
  \multicolumn{1}{c|}{} &
  172 &
  0/0 &
  172/172 \\
WSS &
  11 &
  0/0 &
  10/11 &
  2 &
  0/0 &
  2/2 &
   &
   &
  \multicolumn{1}{c|}{} &
   &
   &
  \multicolumn{1}{c|}{} &
  9 &
  0/0 &
  9/9 \\
ESS &
  78 &
  54/54 &
  75/75 &
   &
   &
   &
   &
   &
  \multicolumn{1}{c|}{} &
   &
   &
  \multicolumn{1}{c|}{} &
   &
   &
   \\
LINCHPIN &
  2 &
  0/0 &
  0/2 &
   &
   &
   &
   &
   &
  \multicolumn{1}{c|}{} &
   &
   &
  \multicolumn{1}{c|}{} &
   &
   &
   \\ \hline
TOTAL &
  1007 &
  690/696 &
  575/897 &
  99 &
  67/68 &
  55/87 &
  475 &
  467/475 &
  \multicolumn{1}{c|}{210/475} &
  602 &
  594/602 &
  \multicolumn{1}{c|}{210/476} &
  823 &
  468/476 &
  505/823 \\ \hline
\end{tabular}%
}
\end{table}

\subsection{Dataset details}

Demographic characteristics and lesion volumes (where data is available) are presented in \textbf{Tables \ref{tab:demo_all}, \ref{tab:demo_train}, \ref{tab:demo_val},} and \textbf{\ref{tab:demo_test}}. Probabilistic spatial maps illustrating the \acrshort{wmh} and \acrshort{isl} distributions for each dataset are provided in \ref{appendix B gt}. A brief overview of each dataset follows below, with further details available in the cited literature.

% Data demographics table all
\begin{table}[htbp]
\centering
\caption{All data demographics and lesions volumes. \{N\} is the number of samples with information available, if not all samples have information. SD: standard deviation, IQR: interquartile range, F: female.}
\label{tab:demo_all}
\resizebox{\textwidth}{!}{%
\begin{tabular}{l|ll|ll}
\hline
Dataset (N)   & \multicolumn{2}{c|}{Demographics} & \multicolumn{2}{c}{Lesion volumes}              \\ \cline{2-5} 
 &
  \multicolumn{1}{c}{\begin{tabular}[c]{@{}c@{}}Age (years),\\ mean $\pm$ SD\end{tabular}} &
  \multicolumn{1}{c|}{\begin{tabular}[c]{@{}c@{}}Sex (F),\\ \%\end{tabular}} &
  \multicolumn{1}{c}{\begin{tabular}[c]{@{}c@{}}WMH (ml),\\ median {[}IQR{]}\end{tabular}} &
  \multicolumn{1}{c}{\begin{tabular}[c]{@{}c@{}}ISL (ml),\\ median {[}IQR{]}\end{tabular}} \\ \hline
MSS1 (94)     & 67 $\pm$ 11          & 23\%          & 10.5 {[}21.2{]}         & 2.0 {[}7.8{]} \{64\}  \\
MSS2 (205)    & 68 $\pm$11 \{92\}    & 36\% \{92\}   & 11.9 {[}27.1{]} \{112\} & 1.8 {[}5.8{]} \{204\} \\
MSS3 (228)    & 66 $\pm$ 11          & 33\%          & 8.3 {[}15.3{]}          & 0.8 {[}2.3{]} \{227\} \\
LBC1936 (667) & 73 $\pm$ 1           & 47\%          & 6.1 {[}11.5{]} \{665\}  & 0.0 {[}0.0{]}         \\
LBC1921 (43)  & 92 $\pm$ 0           & 49\%          & 30.5 {[}42.1{]}         & -                     \\
WMH-ch (170)  & -                 & -             & 11.2 {[}19.8{]}         & -                     \\
BRATS (33)    & -                 & -             & -                       & 0.0 {[}0.0{]}         \\
ISLES (129)   & 74 $\pm$ 12 \{46\}   & 58\% \{59\}   & -                       & 5.0 {[}13.7{]}        \\
SOOP (384)    & -                 & -             & -                       & 4.3 {[}26.0{]}        \\
WSS (22)      & -                 & -             & -                       & 4.3 {[}25.4{]}        \\
ESS (78)      & 65 $\pm$ 17 \{24\}   & 21\% \{24\}   & 13.0 {[}19.6{]} \{54\}  & 2.3 {[}2.8{]} \{75\}  \\
LINCHPIN (2)  & -                 & -             & -                       & 0.0 {[}0.0{]}         \\ \hline
\end{tabular}%
}
\end{table}

% Data demographics table train
\begin{table}[htbp]
\caption{Training data demographics and lesions volumes. \{N\} is the number of samples with information available, if not all samples have information. SD: standard deviation, IQR: interquartile range, F: female.}
\label{tab:demo_train}
\resizebox{\textwidth}{!}{%
\begin{tabular}{l|ll|ll}
\hline
Dataset (N)   & \multicolumn{2}{c|}{Demographics} & \multicolumn{2}{c}{Lesion volumes}             \\ \cline{2-5} 
 &
  \multicolumn{1}{c}{\begin{tabular}[c]{@{}c@{}}Age (years),\\ mean $\pm$ SD\end{tabular}} &
  \multicolumn{1}{c|}{\begin{tabular}[c]{@{}c@{}}Sex (F),\\ N (\%)\end{tabular}} &
  \multicolumn{1}{c}{\begin{tabular}[c]{@{}c@{}}WMH (ml),\\ median {[}IQR{]}\end{tabular}} &
  \multicolumn{1}{c}{\begin{tabular}[c]{@{}c@{}}ISL (ml),\\ median {[}IQR{]}\end{tabular}} \\ \hline
MSS1 (58)     & 66 $\pm$ 11          & 24\%          & 7.1 {[}21.5{]}         & 1.8 {[}6.0{]} \{28\}  \\
MSS2 (143)    & 68 $\pm$ 11 \{53\}   & 36\% \{53\}   & 11.8 {[}23.1{]} \{50\} & 1.4 {[}5.1{]} \{142\} \\
MSS3 (102)    & 65 $\pm$ 12          & 37\%          & 8.1 {[}19.2{]}         & 0.8 {[}2.3{]} \{101\} \\
LBC1936 (300) & 73 $\pm$ 1           & 46\%          & 6.2 {[}11.1{]} \{298\} & 0.0 {[}0.0{]}         \\
LBC1921 (18)  & 92 $\pm$ 0           & 61\%          & 30.6 {[}40.8{]}        & -                     \\
WMH-ch (76)   & -                 & -             & 9.2 {[}14.8{]}         & -                     \\
BRATS (14)    & -                 & -             & -                      & 0.0 {[}0.0{]}         \\
ISLES (57)    & 76 $\pm$ 9 \{24\}    & 62\% \{29\}   & -                      & 5.1 {[}13.1{]}        \\
SOOP (172)    & -                 & -             & -                      & 4.3 {[}21.5{]}        \\
WSS (9)       & -                 & -             & -                      & 1.6 {[}14.9{]}        \\
ESS (0)       & -                 & -             & -                      & -                     \\
LINCHPIN (0)  & -                 & -             & -                      & -                     \\ \hline
\end{tabular}%
}
\end{table}

% Data demographics table val
\begin{table}[htbp]
\caption{Validation data demographics and lesions volumes. \{N\} is the number of samples with information available, if not all samples have information. SD: standard deviation, IQR: interquartile range, F: female.}
\label{tab:demo_val}
\resizebox{\textwidth}{!}{%
\begin{tabular}{l|ll|ll}
\hline
Dataset (N)  & \multicolumn{2}{c|}{Demographics} & \multicolumn{2}{c}{Lesion volumes} \\ \cline{2-5} 
 &
  \multicolumn{1}{c}{\begin{tabular}[c]{@{}c@{}}Age (years),\\ mean $\pm$ SD\end{tabular}} &
  \multicolumn{1}{c|}{\begin{tabular}[c]{@{}c@{}}Sex (F),\\ N (\%)\end{tabular}} &
  \multicolumn{1}{c}{\begin{tabular}[c]{@{}c@{}}WMH (ml),\\ median {[}IQR{]}\end{tabular}} &
  \multicolumn{1}{c}{\begin{tabular}[c]{@{}c@{}}ISL (ml),\\ median {[}IQR{]}\end{tabular}} \\ \hline
MSS1 (4)     & 63 $\pm$ 8          & 25\%           & 21.9 {[}18.4{]} & 4.4 {[}12.7{]}  \\
MSS2 (6)     & -                & -              & 23.0 {[}78.4{]} & 2.7 {[}1.5{]}   \\
MSS3 (12)    & 66 $\pm$ 10         & 17\%           & 7.8 {[}10.7{]}  & 0.7 {[}2.3{]}   \\
LBC1936 (34) & 73 $\pm$ 1          & 44\%           & 6.5 {[}9.5{]}   & 0.0 {[}0.0{]}   \\
LBC1921 (3)  & 92 $\pm$ 0          & 33\%           & 18.0 {[}24.3{]} & -               \\
WMH-ch (9)   & -                & -              & 7.5 {[}14.2{]}  & -               \\
BRATS (2)    & -                & -              & -               & 0.0 {[}0.0{]}   \\
ISLES (7)    & 41 $\pm$ 0 \{1\}    & 100\% \{1\}    & -               & 2.3 {[}8.8{]}   \\
SOOP (20)    & -                & -              & -               & 4.3 {[}27.1{]}  \\
WSS (2)      & -                & -              & -               & 28.8 {[}26.0{]} \\
ESS (0)      & -                & -              & -               & -               \\
LINCHPIN (0) & -                & -              & -               & -               \\ \hline
\end{tabular}%
}
\end{table}

% Data demographics table test
\begin{table}[htbp]
\caption{Test data demographics and lesions volumes. \{N\} is the number of samples with information available, if not all samples have information. SD: standard deviation, IQR: interquartile range, F: female.}
\label{tab:demo_test}
\resizebox{\textwidth}{!}{%
\begin{tabular}{l|ll|ll}
\hline
Dataset (N)   & \multicolumn{2}{c|}{Demographics} & \multicolumn{2}{c}{Lesion volumes}            \\ \cline{2-5} 
 &
  \multicolumn{1}{c}{\begin{tabular}[c]{@{}c@{}}Age (years),\\ mean $\pm$ SD\end{tabular}} &
  \multicolumn{1}{c|}{\begin{tabular}[c]{@{}c@{}}Sex (F),\\ \%\end{tabular}} &
  \multicolumn{1}{c}{\begin{tabular}[c]{@{}c@{}}WMH (ml),\\ median {[}IQR{]}\end{tabular}} &
  \multicolumn{1}{c}{\begin{tabular}[c]{@{}c@{}}ISL (ml),\\ median {[}IQR{]}\end{tabular}} \\ \hline
MSS1 (32)     & 69 $\pm$ 11          & 22\%          & 15.7 {[}18.3{]}        & 2.0 {[}13.2{]}       \\
MSS2 (56)     & 67 $\pm$ 11 \{39\}   & 36\% \{39\}   & 11.9 {[}31.3{]}        & 3.2 {[}10.7{]}       \\
MSS3 (114)    & 67 $\pm$ 11          & 32\%          & 8.5 {[}15.1{]}         & 0.8 {[}2.4{]}        \\
LBC1936 (333) & 73 $\pm$ 1           & 48\%          & 6.1 {[}12.1{]}         & 0.0 {[}0.0{]}        \\
LBC1921 (22)  & 92 $\pm$ 0           & 41\%          & 34.2 {[}43.0{]}        & -                    \\
WMH-ch (85)   & -                 & -             & 12.3 {[}21.5{]}        & -                    \\
BRATS (17)    & -                 & -             & -                      & 0.0 {[}0.0{]}        \\
ISLES (65)    & 73 $\pm$ 13 \{21\}   & 52\% \{29\}   & -                      & 4.8 {[}13.9{]}       \\
SOOP (192)    & -                 & -             & -                      & 4.2 {[}27.6{]}       \\
WSS (11)      & -                 & -             & -                      & 7.0 {[}43.5{]}       \\
ESS (78)      & 65 $\pm$ 17 \{24\}   & 21\% \{24\}   & 13.0 {[}19.6{]} \{54\} & 2.3 {[}2.8{]} \{75\} \\
LINCHPIN (2)  & -                 & -             & -                      & 0.0 {[}0.0{]}        \\ \hline
\end{tabular}%
}
\end{table}

\paragraph{MSS1 \cite{wardlaw2009lacunar}}
This cohort comprises patients presenting with their first clinically evident lacunar or mild cortical stroke, who were recruited prospectively and consecutively at an academic teaching hospital between 2005 and 2007. All neuroimaging data was acquired using a 1.5 Tesla Signa LX General Electric MRI scanner.

\paragraph{MSS2 \cite{wardlaw2017blood}}
This cohort comprises prospectively recruited patients presenting with clinically evident lacunar or mild cortical stroke. All neuroimaging data was acquired using a 1.5 Tesla Signa HDxt General Electric MRI scanner.

\paragraph{MSS3 \cite{clancy2021rationale}}
This dataset originates from a prospective, observational cohort study of patients presenting with clinically evident stroke syndromes. Patient inclusion was based on having either a recent infarct visible on a diagnostic MRI or CT scan, or the absence of other lesions that could explain their symptoms. All neuroimaging data was acquired using a 3 Tesla Siemens Prisma MRI scanner. The data utilised in this paper was obtained from the first study visit, which occurred a maximum of three months after patients initially presented to the Edinburgh and Lothian stroke services.

\paragraph{LBC1936 \cite{wardlaw2011brain}}
This dataset originates from an observational, longitudinal study of relatively healthy individuals who were born in the Lothian area in 1936 and residing in Edinburgh. Neuroimaging was performed at a mean participant age of 73 years. All neuroimaging data was acquired using a 1.5 Tesla Signa Horizon HDx General Electric MRI scanner.

\paragraph{LBC1921 \cite{wardlaw2011brain}}
This dataset originates from an observational, longitudinal study of relatively healthy individuals who were born in the Lothian area in 1921 and residing in Edinburgh. Neuroimaging was performed at a mean participant age of 91 years. All neuroimaging data was acquired using a 1.5 Tesla Signa Horizon HDx General Electric MRI scanner.

\paragraph{WMH-ch \cite{kuijf2019standardized}}
This publicly available dataset served as the basis for a MICCAI challenge conducted between 2017 and 2022 \cite{kuijf2019standardized}. The neuroimaging data was acquired using five different scanners (at 1.5 or 3 Tesla magnetic field strengths) from three distinct vendors, across three clinical sites in the Netherlands and Singapore. \acrshort{wmh} were delineated on the \acrshort{flair} images by a single expert using a contour drawing technique, in accordance with the original STRIVE guidelines \cite{wardlaw2013neuroimaging}. A second qualified observer performed an independent review of all cases, with any identified errors subsequently corrected by the primary annotator. 

\paragraph{BRATS \cite{menze2014multimodal}}
This publicly available dataset contains brain tumour annotations and has served as the basis for several MICCAI challenges \cite{menze2014multimodal}; the 2021 iteration was utilised in this study. Neuroimaging data was acquired under diverse clinical protocols using various scanners across multiple institutions. Ground truth annotations for \acrshort{wmh} and \acrshort{isl} are not provided, as small vessel disease biomarkers were not the focus of the challenge and are presumed to be clinically insignificant or absent in this cohort. To verify this, a single researcher with two years of experience in neuroimage analysis manually reviewed 33 \acrshort{flair} volumes to confirm the absence of \acrshort{isl}. The rationale for incorporating this cohort was to introduce images containing non-SVD pathologies, facilitating an evaluation of model robustness.

\paragraph{ISLES \cite{hernandez2022isles}}
This publicly available dataset of acute and sub-acute \acrshort{isl} has been featured in several MICCAI challenges \cite{hernandez2022isles}; the 2022 iteration was utilised in this study. Neuroimaging data was acquired across four clinical centres using diverse scanners and imaging protocols. Ground truth segmentations were generated via a semi-automated approach: initial masks were produced by a pre-trained U-Net \cite{ronneberger2015u} (trained on \acrshort{dwi}), followed by manual refinement by a specially trained medical student and further revision by a neuroradiologist in training. The segmentations were then reviewed by one of three attending neuroradiologists, each possessing over a decade of experience in stroke imaging. All available MRI sequences were consulted to delineate the final masks.

\paragraph{SOOP \cite{absher2024stroke}}
This publicly available dataset comprises a cohort of all acute ischaemic stroke patients registered in the GWTG database at Prisma Health-Upstate between early 2019 and late 2020. Neuroimaging data was acquired within 30 days of hospital admission (with the majority obtained within 48 hours), utilising heterogeneous scanning protocols. Ground truth \acrshort{isl} segmentations were manually delineated on the \acrshort{dwi} scans by three trained raters.

\paragraph{WSS \cite{wardlaw2013clinical}}
This dataset originates from a prospective study conducted across three UK stroke centres between 2008 and 2010. The cohort comprises patients presenting with potentially disabling acute ischaemic stroke. The study population had a median age of 71 years, of which 39\% were female.

\paragraph{ESS \cite{del2015comparison}}
This dataset originates from a prospective, consecutive study of patients presenting with their first-ever acute ischaemic stroke at a single regional hospital between 2002 and 2005. The included cohort represents a subset of sample 1 detailed in \cite{del2015comparison}, with a mean patient age of 69 years. All neuroimaging data was acquired using a single 1.5 Tesla Signa Horizon HDxt General Electric MRI scanner.

\paragraph{LINCHPIN \cite{rodrigues2018edinburgh}}
This dataset originates from a prospective, population-based cohort of patients presenting with their first ever intracerebral haemorrhage. Ground truth annotations are not provided for this dataset. However, a single researcher with two years of experience in neuroimage analysis manually reviewed two MRI volumes exhibiting pronounced pathology to confirm the absence of \acrshort{isl}. The rationale for including these specific cases was to evaluate the model's susceptibility to predicting false-positive \acrshort{wmh} and \acrshort{isl} in the presence of haemorrhages.

\subsection{Preprocessing}

Because our models were trained exclusively on \acrshort{flair} sequences, datasets containing ground truth annotations delineated in other sequences required spatial alignment. For these cohorts, FSL-FLIRT\footnote{\url{https://fsl.fmrib.ox.ac.uk/fsl/fslwiki/FLIRT}} \cite{jenkinson_global_2001,jenkinson_improved_2002} was employed to perform rigid co-registration, mapping the annotated labels to the \acrshort{flair} space. 

Subsequently, all \acrshort{flair} volumes underwent a standardised preprocessing pipeline comprising the following sequential steps: (1) N4 bias field correction \cite{tustison2010n4itk} to mitigate intensity inhomogeneities; (2) spatial resampling to an isotropic resolution of $1 \times 1 \times 1 \text{ mm}^3$ (employing trilinear interpolation for both images and labels); (3) brain extraction using SynthStrip \cite{hoopes2022synthstrip}, followed by cropping the volume to the extracted brain bounding box and, if necessary, zero-padding to achieve a minimum spatial dimension of $160 \times 160 \times 160$ voxels; and (4) z-score standardisation to normalise voxel intensities to zero mean and unit variance.

A prerequisite for our experiments is that the target pathologies are visually discernible on the \acrshort{flair} sequence alone. This assumption is sometimes violated when ground truth annotations are derived from alternative imaging modalities. Specifically, relying on \acrshort{dwi} sequences for acute \acrshort{isl} delineation introduces a challenge, as the evolution of \acrshort{isl} manifests differently across \acrshort{dwi} and \acrshort{flair} acquisitions over time. To mitigate this, an automated filtering protocol was implemented to exclude volumes from the ISLES and SOOP datasets where the \acrshort{isl} was deemed imperceptible on the corresponding \acrshort{flair} scan. This filtering was executed prior to the aforementioned z-score standardisation using the following algorithmic steps: (1) connected component analysis was performed on each \acrshort{isl} mask; (2) image intensities were min-max normalised to the $[0, 1]$ range, with the 1st and 99th percentiles saturated; (3) SynthSeg \cite{billot2023synthseg} was utilised to generate anatomical brain segmentations, identify the bilateral anatomical region containing the majority of the \acrshort{isl} component voxels, and compute its mean intensity (explicitly excluding all \acrshort{isl} voxels); (4) the mean intensity of the \acrshort{isl} component itself was calculated; (5) scans were excluded if the difference between the regional background mean and the \acrshort{isl} mean was less than 0.05 for ISLES or 0.1 for SOOP, for any component; and (6) finally, scans were additionally excluded if the fraction of \acrshort{isl} voxels closer in intensity to the healthy tissue mean than to the \acrshort{isl} mean exceeded 20\% for ISLES or 10\% for SOOP, for any component. These dataset-specific thresholds were derived empirically, through visual evaluation. The filtering process discarded 48\% of ISLES scans and 71\% of SOOP scans.

\section{Methods}

\subsection{Deep learning pipeline}

\paragraph{Model architecture} 
All segmentation networks were instantiated using the nnU-Net-based \cite{isensee2021nnu} DynUNet architecture from the MONAI library \cite{cardoso2022monai}, implemented within the PyTorch Lightning framework \cite{lightning}. The network features a six-level resolution hierarchy comprising 32, 64, 128, 256, 320, and 320 feature maps, respectively. The architecture incorporates $3 \times 3 \times 3$ convolutional kernels, residual blocks, instance normalisation, Leaky ReLU activation functions (negative slope coefficient $\alpha = 0.01$), and three levels of deep supervision.

\paragraph{Data augmentation and patch sampling} 
Data loading and augmentation pipelines were constructed using TorchIO \cite{perez2021torchio}. We adopted the comprehensive data augmentation scheme proposed by nnU-Net. During training, the following transformations were applied with their respective probabilities: random spatial flipping along all axes ($p=0.5$); random rotations with $\theta \sim \mathcal{U}\left(-90^\circ, 90^\circ\right)$ per axis ($p=0.2$); random isotropic scaling by a factor $f \sim \mathcal{U}\left(0.7, 1.4\right)$ ($p=0.2$); additive Gaussian noise sampled from $\mathcal{N}\left(0, 0.1\right)$ ($p=0.15$); Gaussian blurring with $\sigma \sim \mathcal{U}\left(0.5, 1.5\right)$ ($p=0.1$); brightness adjustment by a factor $f \sim \mathcal{U}\left(0.7, 1.3\right)$ ($p=0.15$); contrast adjustment by a factor $f \sim \mathcal{U}\left(0.65, 1.5\right)$ ($p=0.15$); gamma transformations with $\gamma \sim \mathcal{U}\left(0.7, 1.5\right)$ ($p=0.15$), including a 15\% probability of intensity inversion prior to the transform; and low-resolution simulation with a downsampling factor $f \sim \mathcal{U}\left(1, 4\right)$ ($p=0.25$, expanding upon nnU-Net's maximum downsampling factor of 2). Furthermore, we introduced the following \acrshort{mri}-specific augmentations, using TorchIO default parameters, to build robustness against common acquisition artifacts: random motion simulation ($p=0.05$), ghosting artifacts ($p=0.05$), spike artifacts ($p=0.05$), and bias field inhomogeneities ($p=0.05$). 

For model training, spatial patches of $160 \times 160 \times 160$ voxels were extracted. The sampling strategy was configured such that 30\% of the extracted patches were centred on background voxels, while the remaining 70\% were distributed across the available foreground target classes.

\paragraph{Training and optimisation} 
We used an equally weighted sum of Dice loss and cross-entropy (CE) loss, aligning with the standard nnU-Net configuration \cite{isensee2021nnu}. The background class was excluded from the Dice loss computation (with the exception of the marginal loss formulation). The models were optimised using stochastic gradient descent (SGD) with Nesterov momentum (initial learning rate $= 0.01$, momentum $= 0.99$), governed by a polynomial learning rate decay schedule (power $= 0.9$).

Training was conducted with a batch size of 6 for a maximum of 2,000 epochs, and the model weights yielding the highest mean \acrfull{dsc} on the validation set were retained for final inference. A comparison of key implementation details and computational costs is shown in \textbf{Table \ref{tab:cost}}

To ensure robust predictive performance, each model configuration was trained three times using distinct random initialisation seeds. The final ensemble predictions were derived by computing the average of the probabilistic (post-\texttt{softmax} or post-\texttt{sigmoid}) outputs from these three runs.

\paragraph{Code} 
\url{https://github.com/Jesse-Phitidis/partially_labelled_WMH_ISL_seg}

\begin{table}[]
\centering
\caption{Summary of key implementation details and computational costs for each method. Training stages include stages which can be completed in parallel (multi-model) and stages which must be completed in series (multi-model (TS) stage 3, pseudolabels, phased). Training times are for a workstation equipped with an NVIDIA RTX 4090 GPU and 16-core CPU and will vary depending on hardware. The relative training time between methods may also vary depending on the sizes of the fully labelled subset (FLS) and partially labelled subsets (PLS) respectively. *The number of training stages for the pseudolabels model is one plus the number of training stages for the pseudolabel generating model (in our case the marginal loss model). Params is short for parameters and describes the total number of weights and biases for all networks involved in model inference.}
\label{tab:cost}
\resizebox{\textwidth}{!}{%
\begin{tabular}{l|c|l|c|l}
\hline
Method              & Training stages & Training time & Inference steps & Params \\ \hline
multiclass          & 1               & 61 hours      & 1               & 31.4 M \\ \hline
multi-model         & 2               & 178 hours     & 2               & 62.8 M \\ \hline
multi-model (TS)    & 3               & 178 hours     & 2               & 62.8 M \\ \hline
class-conditional   & 1               & 128 hours     & 1               & 31.4 M \\ \hline
pseudolabels        & 2*              & 245 hours     & 1               & 31.4 M \\ \hline
phased              & 2               & 162 hours     & 1               & 31.4 M \\ \hline
class-adaptive loss & 1               & 125 hours     & 1               & 31.4 M \\ \hline
marginal loss       & 1               & 125 hours     & 1               & 31.4 M \\ \hline
\end{tabular}%
}
\end{table}

\subsection{Supervision methods} \label{sec: supervision}

We systematically evaluated six training strategies for utilising partially labelled data, detailed below and visualised schematically in \textbf{Figure \ref{fig:fig1}}. To formalise these methods, we define five training data subsets: $FLS$ (the fully labelled subset); $PLS_{\text{WMH}}$ and $PLS_{\text{ISL}}$ (the subsets with available ground truth for \acrshort{wmh} or \acrshort{isl}, respectively); $PLS_{\text{all}}$ (the union of all available data, such that $PLS_{\text{all}} = PLS_{\text{WMH}} \cup PLS_{\text{ISL}}$ and $FLS \subset PLS_{\text{all}}$); and $PLS_{\text{pseudo}}$ (which denotes $PLS_{\text{all}}$ with model-generated pseudolabels imputing the missing annotations).

During optimisation, a training instance comprising an input \acrshort{flair} volume, $x$, and its corresponding ground truth segmentation, $y$, is sampled from the designated subset. The networks are optimised using the loss function, $L = L_{CE} + L_{Dice}$, defined as:

\begin{equation*}
    L_{CE} = - \frac{1}{N} \sum_{i=1}^N \sum_{c \in C_{CE}} y_i^{(c)} \log \overline{y}_i^{(c)}
\end{equation*}

\begin{equation*}
    L_{Dice} = - \frac{1}{|C_{Dice}|} \sum_{c \in C_{Dice}} 2 \frac{\sum_{i=1}^N y_i^{(c)} \cdot \overline{y}_i^{(c)}}{\sum_{i=1}^N y_i^{(c)} + \sum_{i=1}^N \overline{y}_i^{(c)}}.
\end{equation*}

\vspace{10pt}
\noindent where $N$ represents the total number of voxels; $C_{CE}$ and $C_{Dice}$ denote the sets of target classes (which may include the background (BG), \acrshort{wmh}, and \acrshort{isl}) over which the cross-entropy and Dice losses are respectively computed; and $\overline{y}$ represents the model's predicted probability map.

\paragraph{Multiclass (baseline)}
This approach serves as the standard experimental baseline, utilising a multiclass architecture trained exclusively on the fully labelled subset ($x, y \sim FLS$). The cross-entropy loss is computed across all available classes ($C_{CE} = \{\text{BG}, \text{WMH}, \text{ISL}\}$), whereas the Dice loss explicitly excludes the background class ($C_{Dice} = C_{CE}\backslash\{\text{BG}\}$).

\paragraph{Multi-model}
This strategy trains two independent binary segmentation networks. The first network is trained exclusively to segment \acrshort{wmh} using its corresponding partially labelled subset ($x, y \sim PLS_{\text{WMH}}$), with the loss computed over $C_{CE} = \{\text{BG}, \text{WMH}\}$ and $C_{Dice} = C_{CE}\backslash\{\text{BG}\}$. The second network is trained to segment \acrshort{isl} using $x, y \sim PLS_{\text{ISL}}$, where $C_{CE} = \{\text{BG}, \text{ISL}\}$ and $C_{Dice} = C_{CE}\backslash\{\text{BG}\}$. During inference, to construct a unified multiclass probabilistic prediction, the minimum of the two independently predicted background probabilities is selected. This derived background probability, along with the two respective foreground probabilities, is subsequently normalised such that the sum across all classes equals one. To reduce the impact of potential calibration discrepancies between the two independent models, temperature scaling (TS) \cite{guo2017calibration} is applied to each network using the validation set. This post hoc calibration technique adjusts the entropy of the predicted distribution by dividing the pre-activation (pre-\texttt{softmax} or pre-\texttt{sigmoid}) logits by a learned scalar parameter. This parameter was optimised using the cross-entropy loss via the L-BFGS optimiser (learning rate = 0.01, maximum of 10,000 iterations). We denote this calibrated variant as the multi-model (TS) approach.

\paragraph{Class-conditional}
This model features a shared feature extractor terminating in two independent segmentation heads (i.e., distinct final convolutional layers) dedicated to \acrshort{wmh} and \acrshort{isl}, respectively. Training instances are sampled from the combined dataset ($x,y \sim PLS_{\text{all}}$), and the target class sets for the loss computation are defined dynamically based on label availability:

\begin{equation*}
    C_{CE} =
\begin{cases}
\{\text{BG}, \text{WMH}\}, & (x, y) \in PLS_{\text{WMH}} \\
\{\text{BG}, \text{ISL}\}, & (x, y) \in PLS_{\text{ISL}}
\end{cases}
\end{equation*}

\vspace{10pt}
\noindent where $C_{Dice} = C_{CE}\backslash\{\text{BG}\}$. For partially annotated instances, the loss is computed and backpropagated exclusively through the segmentation head corresponding to the available ground truth. For fully annotated instances, a forward pass is executed for each head, and the network is optimised using the average of the two resulting losses. This architectural design is analogous to the approach proposed in \cite{he2024vista3d}, omitting the multi-layer perceptron (MLP) component. During inference, unified multiclass probabilistic predictions are derived using the identical normalisation procedure described for the multi-model approach.

\paragraph{Pseudolabels}
This strategy utilises a standard multiclass architecture trained on the fully imputed dataset ($x, y \sim PLS_{\text{pseudo}}$), with the loss computed across all classes ($C_{CE} = \{\text{BG}, \text{WMH}, \text{ISL}\}$) and the background excluded from the Dice formulation ($C_{Dice} = C_{CE}\backslash\{\text{BG}\}$). Missing annotations were imputed using pseudolabels generated by the marginal loss model (detailed below), which was selected for this task due to its strong performance on the validation set, architectural simplicity, and theoretical elegance. During pseudolabel generation, the predictions of the marginal loss model were post-processed using the available partial ground truth. Specifically, to prevent contradictory false-positives, any pseudolabelled predictions were explicitly suppressed in voxels where the alternative foreground class was already confirmed by the available ground truth mask.

\paragraph{Phased}
This strategy employs a sequential, two-stage training paradigm. In the pre-training phase, a binary segmentation network is trained for 2,000 epochs to differentiate the background from a unified foreground class (representing the union of \acrshort{wmh} and \acrshort{isl}). This phase utilises the entirety of the partially labelled data ($x, y \sim PLS_{\text{all}}$), with the target class sets defined as $C_{CE} = \{\text{BG}, \lnot \text{BG}\}$ and $C_{Dice} = C_{CE}\backslash\{\text{BG}\}$. Following this initialisation, the final two-channel segmentation layer is discarded and replaced with a newly initialised three-channel layer. The network is subsequently fine-tuned exclusively on the fully labelled subset ($x, y \sim FLS$) for an additional 1,000 epochs, following the same optimisation protocol established for the multiclass baseline. One obvious limitation of this training strategy is that during the first training stage, predictions of unlabelled lesions are considered to be false positives.

\paragraph{Class-adaptive loss}
This approach employs a standard multiclass architecture trained on the combined dataset ($x,y \sim PLS_{\text{all}}$). The target class sets for the loss computation are dynamically adjusted based on the availability of ground truth annotations for each sample:

\begin{equation*}
    C_{CE} =
\begin{cases}
\{\text{BG}, \text{WMH}, \text{ISL}\}, & (x, y) \in FLS \\
\{\text{WMH}\}, & (x, y) \in PLS_{\text{WMH}} \\
\{\text{ISL}\}, & (x, y) \in PLS_{\text{ISL}}
\end{cases}
\end{equation*}

\vspace{10pt}
\noindent where $C_{Dice} = C_{CE}\backslash\{\text{BG}\}$. Under this class-adaptive formulation, the loss is computed exclusively across the output channels corresponding to the available ground truth labels. If any foreground annotation is absent for a given training instance (i.e., the instance is not in the $FLS$), the background class must be treated as missing; this prevents the model from incorrectly penalising valid (but unannotated)  lesions as false-negative background.

\paragraph{Marginal loss formulation}
This strategy similarly utilises a multiclass model trained on the combined dataset ($x,y \sim PLS_{\text{all}}$), with the target class sets defined as follows:

\begin{equation*}
    C_{CE} =
\begin{cases}
\{\text{BG}, \text{WMH}, \text{ISL}\}, & (x, y) \in FLS \\
\{\lnot \text{WMH}, \text{WMH}\}, & (x, y) \in PLS_{\text{WMH}} \\
\{\lnot \text{ISL},\text{ISL}\}, & (x, y) \in PLS_{\text{ISL}}
\end{cases}
\end{equation*}

\vspace{10pt}
\noindent and, notably, $C_{Dice} = C_{CE}$ (meaning the background is included in the Dice calculation). Rather than selectively computing the loss only for available annotations---as is done in the class-adaptive approach---this method employs the marginal loss principle. Missing target classes are merged into a composite background class (denoted as $\lnot \text{WMH}$ or $\lnot \text{ISL}$), and the network is explicitly trained to predict the sum of probabilities for these merged classes.

% Describe evaluation method
\subsection{Evaluation}

\paragraph{Core segmentation metrics}
The different training and inference methods tested may bias certain approaches toward lower or higher entropy predictions, meaning the ideal operating point may vary. For this reason, it was important to choose our primary evaluation metric appropriately. We selected the \acrfull{ap} (also known as the area under the precision-recall curve) since it is operating-point independent. We additionally employed the \acrfull{dsc}, \acrfull{avd} (as a percentage of intracranial volume (ICV)), \acrfull{asd}, \acrfull{lpre}, and \acrfull{lrec}. 

With the exception of \acrshort{avd}, these metrics are undefined when the ground truth mask is empty. Hence, for \acrshort{isl}, we also report the subject-level false-positive rate among true negative cases. This metric is particularly relevant for our negative control datasets (BRATS and LINCHPIN), which consist entirely of empty \acrshort{isl} masks. We do not report on subject-level false-positives for \acrshort{wmh}. The only scans with empty but available \acrshort{wmh} masks originate from the LBC1936 dataset (mean test set age: 73); upon manual inspection, we found that these cases exhibited ambiguous presentations with trace \acrshort{wmh} volumes. This ambiguity highlights the  subjectivity of \acrshort{wmh} segmentation in elderly cohorts, making strict false-positive reporting less meaningful in this context.

\paragraph{Additional white matter hyperintensity (WMH) metrics} \label{metrics description}
The manual delineation of \acrshort{wmh} is an inherently subjective task, particularly at lesion boundaries and when determining the inclusion of small, punctate, or disconnected hyperintensities. Consequently, inter-rater variability, even among expert annotators, remains a challenge. To account for this spatial uncertainty, we additionally evaluated model performance using the \acrfull{ddsc}, a metric originally introduced in \cite{strumia2016white} for multiple sclerosis lesion segmentation. The \acrshort{ddsc} relaxes the strict spatial overlap requirements of the standard \acrshort{dsc} by ignoring false-positives and negatives that fall within a predefined distance tolerance ($\theta$) of the ground truth boundary. Because it accommodates minor boundary disagreements, it inherently follows that for any $\theta > 0$, \acrshort{ddsc} $\geq$ \acrshort{dsc}. In this study, we adopt a tolerance of $\theta = 2$ mm to facilitate a more nuanced evaluation and discussion regarding the inherent ambiguities in \acrshort{wmh} segmentation assessment.

\paragraph{Statistical analysis}
To evaluate the statistical significance of performance differences between models we employed a non-parametric paired bootstrap hypothesis test. For each metric and class, we generated 10,000 bootstrap samples by resampling the test subjects with replacement, over the set of subject with valid masks. For each bootstrap iteration, the mean performance metrics were calculated, and the distribution of paired differences between all model combinations was constructed. Two-tailed empirical $p$-values were derived from the fraction of bootstrap iterations where the difference inverted across zero. To control the error rate across all pairwise model comparisons, we applied a Holm-Bonferroni correction \cite{aickin1996adjusting} to the $p$-values, using a significance threshold of $\alpha = 0.05$. We used these same bootstrap iterations to derive confidence intervals on the metric values. All pairwise $p$-values are in \ref{appendix C}.

\section{Results and discussion}

\subsection{Utilising additional data boosts model performance}
% Table 1 and 1.5 (extra data helps)

The primary rationale for incorporating partially labelled data is the hypothesis that the additional training data will enhance model generalisation and overall performance. To empirically validate this within our dataset, we trained two independent binary segmentation models---one targeting \acrshort{wmh} and the other \acrshort{isl}. These models were evaluated under two training conditions: exclusively on the \acrshort{fls} and on their respective \acrshort{pls} (the networks trained on the \acrshort{pls} are the same networks which constitute the multi-model approach described in Section \ref{sec: supervision}). We evaluated the models on their respective binary segmentation tasks independently, rather than merging their outputs into a unified multi-class prediction. This isolates the specific impact of the expanded training data for each pathology class, thereby avoiding potential confounding effects introduced by multi-class segmentation dynamics. As demonstrated in \textbf{Table \ref{tab:extra_data}}, leveraging the additional partially labelled data yields statistically significant performance improvements across almost all metrics. The sole exception is \acrshort{lpre} for \acrshort{wmh}, which exhibited statistically non-significant decrease.

% binary model comparison
\begin{table}[htbp]
\caption{Results of independent binary segmentation models when trained on the partially labelled subset (PLS) for the target class, instead of the fully labelled subset (FLS). Reported as mean {[}95\% CI{]} with confidence intervals (CIs) calculated over a 10,000 sample bootstrap. Asterisk (*) shows statistically significant ($p$ \textless 0.05, two sided) difference according to a paired bootstrap test. ASD can only be calculated for cases where there is at least one predicted voxel, so these results are reported as mean [95\% CI] \{N\} where N is the number of test samples for which the metric was calculable, if N is less than the number of non-empty masks in the test set. AP: average precision, DSC: Dice similarity coefficient, AVD: absolute volume difference, ASD: average surface distance, LPRE: lesion-level precision, LREC: lesion-level recall, FP: percentage of empty ground truth masks containing at least one false-positive prediction, WMH: white matter hyperintensities, ISL: ischaemic stroke lesions.}
\label{tab:extra_data}
\resizebox{\textwidth}{!}{%
\begin{tabular}{l|cc|cc|cc|cc|cc|cc|c}
\hline
Train data & \multicolumn{2}{c|}{AP (\%)}                                                                                                              & \multicolumn{2}{c|}{DSC (\%)}                                                                                                             & \multicolumn{2}{c|}{AVD (\% ICV)}                                                                                                               & \multicolumn{2}{c|}{ASD (mm)}                                                                                                                & \multicolumn{2}{c|}{LPRE (\%)}                                                                                                             & \multicolumn{2}{c|}{LREC (\%)}                                                                                                            & FP (\%)                                                           \\ \cline{2-14} 
           & WMH                                                                 & ISL                                                                 & WMH                                                                 & ISL                                                                 & WMH                                                                    & ISL                                                                    & WMH                                                              & ISL                                                                       & WMH                                                                & ISL                                                                   & WMH                                                                 & ISL                                                                 & ISL                                                               \\ \hline
FLS        & \begin{tabular}[c]{@{}c@{}}75.07\\[-5pt] \smalltext{{[}73.61, 76.48{]}}\end{tabular}  & \begin{tabular}[c]{@{}c@{}}44.93\\[-5pt] \smalltext{{[}42.11, 47.74{]}}\end{tabular}  & \begin{tabular}[c]{@{}c@{}}65.99\\[-5pt] \smalltext{{[}64.67, 67.27{]}}\end{tabular}  & \begin{tabular}[c]{@{}c@{}}35.29\\[-5pt] \smalltext{{[}32.57, 37.99{]}}\end{tabular}  & \begin{tabular}[c]{@{}c@{}}0.2526 \\[-5pt] \smalltext{{[}0.2294, 0.2778{]}}\end{tabular} & \begin{tabular}[c]{@{}c@{}}0.6166\\[-5pt] \smalltext{{[}0.4644, 0.7846{]}}\end{tabular}  & \begin{tabular}[c]{@{}c@{}}1.71\\[-5pt] \smalltext{{[}1.39, 2.10{]}}\end{tabular}  & \begin{tabular}[c]{@{}c@{}}11.61 \{431\}\\[-5pt] \smalltext{{[}9.79, 13.50{]}}\end{tabular} & \begin{tabular}[c]{@{}c@{}}65.23\\[-5pt] \smalltext{{[}63.70, 66.79{]}}\end{tabular} & \begin{tabular}[c]{@{}c@{}}79.23\\[-5pt] \smalltext{{[}76.41, 82.03{]}}{]}\end{tabular} & \begin{tabular}[c]{@{}c@{}}39.22\\[-5pt] \smalltext{{[}37.77, 40.67{]}}\end{tabular}  & \begin{tabular}[c]{@{}c@{}}46.06\\[-5pt] \smalltext{{[}42.63, 49.53{]}}\end{tabular}  & \begin{tabular}[c]{@{}c@{}}12.11\\[-5pt] \smalltext{{[}8.61, 15.75{]}}\end{tabular} \\
PLS        & \begin{tabular}[c]{@{}c@{}}75.69*\\[-5pt] \smalltext{{[}74.18, 77.13{]}}\end{tabular} & \begin{tabular}[c]{@{}c@{}}53.80*\\[-5pt] \smalltext{{[}50.91, 56.67{]}}\end{tabular} & \begin{tabular}[c]{@{}c@{}}67.87*\\[-5pt] \smalltext{{[}66.53, 69.17{]}}\end{tabular} & \begin{tabular}[c]{@{}c@{}}44.23*\\[-5pt] \smalltext{{[}41.46, 46.97{]}}\end{tabular} & \begin{tabular}[c]{@{}c@{}}0.2081*\\[-5pt] \smalltext{{[}0.1856, 0.2329{]}}\end{tabular} & \begin{tabular}[c]{@{}c@{}}0.2228*\\[-5pt] \smalltext{{[}0.1791, 0.2726{]}}\end{tabular} & \begin{tabular}[c]{@{}c@{}}1.62*\\[-5pt] \smalltext{{[}1.28, 2.01{]}}\end{tabular} & \begin{tabular}[c]{@{}c@{}}7.43* \{450\}\\[-5pt] \smalltext{{[}6.01, 8.93{]}}\end{tabular}          & \begin{tabular}[c]{@{}c@{}}64.96\\[-5pt] \smalltext{{[}63.33, 66.57{]}}\end{tabular} & \begin{tabular}[c]{@{}c@{}}85.34*\\[-5pt] \smalltext{{[}82.93, 87.70{]}}\end{tabular}   & \begin{tabular}[c]{@{}c@{}}42.62*\\[-5pt] \smalltext{{[}41.12, 44.09{]}}\end{tabular} & \begin{tabular}[c]{@{}c@{}}51.56*\\[-5pt] \smalltext{{[}48.12, 55.05{]}}\end{tabular} & \begin{tabular}[c]{@{}c@{}}6.52*\\[-5pt] \smalltext{{[}3.99, 9.28{]}}\end{tabular}  \\ \hline
\end{tabular}%
}
\end{table}

% percentage change table
\begin{table}[htbp]
\caption{Percentage change in metrics on the test set when independent binary models are trained using the partially labelled subset (PLS) for the target class, instead of the fully labelled subset (FLS). Reported as mean {[}95\% CI{]} with confidence intervals (CIs) calculated over a 10,000 sample bootstrap. For FP, the absolute change is shown with arrows. Asterisk (*) shows statistically significant ($p$ \textless 0.05, two sided) difference according to a paired bootstrap test. Green: better, red: worse, blue: unchanged, AP: average precision, DSC: Dice similarity coefficient, AVD: absolute volume difference, ASD: average surface distance, LPRE: lesion-level precision, LREC: lesion-level recall, FP: percentage of empty ground truth masks containing at least one false-positive prediction ($\rightarrow$ shows absolute change), WMH: white matter hyperintensities, ISL: ischaemic stroke lesions.}
\label{tab:extra_data_2}
\resizebox{\textwidth}{!}{%
\begin{tabular}{l|cc|cc|cc|cc|cc|cc|c}
\hline
Test data &
  \multicolumn{2}{c|}{AP (\%)} &
  \multicolumn{2}{c|}{DSC (\%)} &
  \multicolumn{2}{c|}{AVD (\% ICV)} &
  \multicolumn{2}{c|}{ASD (mm)} &
  \multicolumn{2}{c|}{LPRE (\%)} &
  \multicolumn{2}{c|}{LREC (\%)} &
  FP (\%) \\ \cline{2-14} 
 &
  WMH &
  ISL &
  WMH &
  ISL &
  WMH &
  ISL &
  WMH &
  ISL &
  WMH &
  ISL &
  WMH &
  ISL &
  ISL \\ \hline
MSS1 &
  {\color[HTML]{59CA16} \begin{tabular}[c]{@{}c@{}}+1.91\\[-5pt] \smalltext{{[}+0.18, +4.14{]}}\end{tabular}} &
  {\color[HTML]{59CA16} \begin{tabular}[c]{@{}c@{}}+13.99*\\[-5pt] \smalltext{{[}+4.22, +30.41{]}}\end{tabular}} &
  {\color[HTML]{59CA16} \begin{tabular}[c]{@{}c@{}}+2.24\\[-5pt] \smalltext{{[}+0.00, +5.70{]}}\end{tabular}} &
  {\color[HTML]{D41159} \begin{tabular}[c]{@{}c@{}}-0.25\\[-5pt] \smalltext{{[}-4.36, +3.27{]}}\end{tabular}} &
  {\color[HTML]{D41159} \begin{tabular}[c]{@{}c@{}}+2.50\\[-5pt] \smalltext{{[}-6.12, +9.91{]}}\end{tabular}} &
  {\color[HTML]{59CA16} \begin{tabular}[c]{@{}c@{}}-5.55\\[-5pt] \smalltext{{[}-26.23, +16.80{]}}\end{tabular}} &
  {\color[HTML]{59CA16} \begin{tabular}[c]{@{}c@{}}-13.90\\[-5pt] \smalltext{{[}-35.82, +14.25{]}}\end{tabular}} &
  {\color[HTML]{59CA16} \begin{tabular}[c]{@{}c@{}}-18.61\\[-5pt] \smalltext{{[}-67.47, +39.23{]}}\end{tabular}} &
  {\color[HTML]{D41159} \begin{tabular}[c]{@{}c@{}}-8.10\\[-5pt] \smalltext{{[}-13.70, -2.50{]}}\end{tabular}} &
  {\color[HTML]{59CA16} \begin{tabular}[c]{@{}c@{}}+5.23\\[-5pt] \smalltext{{[}-5.91, +20.55{]}}\end{tabular}} &
  {\color[HTML]{59CA16} \textbf{\begin{tabular}[c]{@{}c@{}}+7.50*\\[-5pt] \smalltext{{[}+3.28, +12.40{]}}\end{tabular}}} &
  {\color[HTML]{59CA16} \begin{tabular}[c]{@{}c@{}}+2.32\\[-5pt] \smalltext{{[}-8.21, +13.82{]}}\end{tabular}} &
  {\color[HTML]{2B82D8} \begin{tabular}[c]{@{}c@{}}0.00\\[-5pt] \smalltext{{[}0.00, 0.00{]}}\\ 0 $\rightarrow$ 0 / 1\end{tabular}} \\
MSS2 &
  {\color[HTML]{D41159} \begin{tabular}[c]{@{}c@{}}-0.44\\[-5pt] \smalltext{{[}-1.57, +0.66{]}}\end{tabular}} &
  {\color[HTML]{59CA16} \begin{tabular}[c]{@{}c@{}}+5.50\\[-5pt] \smalltext{{[}+0.70, +11.91{]}}\end{tabular}} &
  {\color[HTML]{59CA16} \begin{tabular}[c]{@{}c@{}}+0.75\\[-5pt] \smalltext{{[}-0.22, +1.85{]}}\end{tabular}} &
  {\color[HTML]{59CA16} \begin{tabular}[c]{@{}c@{}}+4.41\\[-5pt] \smalltext{{[}-0.88, +11.50{]}}\end{tabular}} &
  {\color[HTML]{59CA16} \begin{tabular}[c]{@{}c@{}}-13.66*\\ {[}-19.89 , -6.62{]}\end{tabular}} &
  {\color[HTML]{D41159} \begin{tabular}[c]{@{}c@{}}+10.86\\[-5pt] \smalltext{{[}-4.84, +38.49{]}}\end{tabular}} &
  {\color[HTML]{D41159} \begin{tabular}[c]{@{}c@{}}+0.12\\[-5pt] \smalltext{{[}-7.59, +10.40{]}}\end{tabular}} &
  {\color[HTML]{59CA16} \begin{tabular}[c]{@{}c@{}}-23.42\\[-5pt] \smalltext{{[}-59.05, -2.98{]}}\end{tabular}} &
  {\color[HTML]{59CA16} \begin{tabular}[c]{@{}c@{}}+2.37\\[-5pt] \smalltext{{[}-1.33, +6.50{]}}\end{tabular}} &
  {\color[HTML]{59CA16} \begin{tabular}[c]{@{}c@{}}+2.79\\[-5pt] \smalltext{{[}-1.99, +9.12{]}}\end{tabular}} &
  {\color[HTML]{59CA16} \begin{tabular}[c]{@{}c@{}}+3.20\\[-5pt] \smalltext{{[}-0.87, +7.58{]}}\end{tabular}} &
  {\color[HTML]{59CA16} \begin{tabular}[c]{@{}c@{}}+7.33\\[-5pt] \smalltext{{[}+1.38, +15.27{]}}\end{tabular}} &
  {\color[HTML]{2B82D8} \begin{tabular}[c]{@{}c@{}}0.00\\[-5pt] \smalltext{{[}0.00, 0.00{]}}\\ 1 $\rightarrow$ 1 / 1\end{tabular}} \\
MSS3 &
  {\color[HTML]{D41159} \begin{tabular}[c]{@{}c@{}}-0.42\\[-5pt] \smalltext{{[}-1.49, +0.55{]}}\end{tabular}} &
  {\color[HTML]{59CA16} \begin{tabular}[c]{@{}c@{}}+4.66\\[-5pt] \smalltext{{[}-2.69, +13.08{]}}\end{tabular}} &
  {\color[HTML]{59CA16} \begin{tabular}[c]{@{}c@{}}+0.48\\[-5pt] \smalltext{{[}-0.23, +1.17{]}}\end{tabular}} &
  {\color[HTML]{59CA16} \begin{tabular}[c]{@{}c@{}}+9.67\\[-5pt] \smalltext{{[}+1.04, +21.32{]}}\end{tabular}} &
  {\color[HTML]{D41159} \begin{tabular}[c]{@{}c@{}}+0.83\\[-5pt] \smalltext{{[}-17.49, +21.09{]}}\end{tabular}} &
  {\color[HTML]{59CA16} \begin{tabular}[c]{@{}c@{}}-25.71\\[-5pt] \smalltext{{[}-49.18, +0.93{]}}\end{tabular}} &
  {\color[HTML]{D41159} \begin{tabular}[c]{@{}c@{}}+9.16\\[-5pt] \smalltext{{[}-0.46, +20.92{]}}\end{tabular}} &
  {\color[HTML]{59CA16} \begin{tabular}[c]{@{}c@{}}-20.07\\[-5pt] \smalltext{{[}-48.29, +11.70{]}}\end{tabular}} &
  {\color[HTML]{59CA16} \begin{tabular}[c]{@{}c@{}}+0.04\\[-5pt] \smalltext{{[}-1.58, +1.64{]}}\end{tabular}} &
  {\color[HTML]{59CA16} \begin{tabular}[c]{@{}c@{}}+2.26\\[-5pt] \smalltext{{[}-3.29, +8.46{]}}\end{tabular}} &
  {\color[HTML]{59CA16} \textbf{\begin{tabular}[c]{@{}c@{}}+3.93*\\[-5pt] \smalltext{{[}+2.59, +5.30{]}}\end{tabular}}} &
  {\color[HTML]{59CA16} \begin{tabular}[c]{@{}c@{}}+4.41\\[-5pt] \smalltext{{[}-4.14, +15.69{]}}\end{tabular}} &
  {\color[HTML]{D41159} \begin{tabular}[c]{@{}c@{}}100.00\\[-5pt] \smalltext{{[}0.00, inf{]}}\\ 1 $\rightarrow$ 2 / 10\end{tabular}} \\
LBC1936 &
  {\color[HTML]{D41159} \begin{tabular}[c]{@{}c@{}}-1.91*\\[-5pt] \smalltext{{[}-2.44, -1.39{]}}\end{tabular}} &
  {\color[HTML]{D41159} \begin{tabular}[c]{@{}c@{}}-2.08\\[-5pt] \smalltext{{[}-13.49, 11.61{]}}\end{tabular}} &
  {\color[HTML]{59CA16} \begin{tabular}[c]{@{}c@{}}+0.04\\[-5pt] \smalltext{{[}-0.43, +0.51{]}}\end{tabular}} &
  {\color[HTML]{D41159} \begin{tabular}[c]{@{}c@{}}-12.96\\[-5pt] \smalltext{{[}-29.46, +2.86{]}}\end{tabular}} &
  {\color[HTML]{59CA16} \begin{tabular}[c]{@{}c@{}}-0.77\\[-5pt] \smalltext{{[}-4.53, +3.06{]}}\end{tabular}} &
  {\color[HTML]{D41159} \begin{tabular}[c]{@{}c@{}}+3.90\\[-5pt] \smalltext{{[}-13.30, +36.51{]}}\end{tabular}} &
  {\color[HTML]{D41159} \begin{tabular}[c]{@{}c@{}}+4.79\\[-5pt] \smalltext{{[}+1.34, +8.79{]}}\end{tabular}} &
  {\color[HTML]{59CA16} \begin{tabular}[c]{@{}c@{}}-5.17\\[-5pt] \smalltext{{[}-61.57, +35.78{]}}\end{tabular}} &
  {\color[HTML]{D41159} \begin{tabular}[c]{@{}c@{}}-9.74*\\[-5pt] \smalltext{{[}-11.25, -8.22{]}}\end{tabular}} &
  {\color[HTML]{D41159} \begin{tabular}[c]{@{}c@{}}-0.75\\[-5pt] \smalltext{{[}-4.26, +2.82{]}}\end{tabular}} &
  {\color[HTML]{59CA16} \begin{tabular}[c]{@{}c@{}}+16.04*\\[-5pt] \smalltext{{[}+13.49, +18.58{]}}\end{tabular}} &
  {\color[HTML]{D41159} \begin{tabular}[c]{@{}c@{}}-20.33\\[-5pt] \smalltext{{[}-41.81, +0.45{]}}\end{tabular}} &
  {\color[HTML]{59CA16} \begin{tabular}[c]{@{}c@{}}-25.00\\[-5pt] \smalltext{{[}-55.56, +16.67{]}}\\ 16 $\rightarrow$ 12 / 296\end{tabular}} \\
LBC1921 &
  {\color[HTML]{59CA16} \begin{tabular}[c]{@{}c@{}}+3.05*\\[-5pt] \smalltext{{[}+2.20, +4.09{]}}\end{tabular}} &
  - &
  {\color[HTML]{59CA16} \begin{tabular}[c]{@{}c@{}}+3.45*\\[-5pt] \smalltext{{[}+2.51, +4.49{]}}\end{tabular}} &
  - &
  {\color[HTML]{59CA16} \begin{tabular}[c]{@{}c@{}}-49.36*\\[-5pt] \smalltext{{[}-61.63, -36.32{]}}\end{tabular}} &
  - &
  {\color[HTML]{59CA16} \begin{tabular}[c]{@{}c@{}}-4.66\\[-5pt] \smalltext{{[}-10.12, +0.96{]}}\end{tabular}} &
  - &
  {\color[HTML]{59CA16} \begin{tabular}[c]{@{}c@{}}+12.02\\[-5pt] \smalltext{{[}+4.88, +19.93{]}}\end{tabular}} &
  - &
  {\color[HTML]{59CA16} \begin{tabular}[c]{@{}c@{}}+0.21\\[-5pt] \smalltext{{[}-4.04, +4.81{]}}\end{tabular}} &
  - &
  - \\
WMH-ch &
  {\color[HTML]{59CA16} \begin{tabular}[c]{@{}c@{}}+13.10*\\[-5pt] \smalltext{{[}+10.65, +15.84{]}}\end{tabular}} &
  - &
  {\color[HTML]{59CA16} \begin{tabular}[c]{@{}c@{}}+20.72*\\[-5pt] \smalltext{{[}+17.64, +24.28{]}}\end{tabular}} &
  - &
  {\color[HTML]{59CA16} \begin{tabular}[c]{@{}c@{}}-73.62*\\[-5pt] \smalltext{{[}-82.60, -61.20{]}}\end{tabular}} &
  - &
  {\color[HTML]{59CA16} \begin{tabular}[c]{@{}c@{}}-61.67*\\[-5pt] \smalltext{{[}-71.17, -51.03{]}}\end{tabular}} &
  - &
  {\color[HTML]{59CA16} \begin{tabular}[c]{@{}c@{}}+39.19*\\[-5pt] \smalltext{{[}+33.38, +45.47{]}}\end{tabular}} &
  - &
  {\color[HTML]{59CA16} \begin{tabular}[c]{@{}c@{}}+10.63*\\[-5pt] \smalltext{{[}+7.23, +14.17{]}}\end{tabular}} &
  - &
  - \\
BRATS &
  - &
  - &
  - &
  - &
  - &
  {\color[HTML]{59CA16} \begin{tabular}[c]{@{}c@{}}-89.59*\\[-5pt] \smalltext{{[}-100.00, -75.16{]}}\end{tabular}} &
  - &
  - &
  - &
  - &
  - &
  - &
  {\color[HTML]{59CA16} \begin{tabular}[c]{@{}c@{}}-82.35*\\[-5pt] \smalltext{{[}-100.00, -64.71{]}}\\ 17 $\rightarrow$ 3 / 17\end{tabular}} \\
ISLES &
  - &
  {\color[HTML]{59CA16} \begin{tabular}[c]{@{}c@{}}+27.00*\\[-5pt] \smalltext{{[}+14.60, +43.32{]}}\end{tabular}} &
  - &
  {\color[HTML]{59CA16} \begin{tabular}[c]{@{}c@{}}+36.28*\\[-5pt] \smalltext{{[}+17.67, +64.83{]}}\end{tabular}} &
  - &
  {\color[HTML]{59CA16} \begin{tabular}[c]{@{}c@{}}-45.08\\[-5pt] \smalltext{{[}-63.66, -13.77{]}}\end{tabular}} &
  - &
  {\color[HTML]{59CA16} \begin{tabular}[c]{@{}c@{}}-38.76*\\[-5pt] \smalltext{{[}-60.32, -17.28{]}}\end{tabular}} &
  - &
  {\color[HTML]{59CA16} \begin{tabular}[c]{@{}c@{}}+13.89*\\[-5pt] \smalltext{{[}+5.92, +24.27{]}}\end{tabular}} &
  - &
  {\color[HTML]{59CA16} \begin{tabular}[c]{@{}c@{}}+36.88*\\[-5pt] \smalltext{{[}+13.78, +71.57{]}}\end{tabular}} &
  {\color[HTML]{2B82D8} \begin{tabular}[c]{@{}c@{}}0.00\\[-5pt] \smalltext{{[}0.00, 0.00{]}}\\ 0 $\rightarrow$ 0 / 1\end{tabular}} \\
SOOP &
  - &
  {\color[HTML]{59CA16} \begin{tabular}[c]{@{}c@{}}+40.68*\\[-5pt] \smalltext{{[}+31.56, +51.29{]}}\end{tabular}} &
  - &
  {\color[HTML]{59CA16} \begin{tabular}[c]{@{}c@{}}+67.92*\\[-5pt] \smalltext{{[}+49.77, +90.76{]}}\end{tabular}} &
  - &
  {\color[HTML]{59CA16} \begin{tabular}[c]{@{}c@{}}-73.11*\\[-5pt] \smalltext{{[}-81.48, -60.86{]}}\end{tabular}} &
  - &
  {\color[HTML]{59CA16} \begin{tabular}[c]{@{}c@{}}-49.01*\\[-5pt] \smalltext{{[}-61.73, -34.77{]}}\end{tabular}} &
  - &
  {\color[HTML]{59CA16} \begin{tabular}[c]{@{}c@{}}+15.67*\\[-5pt] \smalltext{{[}+8.13, +24.10{]}}\end{tabular}} &
  - &
  {\color[HTML]{59CA16} \begin{tabular}[c]{@{}c@{}}+18.73*\\[-5pt] \smalltext{{[}+8.31, +31.05{]}}\end{tabular}} &
  {\color[HTML]{59CA16} \begin{tabular}[c]{@{}c@{}}-50.00\\[-5pt] \smalltext{{[}-100.00, 0.00{]}}\\ 2 $\rightarrow$ 1 / 3\end{tabular}} \\
WSS &
  - &
  {\color[HTML]{59CA16} \begin{tabular}[c]{@{}c@{}}+5.17\\[-5pt] \smalltext{{[}-11.15, +28.09{]}}\end{tabular}} &
  - &
  {\color[HTML]{59CA16} \begin{tabular}[c]{@{}c@{}}+14.28\\[-5pt] \smalltext{{[}-35.62, +129.85{]}}\end{tabular}} &
  - &
  {\color[HTML]{59CA16} \begin{tabular}[c]{@{}c@{}}-44.40\\[-5pt] \smalltext{{[}-63.02, +24.06{]}}\end{tabular}} &
  - &
  {\color[HTML]{59CA16} \begin{tabular}[c]{@{}c@{}}-77.66\\[-5pt] \smalltext{{[}-87.29, +10.62{]}}\end{tabular}} &
  - &
  {\color[HTML]{59CA16} \begin{tabular}[c]{@{}c@{}}+4.88\\[-5pt] \smalltext{{[}-5.46, +27.31{]}}\end{tabular}} &
  - &
  {\color[HTML]{59CA16} \begin{tabular}[c]{@{}c@{}}+61.60\\[-5pt] \smalltext{{[}-24.05, +558.76{]}}\end{tabular}} &
  {\color[HTML]{2B82D8} \begin{tabular}[c]{@{}c@{}}0.00\\[-5pt] \smalltext{{[}0.00, 0.00{]}}\\ 0 $\rightarrow$ 0 / 1\end{tabular}} \\
ESS &
  {\color[HTML]{D41159} \begin{tabular}[c]{@{}c@{}}-0.08\\[-5pt] \smalltext{{[}-1.16, +1.06{]}}\end{tabular}} &
  {\color[HTML]{59CA16} \begin{tabular}[c]{@{}c@{}}+17.66*\\[-5pt] \smalltext{{[}+8.77, +29.52{]}}{]}\end{tabular}} &
  {\color[HTML]{D41159} \begin{tabular}[c]{@{}c@{}}-0.97\\[-5pt] \smalltext{{[}-1.98, +0.16{]}}\end{tabular}} &
  {\color[HTML]{59CA16} \begin{tabular}[c]{@{}c@{}}+8.99\\[-5pt] \smalltext{{[}-2.41, +23.66{]}}\end{tabular}} &
  {\color[HTML]{D41159} \begin{tabular}[c]{@{}c@{}}+10.02\\[-5pt] \smalltext{{[}+2.20, +19.24{]}}\end{tabular}} &
  {\color[HTML]{59CA16} \begin{tabular}[c]{@{}c@{}}-0.68\\[-5pt] \smalltext{{[}-12.67, +12.54{]}}\end{tabular}} &
  {\color[HTML]{D41159} \begin{tabular}[c]{@{}c@{}}+9.26\\[-5pt] \smalltext{{[}+2.77, +15.75{]}}\end{tabular}} &
  {\color[HTML]{59CA16} \begin{tabular}[c]{@{}c@{}}-24.18\\[-5pt] \smalltext{{[}-40.67, -7.05{]}}\end{tabular}} &
  {\color[HTML]{D41159} \begin{tabular}[c]{@{}c@{}}-4.02\\[-5pt] \smalltext{{[}-7.40, -0.54{]}}\end{tabular}} &
  {\color[HTML]{59CA16} \begin{tabular}[c]{@{}c@{}}+4.18\\[-5pt] \smalltext{{[}-1.24, +11.53{]}}\end{tabular}} &
  {\color[HTML]{D41159} \begin{tabular}[c]{@{}c@{}}-6.98*\\[-5pt] \smalltext{{[}-10.27, -3.64{]}}\end{tabular}} &
  {\color[HTML]{59CA16} \begin{tabular}[c]{@{}c@{}}+7.80\\[-5pt] \smalltext{{[}-4.68, +22.99{]}}\end{tabular}} &
  - \\
LINCHPIN &
  - &
  - &
  - &
  - &
  - &
  {\color[HTML]{009901} \begin{tabular}[c]{@{}c@{}}-27.78*\\[-5pt] \smalltext{{[}-28.53, -21.54{]}}\end{tabular}} &
  - &
  - &
  - &
  - &
  - &
  - &
  {\color[HTML]{3166FF} \begin{tabular}[c]{@{}c@{}}0.00\\[-5pt] \smalltext{{[}0.00, 0.00{]}}\\ 2 $\rightarrow$ 2 / 2\end{tabular}} \\ \hline
\end{tabular}%
}
\end{table}

A detailed stratification of test performance across individual datasets is presented in \textbf{Table \ref{tab:extra_data_2}}. The most pronounced improvements are observed within the partially labelled test datasets (LBC1921, WMH-ch, BRATS, ISLES, SOOP, WSS, LINCHPIN). This is an expected outcome, as these datasets are absent from the \acrshort{fls} but are incorporated within the \acrshort{pls} during training. This underscores the primary advantage of expanding the training corpus: it reduces covariate shift and the probability of encountering out-of-distribution samples during inference.

Conversely, on test datasets originating from the \acrshort{fls} (MSS1, MSS2, MSS3, LBC1936), performance exhibits greater variability, demonstrating both metric- and pathology-dependent improvements and degradations. Rather than interpreting this strictly as a model deficit, this behaviour highlights the inherent complexities of heterogeneous data integration. As detailed in subsequent sections, ground truth annotation protocols vary significantly across datasets. Consequently, a model trained on a composite dataset must infer a consensus segmentation policy, which is implicitly weighted by the proportional representation of each dataset's annotation standard within the overall training pool. Therefore, the observed metric degradations on \acrshort{fls} datasets, particularly for \acrshort{wmh}, likely result from the model's consensus policy diverging from the annotation conventions of those specific datasets, whose relative influence is diluted once the \acrshort{pls} data is introduced. This hypothesis is supported by the qualitative analysis presented in Section \ref{sec: qualitative}.

Finally, the results on the BRATS dataset demonstrate the utility of incorporating labelled negative cases containing alternative pathologies. The \acrshort{fls}-trained model initially generated false-positive \acrshort{isl} predictions across all 17 BRATS test scans; however, integrating the \acrshort{pls} training data reduced these false-positive occurrences to only 3 cases, significantly improving the model's robustness to confounding pathologies.

\subsection{Comparison of training strategies}

We evaluated the performance of models trained under various supervision paradigms. Class-specific and macro-averaged results are detailed in \textbf{Table \ref{tab:results_classes}} and \textbf{Table \ref{tab:results_avg}}, respectively.

% Main results per class
\begin{table}[htbp]
\caption{Results of the different training approaches for utilising the partially labelled subset (PLS) of data, compared to a multiclass baselines, trained on the fully labelled subset of the data (FLS). Reported as mean {[}95\% CI{]} with confidence intervals (CIs) calculated over a 10,000 sample bootstrap. Asterisk (*) and dagger (\textdagger) show statistically significant ($p$ \textless 0.05, two sided) improvement and deterioration respectively, in comparison to the multiclass model, according to a paired bootstrap test. ASD can only be calculated for cases where there is a least one predicted voxel, so these results are reported as mean [95\% CI] \{N\} where N is the number of test samples for which the metric was calculable, if N is less than the number of non-empty masks in the test set. Best result per column in bold. AP: average precision, DSC: Dice similarity coefficient, AVD: absolute volume difference, ASD: average surface distance, LPRE: lesion-level precision, LREC: lesion-level recall, FP: percentage of empty ground truth masks containing at least one false-positive prediction, WMH: white matter hyperintensities, ISL: ischaemic stroke lesions.}
\label{tab:results_classes}
\resizebox{\textwidth}{!}{%
\begin{tabular}{l|cc|cc|cc|cc|cc|cc|c}
\hline
Method &
  \multicolumn{2}{c|}{AP (\%)} &
  \multicolumn{2}{c|}{DSC (\%)} &
  \multicolumn{2}{c|}{AVD (\% of ICV)} &
  \multicolumn{2}{c|}{ASD (mm)} &
  \multicolumn{2}{c|}{LPRE (\%)} &
  \multicolumn{2}{c|}{LREC (\%)} &
  FP (\%) \\ \cline{2-14} 
 &
  WMH &
  ISL &
  WMH &
  ISL &
  WMH &
  ISL &
  WMH &
  ISL &
  WMH &
  ISL &
  WMH &
  ISL &
  ISL \\ \hline
multiclass &
  \begin{tabular}[c]{@{}c@{}}75.62\\[-5pt] \smalltext{{[}74.18, 77.05{]}}\end{tabular} &
  \begin{tabular}[c]{@{}c@{}}48.05\\[-5pt] \smalltext{{[}45.21, 50.87{]}}\end{tabular} &
  \begin{tabular}[c]{@{}c@{}}65.91\\[-5pt] \smalltext{{[}64.63, 67.19{]}}\end{tabular} &
  \begin{tabular}[c]{@{}c@{}}36.50\\[-5pt] \smalltext{{[}33.82, 39.15{]}}\end{tabular} &
  \begin{tabular}[c]{@{}c@{}}0.2526\\[-5pt] \smalltext{{[}0.2298, 0.2776{]}}\end{tabular} &
  \begin{tabular}[c]{@{}c@{}}0.6201\\[-5pt] \smalltext{{[}0.4691, 0.7927{]}}\end{tabular} &
  \begin{tabular}[c]{@{}c@{}}1.67\\[-5pt] \smalltext{{[}1.36, 2.06{]}}\end{tabular} &
  \begin{tabular}[c]{@{}c@{}}11.28 \{446\}\\[-5pt] \smalltext{{[}9.52, 13.03{]}}\end{tabular} &
  \begin{tabular}[c]{@{}c@{}}66.92\\[-5pt] \smalltext{{[}65.42, 68.44{]}}\end{tabular} &
  \begin{tabular}[c]{@{}c@{}}75.45\\[-5pt] \smalltext{{[}72.65, 78.27{]}}\end{tabular} &
  \begin{tabular}[c]{@{}c@{}}38.98\\[-5pt] \smalltext{{[}37.47, 40.48{]}}\end{tabular} &
  \begin{tabular}[c]{@{}c@{}}49.58\\[-5pt] \smalltext{{[}40.20, 52.95{]}}\end{tabular} &
  \begin{tabular}[c]{@{}c@{}}12.42\\[-5pt] \smalltext{{[}8.89, 16.20{]}}\end{tabular} \\ \hline
multi-model &
  \begin{tabular}[c]{@{}c@{}}75.67*\\[-5pt] \smalltext{{[}74.20, 77.13{]}}\end{tabular} &
  \begin{tabular}[c]{@{}c@{}}52.11*\\[-5pt] \smalltext{{[}49.23, 54.95{]}}\end{tabular} &
  \begin{tabular}[c]{@{}c@{}}67.88*\\[-5pt] \smalltext{{[}66.57, 69.19{]}}\end{tabular} &
  \begin{tabular}[c]{@{}c@{}}42.73*\\[-5pt] \smalltext{{[}39.95, 45.45{]}}\end{tabular} &
  \textbf{\begin{tabular}[c]{@{}c@{}}0.2076*\\[-5pt] \smalltext{{[}0.1856, 0.2320{]}}\end{tabular}} &
  \begin{tabular}[c]{@{}c@{}}0.2269*\\[-5pt] \smalltext{{[}0.1814, 0.2789{]}}\end{tabular} &
  \begin{tabular}[c]{@{}c@{}}1.61\\[-5pt] \smalltext{{[}1.28, 2.01{]}}\end{tabular} &
  \begin{tabular}[c]{@{}c@{}}7.46* \{447\}\\[-5pt] \smalltext{{[}6.06, 8.95{]}}\end{tabular} &
  \begin{tabular}[c]{@{}c@{}}64.85$^{\dagger}$\\[-5pt] \smalltext{{[}63.23, 66.50{]}}\end{tabular} &
  \begin{tabular}[c]{@{}c@{}}81.68*\\[-5pt] \smalltext{{[}79.13, 84.16{]}}\end{tabular} &
  \begin{tabular}[c]{@{}c@{}}42.66*\\[-5pt] \smalltext{{[}41.16, 44.13{]}}\end{tabular} &
  \begin{tabular}[c]{@{}c@{}}52.18\\[-5pt] \smalltext{{[}48.73, 55.63{]}}\end{tabular} &
  \textbf{\begin{tabular}[c]{@{}c@{}}6.52*\\[-5pt] \smalltext{{[}3.95, 9.40{]}}\end{tabular}} \\ \hline
multi-model (TS) &
  \begin{tabular}[c]{@{}c@{}}75.87*\\[-5pt] \smalltext{{[}74.42, 77.30{]}}\end{tabular} &
  \begin{tabular}[c]{@{}c@{}}52.73*\\[-5pt] \smalltext{{[}49.82, 55.89{]}}\end{tabular} &
  \textbf{\begin{tabular}[c]{@{}c@{}}67.91*\\[-5pt] \smalltext{{[}66.60, 69.22{]}}\end{tabular}} &
  \begin{tabular}[c]{@{}c@{}}42.38*\\[-5pt] \smalltext{{[}39.61, 45.12{]}}\end{tabular} &
  \begin{tabular}[c]{@{}c@{}}0.2077*\\[-5pt] \smalltext{{[}0.1857, 0.2321{]}}\end{tabular} &
  \begin{tabular}[c]{@{}c@{}}0.2306*\\[-5pt] \smalltext{{[}0.1841, 0.2840{]}}\end{tabular} &
  \begin{tabular}[c]{@{}c@{}}1.60\\[-5pt] \smalltext{{[}1.27, 2.01{]}}\end{tabular} &
  \begin{tabular}[c]{@{}c@{}}7.47* \{446\}\\[-5pt] \smalltext{{[}6.07, 8.96{]}}\end{tabular} &
  \begin{tabular}[c]{@{}c@{}}65.00$^{\dagger}$\\[-5pt] \smalltext{{[}63.35, 66.66{]}}\end{tabular} &
  \begin{tabular}[c]{@{}c@{}}81.83*\\[-5pt] \smalltext{{[}79.28, 84.34{]}}\end{tabular} &
  \textbf{\begin{tabular}[c]{@{}c@{}}42.70*\\[-5pt] \smalltext{{[}41.19, 44.17{]}}\end{tabular}} &
  \begin{tabular}[c]{@{}c@{}}51.48\\[-5pt] \smalltext{{[}48.01, 54.89{]}}\end{tabular} &
  \textbf{\begin{tabular}[c]{@{}c@{}}6.52*\\[-5pt] \smalltext{{[}3.95, 9.40{]}}\end{tabular}} \\ \hline
class-conditional &
  \begin{tabular}[c]{@{}c@{}}69.06$^{\dagger}$\\[-5pt] \smalltext{{[}67.51, 70.62{]}}\end{tabular} &
  \begin{tabular}[c]{@{}c@{}}49.93\\[-5pt] \smalltext{{[}47.22, 52.64{]}}\end{tabular} &
  \begin{tabular}[c]{@{}c@{}}61.42$^{\dagger}$\\[-5pt] \smalltext{{[}60.08, 62.76{]}}\end{tabular} &
  \begin{tabular}[c]{@{}c@{}}44.21*\\[-5pt] \smalltext{{[}41.43, 46.93{]}}\end{tabular} &
  \begin{tabular}[c]{@{}c@{}}0.2605\\[-5pt] \smalltext{{[}0.2362, 0.2858{]}}\end{tabular} &
  \begin{tabular}[c]{@{}c@{}}0.2697*\\[-5pt] \smalltext{{[}0.2192, 32.77{]}}\end{tabular} &
  \begin{tabular}[c]{@{}c@{}}1.80\\[-5pt] \smalltext{{[}1.48, 2.20{]}}\end{tabular} &
  \begin{tabular}[c]{@{}c@{}}9.23* \{487\}\\[-5pt] \smalltext{{[}7.75, 10.80{]}}\end{tabular} &
  \textbf{\begin{tabular}[c]{@{}c@{}}76.54*\\[-5pt] \smalltext{{[}75.01, 78.05{]}}\end{tabular}} &
  \begin{tabular}[c]{@{}c@{}}71.33\\[-5pt] \smalltext{{[}68.38, 74.30{]}}\end{tabular} &
  \begin{tabular}[c]{@{}c@{}}29.64$^{\dagger}$\\[-5pt] \smalltext{{[}28.19, 31.06{]}}\end{tabular} &
  \textbf{\begin{tabular}[c]{@{}c@{}}58.36*\\[-5pt] \smalltext{{[}54.98, 61.71{]}}\end{tabular}} &
  \begin{tabular}[c]{@{}c@{}}22.67$^{\dagger}$\\[-5pt] \smalltext{{[}18.12, 27.44{]}}\end{tabular} \\ \hline
pseudolabels &
  \begin{tabular}[c]{@{}c@{}}76.00\\[-5pt] \smalltext{{[}74.55, 77.42{]}}\end{tabular} &
  \textbf{\begin{tabular}[c]{@{}c@{}}55.17*\\[-5pt] \smalltext{{[}52.30, 57.97{]}}\end{tabular}} &
  \begin{tabular}[c]{@{}c@{}}67.22*\\[-5pt] \smalltext{{[}65.93, 68.51{]}}\end{tabular} &
  \textbf{\begin{tabular}[c]{@{}c@{}}46.95*\\[-5pt] \smalltext{{[}44.24, 49.58{]}}\end{tabular}} &
  \begin{tabular}[c]{@{}c@{}}0.2151*\\[-5pt] \smalltext{{[}0.1928, 0.2389{]}}\end{tabular} &
  \textbf{\begin{tabular}[c]{@{}c@{}}0.2115*\\[-5pt] \smalltext{{[}0.16.86, 0.2602{]}}\end{tabular}} &
  \textbf{\begin{tabular}[c]{@{}c@{}}1.54\\[-5pt] \smalltext{{[}1.20, 2.00{]}}\end{tabular}} &
  \begin{tabular}[c]{@{}c@{}}9.96 \{503\}\\[-5pt] \smalltext{{[}8.43, 11.55{]}}\end{tabular} &
  \begin{tabular}[c]{@{}c@{}}71.16*\\[-5pt] \smalltext{{[}69.65, 72.68{]}}\end{tabular} &
  \begin{tabular}[c]{@{}c@{}}70.60$^{\dagger}$\\[-5pt] \smalltext{{[}67.62, 73.52{]}}\end{tabular} &
  \begin{tabular}[c]{@{}c@{}}37.76$^{\dagger}$\\[-5pt] \smalltext{{[}36.25, 39.24{]}}\end{tabular} &
  \begin{tabular}[c]{@{}c@{}}57.82*\\[-5pt] \smalltext{{[}54.38, 61.08{]}}\end{tabular} &
  \begin{tabular}[c]{@{}c@{}}21.12$^{\dagger}$\\[-5pt] \smalltext{{[}16.72, 25.71{]}}\end{tabular} \\ \hline
phased &
  \begin{tabular}[c]{@{}c@{}}75.05$^{\dagger}$\\[-5pt] \smalltext{{[}73.58, 76.51{]}}\end{tabular} &
  \begin{tabular}[c]{@{}c@{}}51.10*\\[-5pt] \smalltext{{[}48.26, 53.90{]}}\end{tabular} &
  \begin{tabular}[c]{@{}c@{}}66.71*\\[-5pt] \smalltext{{[}65.42, 68.02{]}}\end{tabular} &
  \begin{tabular}[c]{@{}c@{}}44.48*\\[-5pt] \smalltext{{[}41.76, 47.11{]}}\end{tabular} &
  \begin{tabular}[c]{@{}c@{}}0.2234*\\[-5pt] \smalltext{{[}0.2004, 0.2482{]}}\end{tabular} &
  \begin{tabular}[c]{@{}c@{}}0.3748*\\[-5pt] \smalltext{{[}0.2818, 48.22{]}}\end{tabular} &
  \begin{tabular}[c]{@{}c@{}}1.56\\[-5pt] \smalltext{{[}1.22, 1.98{]}}\end{tabular} &
  \begin{tabular}[c]{@{}c@{}}12.03 \{510\}\\[-5pt] \smalltext{{[}10.35, 13.75{]}}\end{tabular} &
  \begin{tabular}[c]{@{}c@{}}69.63*\\[-5pt] \smalltext{{[}68.10, 71.14{]}}\end{tabular} &
  \begin{tabular}[c]{@{}c@{}}68.00$^{\dagger}$\\[-5pt] \smalltext{{[}64.91, 71.03{]}}\end{tabular} &
  \begin{tabular}[c]{@{}c@{}}37.55$^{\dagger}$\\[-5pt] \smalltext{{[}36.16, 38.94{]}}\end{tabular} &
  \begin{tabular}[c]{@{}c@{}}57.07*\\[-5pt] \smalltext{{[}53.66, 60.36{]}}\end{tabular} &
  \begin{tabular}[c]{@{}c@{}}19.88$^{\dagger}$\\[-5pt] \smalltext{{[}15.58, 24.39{]}}\end{tabular} \\ \hline
class-adaptive loss &
  \begin{tabular}[c]{@{}c@{}}76.62*\\[-5pt] \smalltext{{[}75.18, 78.05{]}}\end{tabular} &
  \begin{tabular}[c]{@{}c@{}}53.64*\\[-5pt] \smalltext{{[}50.75, 56.49{]}}\end{tabular} &
  \begin{tabular}[c]{@{}c@{}}67.71*\\[-5pt] \smalltext{{[}66.41, 69.00{]}}\end{tabular} &
  \begin{tabular}[c]{@{}c@{}}43.82*\\[-5pt] \smalltext{{[}41.06, 46.56{]}}\end{tabular} &
  \begin{tabular}[c]{@{}c@{}}0.2136*\\[-5pt] \smalltext{{[}0.1911, 0.2381{]}}\end{tabular} &
  \begin{tabular}[c]{@{}c@{}}0.3230*\\[-5pt] \smalltext{{[}0.2571, 0.3962{]}}\end{tabular} &
  \begin{tabular}[c]{@{}c@{}}1.56\\[-5pt] \smalltext{{[}1.24, 1.97{]}}\end{tabular} &
  \textbf{\begin{tabular}[c]{@{}c@{}}6.90* \{454\}\\[-5pt] \smalltext{{[}5.61, 8.24{]}}\end{tabular}} &
  \begin{tabular}[c]{@{}c@{}}68.01*\\[-5pt] \smalltext{{[}66.45, 69.57{]}}\end{tabular} &
  \begin{tabular}[c]{@{}c@{}}81.38*\\[-5pt] \smalltext{{[}78.84, 83.88{]}}\end{tabular} &
  \begin{tabular}[c]{@{}c@{}}40.45*\\[-5pt] \smalltext{{[}38.97, 41.91{]}}\end{tabular} &
  \begin{tabular}[c]{@{}c@{}}51.53\\[-5pt] \smalltext{{[}48.14, 54.84{]}}\end{tabular} &
  \begin{tabular}[c]{@{}c@{}}9.60\\[-5pt] \smalltext{{[}6.54, 12.98{]}}\end{tabular} \\ \hline
marginal loss &
  \textbf{\begin{tabular}[c]{@{}c@{}}76.67*\\[-5pt] \smalltext{{[}75.24, 78.08{]}}\end{tabular}} &
  \begin{tabular}[c]{@{}c@{}}52.77*\\[-5pt] \smalltext{{[}49.76, 55.65{]}}\end{tabular} &
  \begin{tabular}[c]{@{}c@{}}67.37*\\[-5pt] \smalltext{{[}66.07, 68.66{]}}\end{tabular} &
  \begin{tabular}[c]{@{}c@{}}43.09*\\[-5pt] \smalltext{{[}40.31, 45.83{]}}\end{tabular} &
  \begin{tabular}[c]{@{}c@{}}0.2153*\\[-5pt] \smalltext{{[}0.1927, 0.2402{]}}\end{tabular} &
  \begin{tabular}[c]{@{}c@{}}0.2322*\\[-5pt] \smalltext{{[}0.1844, 0.2858{]}}\end{tabular} &
  \begin{tabular}[c]{@{}c@{}}1.57\\[-5pt] \smalltext{{[}1.23, 1.99{]}}\end{tabular} &
  \begin{tabular}[c]{@{}c@{}}7.88* \{451\}\\[-5pt] \smalltext{{[}6.45, 9.37{]}}\end{tabular} &
  \begin{tabular}[c]{@{}c@{}}71.18*\\[-5pt] \smalltext{{[}69.68, 72.68{]}}\end{tabular} &
  \textbf{\begin{tabular}[c]{@{}c@{}}84.40*\\[-5pt] \smalltext{{[}81.91, 86.95{]}}\end{tabular}} &
  \begin{tabular}[c]{@{}c@{}}37.90$^{\dagger}$\\[-5pt] \smalltext{{[}36.39, 39.38{]}}\end{tabular} &
  \begin{tabular}[c]{@{}c@{}}51.20\\[-5pt] \smalltext{{[}47.74, 54.59{]}}\end{tabular} &
  \begin{tabular}[c]{@{}c@{}}7.14*\\[-5pt] \smalltext{{[}4.44, 10.17{]}}\end{tabular} \\ \hline
\end{tabular}%
}
\end{table}

% Main results macro-avg
\begin{table}[htbp]
\caption{Macro average results of different training approaches for utilising the partially labelled subset (PLS) of data, compared to a multiclass baselines, trained on the fully labelled subset of the data (FLS). Reported as mean {[}95\% CI{]} with confidence intervals (CIs) calculated over a 10,000 sample bootstrap. Asterisk (*) and dagger (\textdagger) show statistically significant ($p$ \textless 0.05, two sided) improvement and deterioration respectively, in comparison to the multiclass model, according to a paired bootstrap test. Average for cases with only one ground truth class default to the value of the metric for that class. Best result per column in bold. AP: average precision, DSC: Dice similarity coefficient, AVD: absolute volume difference, ASD: average surface distance, LPRE: lesion-level precision, LREC: lesion-level recall.}
\label{tab:results_avg}
\resizebox{\textwidth}{!}{%
\begin{tabular}{l|cl|cl|cl|cl|cl|cl}
\hline
Method &
  \multicolumn{2}{c|}{AP (\%)} &
  \multicolumn{2}{c|}{DSC (\%)} &
  \multicolumn{2}{c|}{AVD (\% ICV)} &
  \multicolumn{2}{c|}{ASD (mm)} &
  \multicolumn{2}{c|}{LPRE (\%)} &
  \multicolumn{2}{c}{LREC (\%)} \\ \hline
multiclass &
  \multicolumn{2}{c|}{\begin{tabular}[c]{@{}c@{}}63.06\\[-5pt] \smalltext{{[}61.30, 64.75{]}}\end{tabular}} &
  \multicolumn{2}{c|}{\begin{tabular}[c]{@{}c@{}}52.15\\[-5pt] \smalltext{{[}50.49, 53.76{]}}\end{tabular}} &
  \multicolumn{2}{c|}{\begin{tabular}[c]{@{}c@{}}0.6401\\[-5pt] \smalltext{{[}0.5071, 0.7921{]}}\end{tabular}} &
  \multicolumn{2}{c|}{\begin{tabular}[c]{@{}c@{}}5.46\\[-5pt] \smalltext{{[}4.70, 6.25{]}}\end{tabular}} &
  \multicolumn{2}{c|}{\begin{tabular}[c]{@{}c@{}}69.56\\[-5pt] \smalltext{{[}67.93, 71.16{]}}\end{tabular}} &
  \multicolumn{2}{c}{\begin{tabular}[c]{@{}c@{}}44.14\\[-5pt] \smalltext{{[}42.26, 46.02{]}}\end{tabular}} \\ \hline
multi-model &
  \multicolumn{2}{c|}{\begin{tabular}[c]{@{}c@{}}65.75*\\[-5pt] \smalltext{{[}64.09, 67.40{]}}\end{tabular}} &
  \multicolumn{2}{c|}{\begin{tabular}[c]{@{}c@{}}57.40*\\[-5pt] \smalltext{{[}55.82, 58.97{]}}\end{tabular}} &
  \multicolumn{2}{c|}{\begin{tabular}[c]{@{}c@{}}0.2586*\\[-5pt] \smalltext{{[}0.2187, 0.3046{]}}\end{tabular}} &
  \multicolumn{2}{c|}{\begin{tabular}[c]{@{}c@{}}3.85*\\[-5pt] \smalltext{{[}3.23, 4.51{]}}\end{tabular}} &
  \multicolumn{2}{c|}{\begin{tabular}[c]{@{}c@{}}71.57\\[-5pt] \smalltext{{[}69.97, 73.14{]}}\end{tabular}} &
  \multicolumn{2}{c}{\textbf{\begin{tabular}[c]{@{}c@{}}48.39*\\[-5pt] \smalltext{{[}46.47, 50.28{]}}\end{tabular}}} \\ \hline
multi-model (TS) &
  \multicolumn{2}{c|}{\begin{tabular}[c]{@{}c@{}}66.15*\\[-5pt] \smalltext{{[}64.49, 67.79{]}}\end{tabular}} &
  \multicolumn{2}{c|}{\begin{tabular}[c]{@{}c@{}}57.26*\\[-5pt] \smalltext{{[}55.67, 58.83{]}}\end{tabular}} &
  \multicolumn{2}{c|}{\begin{tabular}[c]{@{}c@{}}0.2616*\\[-5pt] \smalltext{{[}0.2206, 0.3088{]}}\end{tabular}} &
  \multicolumn{2}{c|}{\begin{tabular}[c]{@{}c@{}}3.84*\\[-5pt] \smalltext{{[}3.22, 4.50{]}}\end{tabular}} &
  \multicolumn{2}{c|}{\begin{tabular}[c]{@{}c@{}}71.62\\[-5pt] \smalltext{{[}70.03, 73.22{]}}\end{tabular}} &
  \multicolumn{2}{c}{\begin{tabular}[c]{@{}c@{}}48.09*\\[-5pt] \smalltext{{[}46.18, 49.97{]}}\end{tabular}} \\ \hline
class-conditional &
  \multicolumn{2}{c|}{\begin{tabular}[c]{@{}c@{}}61.32$^{\dagger}$\\[-5pt] \smalltext{{[}59.71, 62.92{]}}\end{tabular}} &
  \multicolumn{2}{c|}{\begin{tabular}[c]{@{}c@{}}54.75*\\[-5pt] \smalltext{{[}53.20, 56.28{]}}\end{tabular}} &
  \multicolumn{2}{c|}{\begin{tabular}[c]{@{}c@{}}0.3160*\\[-5pt] \smalltext{{[}0.2708, 0.3677{]}}\end{tabular}} &
  \multicolumn{2}{c|}{\begin{tabular}[c]{@{}c@{}}4.79\\[-5pt] \smalltext{{[}4.11, 5.51{]}}\end{tabular}} &
  \multicolumn{2}{c|}{\begin{tabular}[c]{@{}c@{}}73.74*\\[-5pt] \smalltext{{[}72.07, 75.37{]}}\end{tabular}} &
  \multicolumn{2}{c}{\begin{tabular}[c]{@{}c@{}}43.24\\[-5pt] \smalltext{{[}41.25, 45.24{]}}\end{tabular}} \\ \hline
pseudolabels &
  \multicolumn{2}{c|}{\textbf{\begin{tabular}[c]{@{}c@{}}67.09*\\[-5pt] \smalltext{{[}65.46, 68.72{]}}\end{tabular}}} &
  \multicolumn{2}{c|}{\textbf{\begin{tabular}[c]{@{}c@{}}58.54*\\[-5pt] \smalltext{{[}57.02, 60.05{]}}\end{tabular}}} &
  \multicolumn{2}{c|}{\textbf{\begin{tabular}[c]{@{}c@{}}0.2518*\\[-5pt] \smalltext{{[}0.2147, 0.2950{]}}\end{tabular}}} &
  \multicolumn{2}{c|}{\begin{tabular}[c]{@{}c@{}}4.97\\[-5pt] \smalltext{{[}4.25, 5.72{]}}\end{tabular}} &
  \multicolumn{2}{c|}{\begin{tabular}[c]{@{}c@{}}70.19\\[-5pt] \smalltext{{[}68.51, 71.86{]}}\end{tabular}} &
  \multicolumn{2}{c}{\begin{tabular}[c]{@{}c@{}}47.36*\\[-5pt] \smalltext{{[}45.38, 49.30{]}}\end{tabular}} \\ \hline
phased &
  \multicolumn{2}{c|}{\begin{tabular}[c]{@{}c@{}}64.39*\\[-5pt] \smalltext{{[}62.68, 66.05{]}}\end{tabular}} &
  \multicolumn{2}{c|}{\begin{tabular}[c]{@{}c@{}}57.02*\\[-5pt] \smalltext{{[}55.46, 58.54{]}}\end{tabular}} &
  \multicolumn{2}{c|}{\begin{tabular}[c]{@{}c@{}}0.4016*\\[-5pt] \smalltext{{[}0.3192, 0.4970{]}}\end{tabular}} &
  \multicolumn{2}{c|}{\begin{tabular}[c]{@{}c@{}}5.88\\[-5pt] \smalltext{{[}5.08, 6.70{]}}\end{tabular}} &
  \multicolumn{2}{c|}{\begin{tabular}[c]{@{}c@{}}68.02\\[-5pt] \smalltext{{[}66.31, 69.72{]}}\end{tabular}} &
  \multicolumn{2}{c}{\begin{tabular}[c]{@{}c@{}}46.95*\\[-5pt] \smalltext{{[}45.06, 48.83{]}}\end{tabular}} \\ \hline
class-adaptive loss &
  \multicolumn{2}{c|}{\begin{tabular}[c]{@{}c@{}}66.82*\\[-5pt] \smalltext{{[}65.13, 68.47{]}}\end{tabular}} &
  \multicolumn{2}{c|}{\begin{tabular}[c]{@{}c@{}}57.51*\\[-5pt] \smalltext{{[}55.92, 59.09{]}}\end{tabular}} &
  \multicolumn{2}{c|}{\begin{tabular}[c]{@{}c@{}}0.3491*\\[-5pt] \smalltext{{[}0.2902, 0.4144{]}}\end{tabular}} &
  \multicolumn{2}{c|}{\textbf{\begin{tabular}[c]{@{}c@{}}3.64*\\[-5pt] \smalltext{{[}3.05, 4.26{]}}\end{tabular}}} &
  \multicolumn{2}{c|}{\begin{tabular}[c]{@{}c@{}}73.07*\\[-5pt] \smalltext{{[}71.43, 74.68{]}}\end{tabular}} &
  \multicolumn{2}{c}{\begin{tabular}[c]{@{}c@{}}46.65\\[-5pt] \smalltext{{[}44.75, 48.51{]}}\end{tabular}} \\ \hline
marginal loss &
  \multicolumn{2}{c|}{\begin{tabular}[c]{@{}c@{}}66.71*\\[-5pt] \smalltext{{[}65.00, 68.36{]}}\end{tabular}} &
  \multicolumn{2}{c|}{\begin{tabular}[c]{@{}c@{}}57.19*\\[-5pt] \smalltext{{[}55.61, 58.75{]}}\end{tabular}} &
  \multicolumn{2}{c|}{\begin{tabular}[c]{@{}c@{}}0.2684*\\[-5pt] \smalltext{{[}0.2265, 0.3169{]}}\end{tabular}} &
  \multicolumn{2}{c|}{\begin{tabular}[c]{@{}c@{}}4.04*\\[-5pt] \smalltext{{[}3.39, 4.72{]}}\end{tabular}} &
  \multicolumn{2}{c|}{\textbf{\begin{tabular}[c]{@{}c@{}}76.51*\\[-5pt] \smalltext{{[}74.98, 78.03{]}}\end{tabular}}} &
  \multicolumn{2}{c}{\begin{tabular}[c]{@{}c@{}}44.90\\[-5pt] \smalltext{{[}42.97, 46.84{]}}\end{tabular}} \\ \hline
\end{tabular}%
}
\end{table}

% Boxstrips for WMH and ISL
\begin{figure}
     \centering
     \begin{subfigure}[]{0.49\textwidth}
         \centering
         \includegraphics[width=\textwidth]{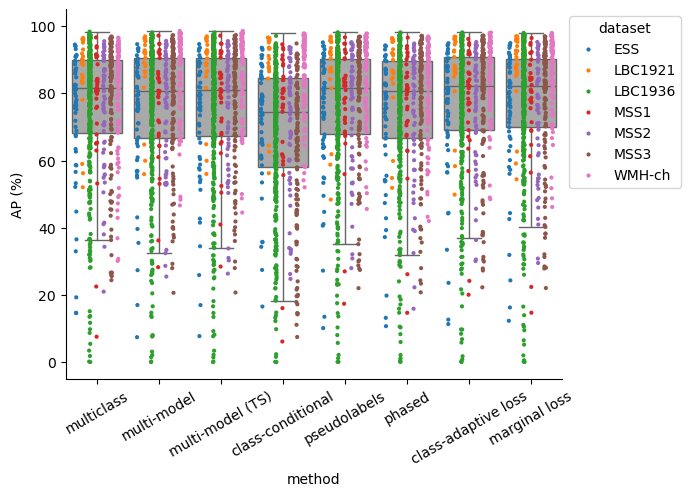}
         \caption{}
         \label{fig:AP boxstrip WMH}
     \end{subfigure}
     \hfill
     \begin{subfigure}[]{0.49\textwidth}
         \centering
         \includegraphics[width=\textwidth]{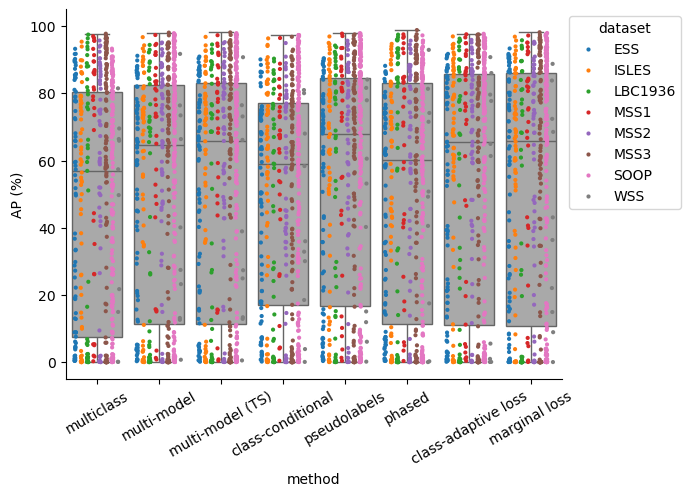}
         \caption{}
         \label{fig:AP boxstrip ISL}
     \end{subfigure}
        \caption{Average precision (AP) per dataset and per training strategy for: (a) white matter hyperintensities (WMH); and (b) ischaemic stroke lesions (ISL) Boxplot whiskers extend to 1.5 times the interquartile range.}
        \label{fig:box_strip}
\end{figure}

\paragraph{White matter hyperintensities (WMH)}
Regarding \acrshort{wmh} segmentation, the multi-model, class-adaptive loss, and marginal loss training strategies effectively leveraged the \acrshort{pls} data to yield statistically significant improvements over the multiclass baseline on the primary \acrshort{ap} metric, with the marginal loss approach achieving the most substantial gain. Conversely, the class-conditional and phased models exhibited a statistically significant degradation in \acrshort{ap}. Furthermore, the class-conditional model demonstrated the highest \acrshort{lpre} for \acrshort{wmh} alongside the highest \acrshort{lrec} for \acrshort{isl}, indicating a comparative bias toward predicting \acrshort{isl} in regions where alternative models might predict \acrshort{wmh}.

\paragraph{Ischaemic stroke lesions (ISL)}
For \acrshort{isl} segmentation, all approaches, with the exception of the class-conditional model, achieved statistically significant improvements in \acrshort{ap} relative to the multiclass baseline. The pseudolabels model achieved the highest performance on three of the four evaluated voxel-level metrics (\acrshort{ap}, \acrshort{dsc}, \acrshort{avd}, and \acrshort{asd}). Analysis of \acrshort{lpre} and \acrshort{lrec} for \acrshort{isl} reveals that the class-conditional, pseudolabels, and phased models adopt a less conservative predictive threshold for \acrshort{isl} compared to the alternative strategies. On the non-\acrshort{isl} cases, the multi-model configurations generated the fewest scan-level false-positive \acrshort{isl} predictions and, along with the marginal loss model, showed statistically significant improvement over the multiclass baseline on this metric; the class-conditional, pseudolabels, and phased models showed significant degradation.

\paragraph{Macro-average}
Macro-average scores for the operating-point-independent primary metric (\acrshort{ap}) and the \acrshort{dsc} metric---widely regarded as the standard measure of segmentation quality---indicate that the pseudolabels model achieves the highest overall performance. This superiority is primarily driven by its robust performance in \acrshort{isl} segmentation, a task characterised by greater inter-model variability and more substantial gains relative to the multiclass baseline. Given its superior performance, this model was selected for the subsequent qualitative analysis and clinical utility evaluation detailed in the following sections.

Furthermore, we conducted a comprehensive region-of-interest (ROI) analysis of the pseudolabel model's predictions. This included evaluating the segmentation of periventricular versus deep \acrshort{wmh}, cortical versus subcortical \acrshort{isl}, \acrshort{isl} distributions within specific arterial territories, and the volumetric dependency of \acrshort{wmh} and \acrshort{isl} segmentation accuracy. The novel and robust ROI-assignment methodology developed for this analysis, alongside the corresponding results, are detailed in \ref{appendix A}.

\paragraph{Per-dataset trends}
As illustrated by the distributions in \textbf{Figure \ref{fig:box_strip}}, performance metrics exhibited variations across the evaluated datasets. For \acrshort{wmh} segmentation, subjects from the LBC1921 cohort were consistently well-segmented across all evaluated models. Notably, on the ESS dataset---which was withheld during training to serve as an out-of-distribution (OOD) test set---segmentation performance for both \acrshort{wmh} and \acrshort{isl} remained highly comparable to that observed on in-distribution datasets. This provides encouraging evidence regarding the out-of-domain generalisation capabilities of the models. To supplement these findings, probabilistic false-positive and false-negative error maps stratified by dataset are provided in \ref{appendix B fp fn}.

\paragraph{Multi-model training: simplicity and robustness}
A key investigative focus of this study was to determine whether training two independent binary segmentation networks (the multi-model approach) would be inherently disadvantaged by the lack of explicit joint optimisation over mutually exclusive, visually confounding, and physiologically interrelated pathologies. Our results indicate that the decoupled nature of the multi-model framework did not introduce significant performance penalties. In fact, the multi-model configurations achieved the highest \acrshort{dsc} for \acrshort{wmh} segmentation, although they were outperformed by the marginal loss model on the primary \acrshort{ap} metric, which evaluates the precision-recall trade-off across the entire operational spectrum. 

Integrating temperature scaling using the validation set---denoted as the multi-model (TS) approach---yielded marginal improvements in \acrshort{ap} for both \acrshort{wmh} and \acrshort{isl}. This finding highlights the calibration challenges inherent to post-hoc combination of independent binary networks into a unified multi-class probability distribution, while demonstrating that temperature scaling serves as an effective mechanism to partially mitigate these limitations. Nevertheless, we urge caution when extrapolating this decoupled approach to alternative data distributions. The success of the multi-model framework in this setting may be partially attributable to specific characteristics of our training data distribution: namely, the inclusion of a substantially sized \acrshort{fls}, containing both classes, which facilitates implicit calibration while ensuring that each binary network is exposed to the opposing pathology class as a negative control during training.

\paragraph{Negative controls with tumours}
Leveraging the established ground truth tumour annotations within the BRATS dataset, we quantified the false-positive rates for both \acrshort{wmh} and \acrshort{isl} within these regions. These findings are summarised in \textbf{Table \ref{tab:results_brats}}.

%%% BRATS FP table %%%
\begin{table}[htbp]
\footnotesize
\centering
\caption{Number of times that one or more false-positive voxels were predicted within the tumour ground truth mask on the BRATS test set, out of a total of 17 subjects. Reported as mean {[}95\% CI{]} with confidence intervals (CIs) calculated over a 10,000 sample bootstrap. Asterisk (*) and dagger (\textdagger) show statistically significant ($p$ \textless 0.05, two sided) improvement and deterioration respectively, in comparison to the multiclass model, according to a paired bootstrap test. Best result per column in bold. WMH: white matter hyperintensities, ISL: ischaemic stroke lesions.}
\label{tab:results_brats}
\begin{tabular}{l|cc}
\hline
Method       & WMH                                                       & ISL                                                       \\ \hline
multiclass   & \begin{tabular}[c]{@{}c@{}}12\\[-5pt] \smalltext{{[}8, 15{]}}\end{tabular}  & \begin{tabular}[c]{@{}c@{}}17\\[-5pt] \smalltext{{[}17, 17{]}}\end{tabular} \\ \hline
multi-model         & \textbf{\begin{tabular}[c]{@{}c@{}}11\\[-5pt] \smalltext{{[}7, 15{]}}\end{tabular}} & \textbf{\begin{tabular}[c]{@{}c@{}}3*\\[-5pt] \smalltext{{[}0, 6{]}}\end{tabular}} \\ \hline
multi-model (TS)    & \textbf{\begin{tabular}[c]{@{}c@{}}11\\[-5pt] \smalltext{{[}7, 15{]}}\end{tabular}} & \textbf{\begin{tabular}[c]{@{}c@{}}3*\\[-5pt] \smalltext{{[}0, 6{]}}\end{tabular}} \\ \hline
class-conditional   & \begin{tabular}[c]{@{}c@{}}17\\[-5pt] \smalltext{{[}17, 17{]}}\end{tabular}         & \begin{tabular}[c]{@{}c@{}}17\\[-5pt] \smalltext{{[}17, 17{]}}\end{tabular}        \\ \hline
pseudolabels & \begin{tabular}[c]{@{}c@{}}17\\[-5pt] \smalltext{{[}17, 17{]}}\end{tabular} & \begin{tabular}[c]{@{}c@{}}13\\[-5pt] \smalltext{{[}9, 16{]}}\end{tabular}  \\ \hline
phased       & \begin{tabular}[c]{@{}c@{}}13\\[-5pt] \smalltext{{[}9, 16{]}}\end{tabular}  & \begin{tabular}[c]{@{}c@{}}17\\[-5pt] \smalltext{{[}17, 17{]}}\end{tabular} \\ \hline
class-adaptive loss & \textbf{\begin{tabular}[c]{@{}c@{}}11\\[-5pt] \smalltext{{[}7, 15{]}}\end{tabular}} & \begin{tabular}[c]{@{}c@{}}17\\[-5pt] \smalltext{{[}17, 17{]}}\end{tabular}        \\ \hline
marginal loss       & \begin{tabular}[c]{@{}c@{}}17\\[-5pt] \smalltext{{[}17, 17{]}}\end{tabular}         & \textbf{\begin{tabular}[c]{@{}c@{}}3*\\[-5pt] \smalltext{{[}0, 6{]}}\end{tabular}} \\ \hline
\end{tabular}
\normalsize
\end{table}

\begin{figure}[htbp]
     \centering
     \begin{subfigure}[]{0.49\textwidth}
         \centering
         \includegraphics[width=\textwidth]{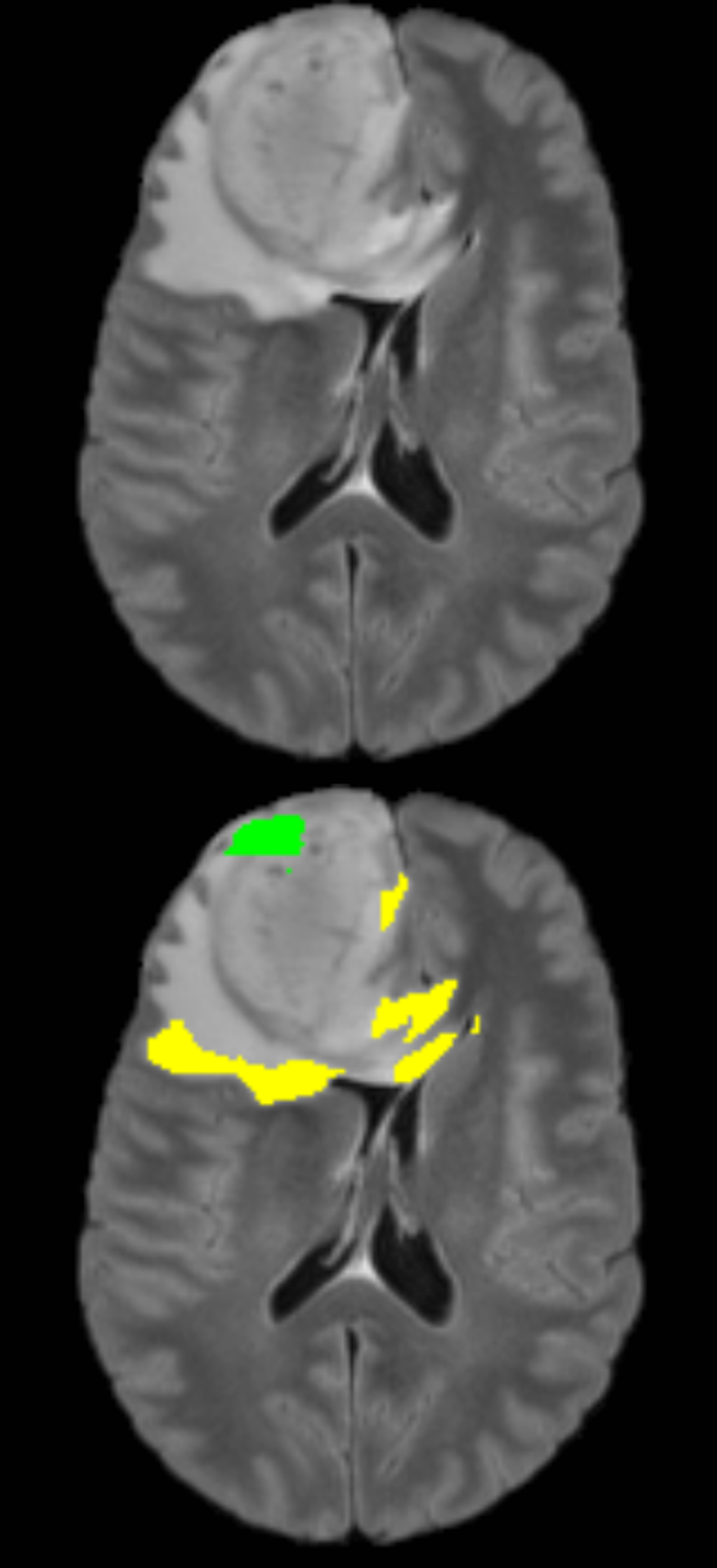}
         \caption{BRATS dataset (tumour)}
         \label{fig:neg controls brats}
     \end{subfigure}
     \hfill
     \begin{subfigure}[]{0.49\textwidth}
         \centering
         \includegraphics[width=\textwidth]{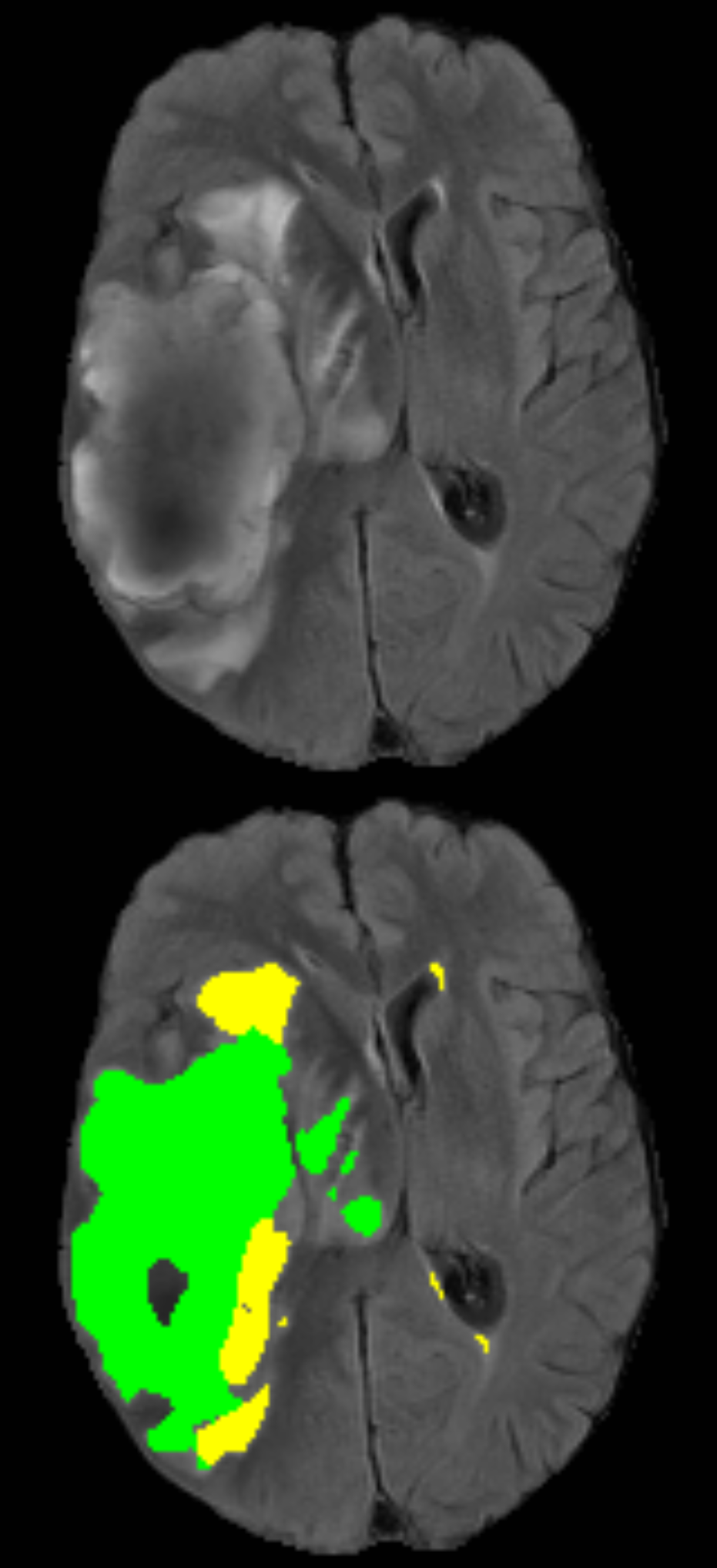}
         \caption{LINCHPIN dataset (haemorrhage)}
         \label{fig:neg controls linchpin}
     \end{subfigure}
        \caption{Test set predictions of the pseudolabels model on negative controls with visible pathology. Yellow: white matter hyperintensities (WMH), green: ischaemic stroke lesions (ISL).}
        \label{fig:neg controls}
\end{figure}

The multi-model configurations demonstrated the best performance, achieving the joint lowest false-positive rates for each individual class and the lowest overall. Both the multi-model variants and the marginal loss model demonstrated statistically significant improvements over the multiclass baseline for \acrshort{isl} segmentation. While the performance variability among the models is considerable, its direct clinical impact is likely minimal; patients presenting with brain tumours typically follow distinct clinical and diagnostic pathways. Moreover, in a high-throughput research or clinical trial setting, automated anomaly detection pipelines could be deployed upfront to filter out confounding oncological cases. Given that scan-level classification does not require laborious voxel-wise annotations, acquiring sufficient datasets to train robust triage models is highly tractable. A representative prediction from the pseudolabels model on a BRATS case is shown in \textbf{Figure \ref{fig:neg controls brats}}.

% negative contol images

\paragraph{Negative controls with haemorrhages}

The LINCHPIN dataset, reserved exclusively for out-of-distribution evaluation, lacks ground truth haemorrhage masks; however, due to the limited sample size ($n=2$), we conducted a qualitative assessment of these cases. Every evaluated model generated false-positive \acrshort{wmh} and \acrshort{isl} predictions within the haemorrhage-affected tissue of both subjects. Again, this is likely to be of little clinical impact, since acute haemorrhage represents a distinct neurovascular emergency, meaning that these patients are managed via dedicated acute clinical pathways. In a research setting, similarly to brain brain tumour subjects, automated anomaly detectors could be effectively leveraged to filter out such scans. An illustrative prediction from the pseudolabels model on this cohort is displayed in \textbf{Figure \ref{fig:neg controls linchpin}}.

\subsection{Pseudolabel model outputs are as plausible as ground truth in most cases} \label{sec: qualitative}

% worse - different - better (images)
\begin{figure}
    \centering
    \includegraphics[width=0.8\textwidth]{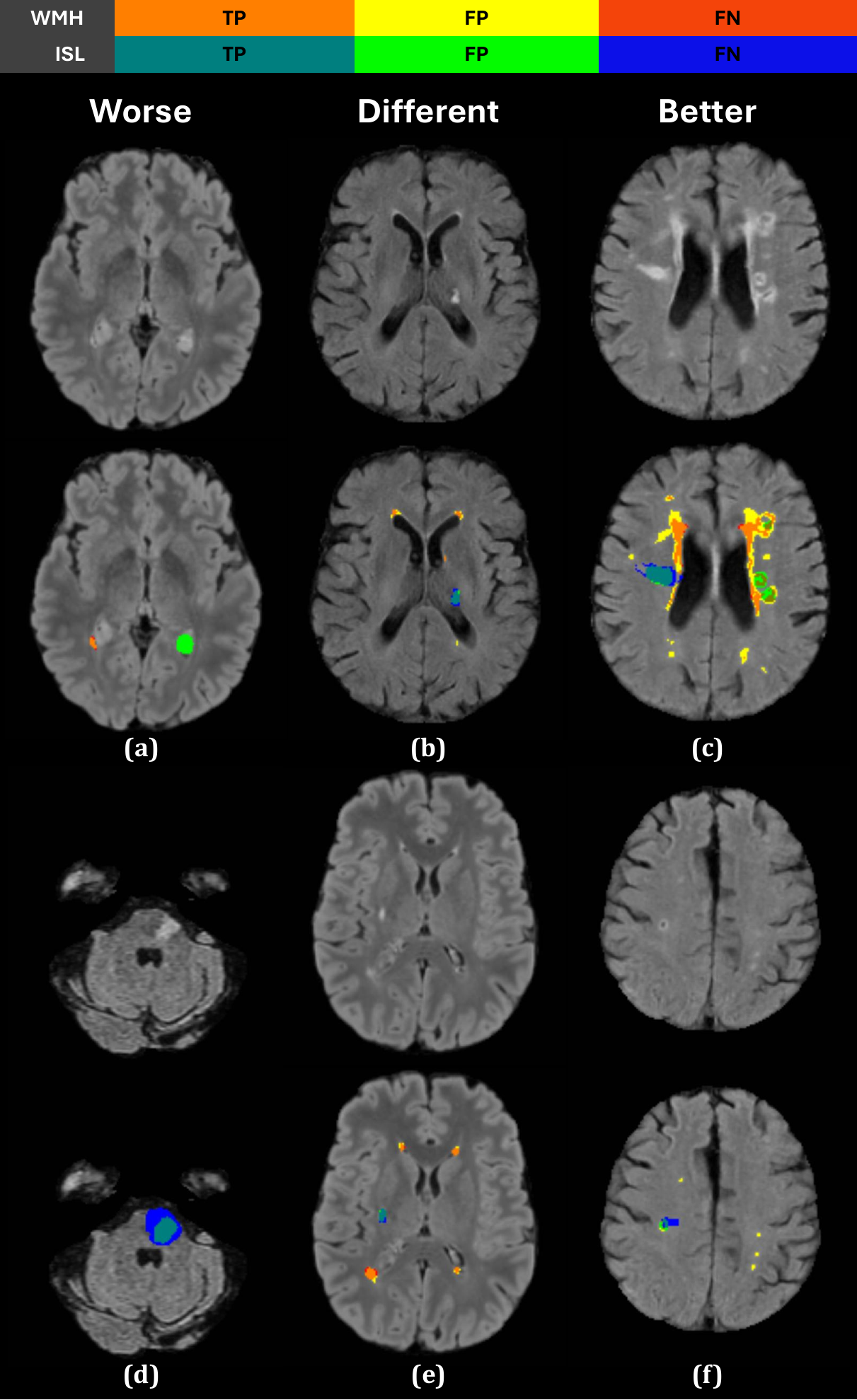}
    \caption{Test set predictions of the pseudolabels model, grouped into predictions that appear worse, different, or better than the ground truth labels according to qualitative evaluation. WMH: white matter hyperintensity, ISL: ischaemic stroke lesion, TP: true positive, FP: false-positive, FN: false-negative.}
    \label{fig:qualitative analysis}
\end{figure}

Given its strong quantitative performance, a qualitative analysis of the pseudolabels model's predictions was conducted. Detailed evaluations of representative cases, presented in \textbf{Figure \ref{fig:qualitative analysis}}, are discussed below.

% a
\paragraph{\textbf{Figure \ref{fig:qualitative analysis}a}}
Here we can see a false-positive \acrshort{isl} predicted by the model. This is objectively a bad prediction, since a human expert would not make this error. It is interesting to note that this anatomical feature mistaken for an \acrshort{isl} is relatively symmetrical, but only the right hemisphere has a predicted \acrshort{isl}.

% d
\paragraph{\textbf{Figure \ref{fig:qualitative analysis}d}}
This example depicts an \acrshort{isl} that was correctly detected, albeit with a significantly underestimated spatial extent relative to the ground truth. While approximately half of the ground truth annotation encompasses hyperintense tissue, the remainder is hypointense, exhibiting textural differences from healthy tissue. Although comprehensive segmentation of such heterogeneous lesions is clinically desirable, the model's failure to capture the hypointense region is an anticipated consequence of our data curation strategy which was implemented for the purpose of filtering out lesions annotated on DWI which were not visible on FLAIR, but which likely also filtered out some low contrast, visible lesions. Consequently, the model's prediction is largely restricted to the hyperintense core.

% b
\paragraph{\textbf{Figure \ref{fig:qualitative analysis}b}} 
This example demonstrates a more expansive prediction of periventricular \acrshort{wmh} by the model compared to the ground truth. This discrepancy reflects a more conservative ground truth segmentation policy regarding regions of subtle hyperintensity. Conversely, the \acrshort{isl} is accurately localised, though a marginal boundary mismatch is evident; the model yields a tighter boundary, adhering strictly to the hyperintense tissue, whereas the ground truth is slightly more expansive.

% e
\paragraph{\textbf{Figure \ref{fig:qualitative analysis}e}}
This case exhibits good agreement between the ground truth and the model prediction for both \acrshort{wmh} and \acrshort{isl} (with a slight offset). The example illustrates well a case where two different masks, while not identical, are likely to be of the same clinical utility.

% c
\paragraph{\textbf{Figure \ref{fig:qualitative analysis}c}}
This example presents a complex case with a high \acrshort{wmh} burden. The model segments a broader extent of periventricular \acrshort{wmh} than the ground truth. This divergence highlights differing implicit segmentation policies; the ground truth is conservative in regions of subtle hyperintensity, whereas the model extends its prediction until the tissue reaches normal appearance. Given the subjective nature of boundary delineation across continuous intensity gradients, both interpretations possess clinical validity. Notably, the model correctly detects several deep \acrshort{wmh} instances omitted from the ground truth. In the left hemisphere, an \acrshort{isl} is correctly identified, with the model's prediction conforming more tightly to the apparent FLAIR-visible boundaries than the ground truth. This discrepancy exemplifies a persistent challenge in dataset curation: ground truth annotations for \acrshort{isl} are frequently derived using complementary modalities (e.g., DWI). This introduces cross-modal discrepancies, such as the DWI-FLAIR mismatch phenomenon, or morphological distortions resulting from the superimposition of lower-resolution DWI masks onto higher-resolution FLAIR scans. Moreover, sometimes the ground truth is produced at an earlier time point and propagated to the follow-up scan via registration, resulting in border differences due to lesion progression. As the model is trained only on FLAIR scans, it naturally diverges from ground truth masks influenced by these external factors. Finally, in the right hemisphere, the model identifies three cavitated hyperintense regions as \acrshort{isl}. Retrospective evaluation by an expert neuroradiologist confirmed that, based exclusively on the provided FLAIR sequence, these predictions are radiologically plausible.

% f
\paragraph{\textbf{Figure \ref{fig:qualitative analysis}f}}
Here the model accurately delineates the FLAIR-visible border of a cavitated \acrshort{isl}, whereas the ground truth annotation diverges significantly from the apparent lesion boundaries in the FLAIR sequence. Additionally, the model successfully identifies deep \acrshort{wmh} that were not segmented in the reference standard.

\vspace{10pt}
In summary, the pseudolabel model generates \acrshort{wmh} predictions that are consistently comparable in radiological plausibility to the ground truth, often exhibiting smoother boundary delineations. The model demonstrates a systemic tendency toward more expansive segmentation of periventricular \acrshort{wmh} and generally exhibits heightened sensitivity to subtle deep \acrshort{wmh} compared to the ground truth. While the model reliably detects both prominent and subtle \acrshort{isl} manifestations, it exhibits diminished sensitivity to lesions (or parts of lesions) characterised by low contrast relative to the surrounding healthy tissue.

\subsection{Evaluation challenges and (pseudolabel-trained) model utility} \label{eval and util}

\begin{figure}[htbp]
     \centering
     \begin{subfigure}[]{0.49\textwidth}
         \centering
         \includegraphics[width=\textwidth]{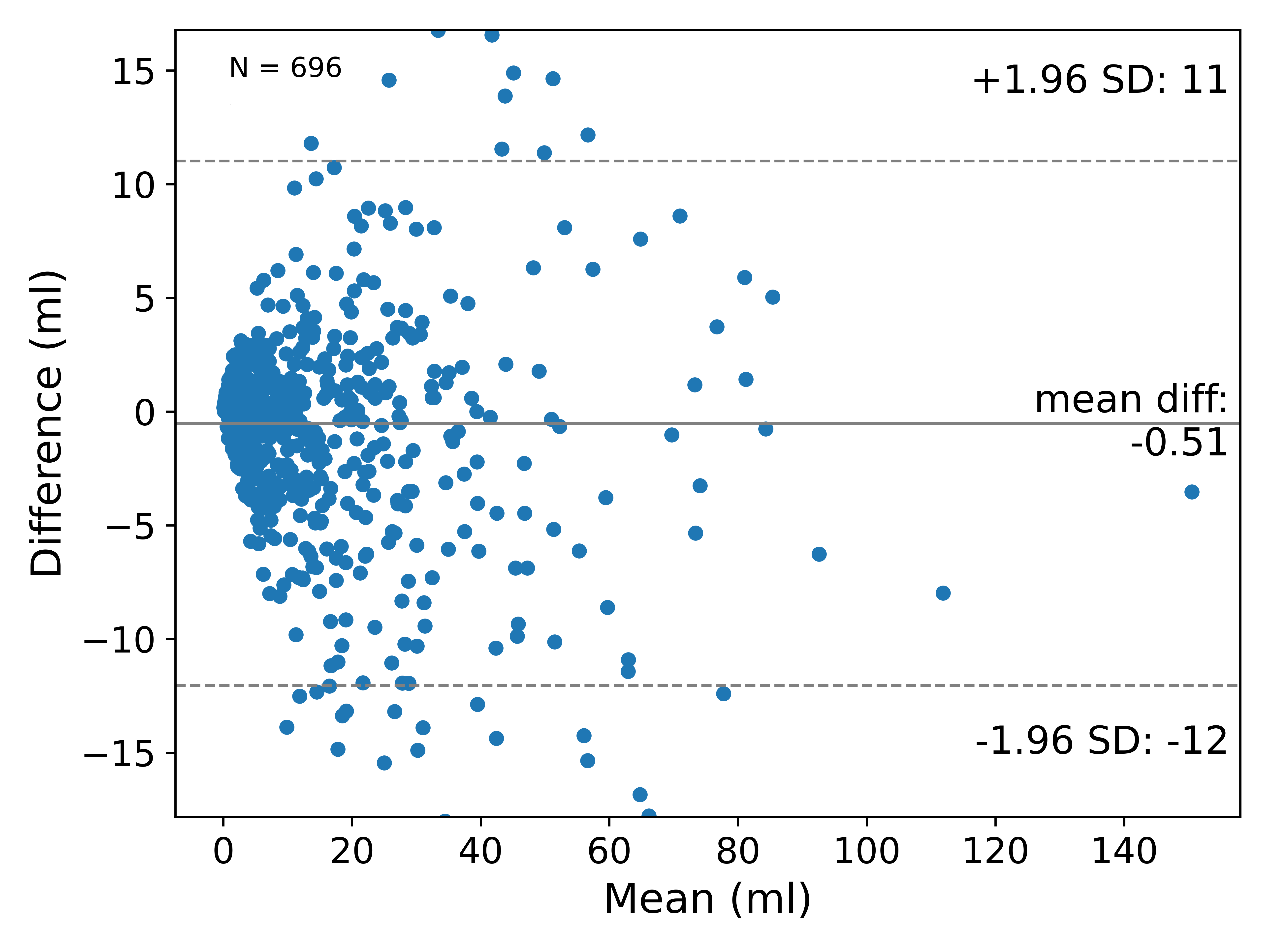}
         \caption{Bland-Altman plot, showing no significant pattern of over or under segmentation. SD: standard deviation.}
         \label{fig:bland altman}
     \end{subfigure}
     \hfill
     \begin{subfigure}[]{0.49\textwidth}
         \centering
         \includegraphics[width=\textwidth]{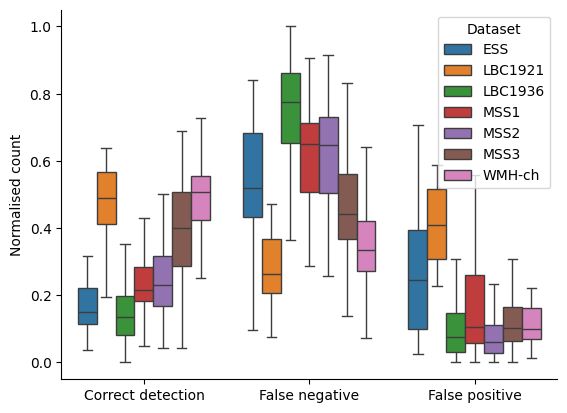}
         \caption{Lesion-level overlap error type distribution per test dataset, normalised over ground truth lesion count. Outliers were removed to aid visualisation and boxplot whiskers extend to 1.5 times the interquartile range.}
         \label{fig:nascimento boxplot 1}
     \end{subfigure}
        \caption{White matter hyperintensity (WMH) volumetric and lesion-level performance plots for the pseudolabels model.}
        \label{fig:nascimento boxplots}
\end{figure}

\paragraph{White matter hyperintensities (WMH)}
Quantitative evaluation of \acrshort{wmh} segmentation presents two primary challenges: 1) high inter-rater variability among experts, meaning that conventional metric degradation may reflect policy divergence rather than segmentation failure \cite{philps2024stochastic}; and 2) the high morphological variance of \acrshort{wmh} instances, where the successful delineation of large confluent lesions can mask the omission of smaller punctate lesions, complicating the interpretation of voxel-wise overlap metrics. To address the first challenge, we employ the \acrfull{ddsc} introduced in Section \ref{metrics description}. The \acrshort{ddsc} mitigates the impact of boundary uncertainty and the penalisation of sub-threshold-sized lesion detection discrepancies (diameter $< \theta$). Setting $\theta=2$ mm, the model achieves a mean \acrshort{ddsc} of 94.67\%, substantially higher than the 67.22\% obtained using the standard \acrshort{dsc}, highlighting the general plausibility of the predictions. Regarding the second challenge, while lesion-level metrics are theoretically advantageous since lesions of all sizes are equally weighted, our qualitative analysis revealed considerable noise, disconnected components, and systematic omission of deep \acrshort{wmh} within several ground truth cohorts. Consequently, \acrshort{lpre} and \acrshort{lrec} become less reliable. Instead, we pivot to the metric of highest clinical consequence: total \acrshort{wmh} volume, a direct surrogate for disease burden tracking today. The Bland-Altman analysis (\textbf{Figure \ref{fig:bland altman}}) demonstrates a minimal mean discrepancy of -0.51 ml with no systematic over- or under-segmentation biases. While future pathophysiological insights may allow us to take advantage of lower-level topological features (e.g., fine-grained spatial distribution or connectivity), these robust volumetric results strongly validate the immediate clinical utility of the model's \acrshort{wmh} segmentation capabilities.

\paragraph{Ischaemic stroke lesions (ISL)}
While the performance metrics for \acrshort{isl} segmentation are competitive with existing literature, they are inherently constrained by the reference standards. As noted in the qualitative analysis, sometimes the ground truth is derived using other sequences or even time points, leading to unavoidable disagreement between model and ground truth segmentation. Based on extensive visual inspection across the large test cohort, the pseudolabel-trained model reliably detects the majority of \acrshort{isl} presentations that a human expert could reasonably identify from FLAIR imaging alone. Consequently, we would suggest that the main utility of the \acrshort{isl} segmentation capabilities of the model are in identify relatively obvious FLAIR-positive \acrshort{isl} in large datasets, for clinical research purposes, or tracking the progression of elderly \acrshort{svd} patients with confirmed FLAIR-positive \acrshort{isl}.

% Boxplots showing huge deviation within overlap error types between datasets, suggest diff annotation policy
% Confirm with qualitative analysis that annotation policy is the difference, not model policy

% LBC1921 vs 1936
\begin{figure}[htbp]
     \centering
     \begin{subfigure}[]{0.49\textwidth}
         \centering
         \includegraphics[width=\textwidth]{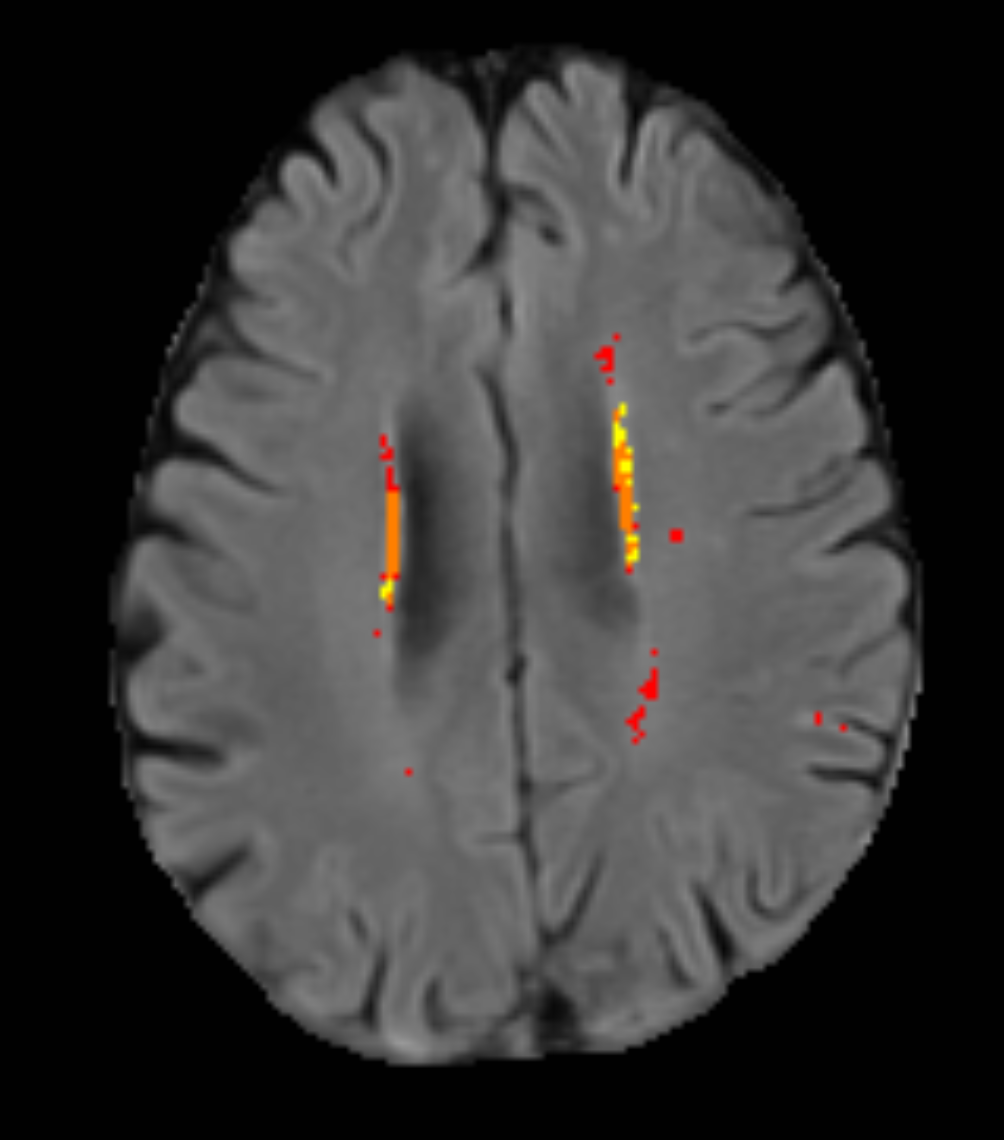}
         \caption{}
         \label{fig:LBC1936 lab}
     \end{subfigure}
     \hfill
     \begin{subfigure}[]{0.49\textwidth}
         \centering
         \includegraphics[width=\textwidth]{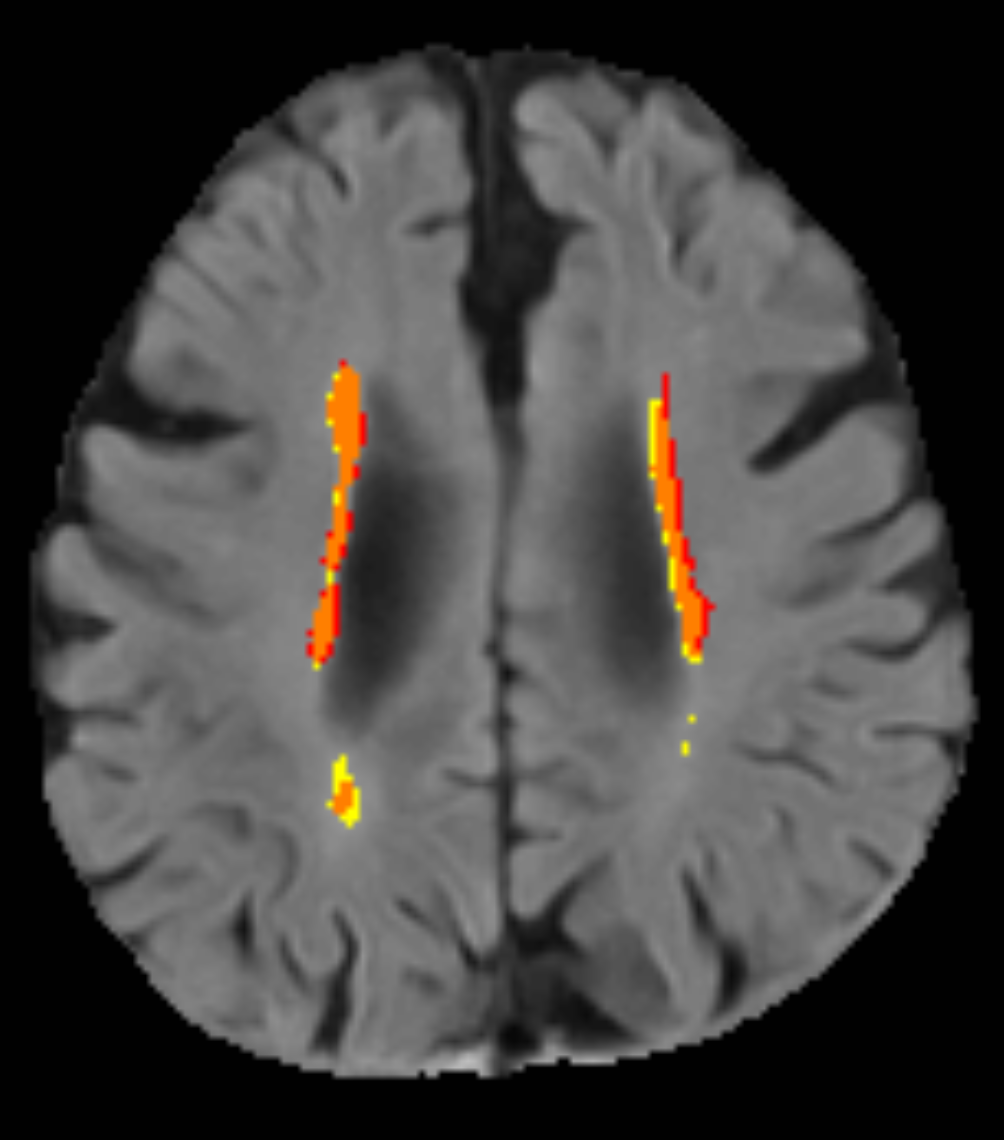}
         \caption{}
         \label{fig:LBC1921 lab}
    \end{subfigure}
    \caption{Ground truth (red) and predicted (yellow)(pseudolabels model) white matter hyperintensities (WMH) for two datasets. (a) An LBC1936 \acrshort{flair} scan where it is clear that the ground truth includes very small punctate areas of hyperintensity, often no larger than a single voxel, while the prediction is more conservative; (b) An LBC1921 \acrshort{flair} scan where the ground truth does not include small disconnected \acrshort{wmh}, but rather smoothly segmented areas of hyperintensity. This figure highlights the difference in annotation policy between LBC1936 and LBC1921.}
    \label{fig:nascimento images}
\end{figure}

\paragraph{Consistent segmentation policy}
The qualitative evaluation further highlights a secondary utility of the proposed model. While ground truth annotation policies fluctuate significantly across datasets, the pseudolabel model applies a deterministic and consistent segmentation policy. This consistency provides a mechanism to retroactively audit and quantify the implicit annotation biases of different training datasets, assuming comparable underlying patient demographics. This capability is useful for researchers seeking to curate datasets that align with specific annotation philosophies (e.g., highly conservative versus inclusive boundary delineation). For instance, \textbf{Figure \ref{fig:nascimento boxplot 1}} illustrates that, for \acrshort{wmh}, lesion-level false-negatives are systematically more prevalent in the LBC1936 dataset compared to LBC1921, while the inverse is true for false-positives. Given the model's consistent segmentation policy, this discrepancy suggests that the ground truth annotation policy for LBC1921 is more conservative than that of LBC1936. This conclusion is visually corroborated by \textbf{Figure \ref{fig:nascimento images}} and by the comparative density of probabilistic false-positive and false-negative error maps between \textbf{Figures \ref{fig:lbc1936_prob_maps_full_gt}} and \textbf{\ref{fig:lbc1921_prob_maps_full_gt}} in \ref{appendix B fp fn}.

\section{Conclusions}

While complex and highly specialised architectures are frequently proposed for partially labelled medical image segmentation, this study demonstrates the efficacy of foundational, straightforward training strategies. Aligning with the established principle that simple, robust techniques often prove to be the most effective, practical, and widely adopted (e.g., SGD, combined Dice and cross-entropy loss) \cite{isensee2021nnu,isensee2024nnu}, we systematically evaluated six accessible training paradigms.

Our findings confirm that incorporating partially labelled training data is both feasible and beneficial for the simultaneous segmentation of the visually confounding \acrshort{wmh} and \acrshort{isl} pathologies. With the exception of the class-conditional model, all evaluated methods successfully leveraged the partially labelled data to improve upon the multiclass baseline in our primary metric, macro-average \acrfull{ap}. While the pseudolabel-trained model achieved the strongest overall quantitative performance, its training pipeline is more computationally demanding than the class-adaptive and marginal loss models, which also delivered robust results. Additionally, the multi-model approach proved highly competitive; despite its simplicity, it achieved the highest \acrshort{dsc} for \acrshort{wmh} segmentation and the lowest number of scan- and tumour-level false positives. Ultimately, practitioners can confidently integrate most of these methods to effectively utilise partially labelled data, provided they weigh their specific practical trade-offs: the class-adaptive and marginal loss approaches are best if training time is a priority, whereas the multi-model approach should be avoided if fast inference speed is critical.

Future research should investigate how these practical methods translate to training regimes entirely devoid of fully labelled cases, or to cohorts lacking scans with both pathologies concurrently visible. Finally, given the superior performance of the pseudolabels approach, subsequent studies should examine the sensitivity of the final model's performance to the choice of pseudolabel generation method.

\paragraph{\textbf{Limitations}}
Despite our best efforts to ensure the robustness of our experimental design, several limitations must be acknowledged. \textbf{First}, the absence of comprehensive demographic metadata across all datasets limits our ability to assess potential algorithmic biases or stratify performance across distinct patient populations. \textbf{Second}, our evaluation benefited from a significant number of fully labelled scans, thanks to our privately held datasets. Given the current lack of publicly available data featuring dual annotations for both \acrshort{wmh} and \acrshort{isl}, the generalisability of these training paradigms to purely partially labelled environments remains unknown. Relatedly, our training data contained a high prevalence of cases where both pathologies co-occurred, even when only one was formally annotated. The resilience of the decoupled multi-model approach when trained on data where such co-occurrence is rare requires further investigation. \textbf{Third}, the automated curation strategy used to exclude \acrshort{isl} cases unidentifiable on the FLAIR sequence alone may have inadvertently filtered out subtle, low-contrast lesions. Consequently, the models were not trained or evaluated on these diagnostically challenging cases. \textbf{Fourth}, the inclusion of partially labelled data within the evaluation splits inherently limits our capacity to systematically quantify false-positive predictions for unannotated classes across the entire test set. \textbf{Fifth}, driven by the objective of establishing accessible and generalisable baselines, we explicitly excluded any complex and highly bespoke methods from the literature from our comparative analysis. \textbf{Sixth}, as discussed at length in this paper, the ground truth labels were derived using several different implicit segmentation policies, and were of varying quality. \textbf{Seventh}, we chose to utilise an ensemble of three models (trained with different random seeds) to evaluate each method, since this reflects a more realistic deployment scenario. As a consequence of this, while our confidence intervals express uncertainty over the test set, we did not report measures of variability over training runs. \textbf{Eighth}, and directly related to the previous point, while our bootstrapping simulates variability in the test set, our use of a single training and validation split is a limitation. \textbf{Finally}, the generality of the observed rankings remains to be demonstrated with different base neural network architectures.

\section*{Acknowledgements}

We thank the Lothian Birth Cohorts (LBC), Mild Stroke Studies, and LINCHPIN research groups, patients/study participants, and radiographers at the Brain Research Imaging Centre and Edinburgh Hospitals for their contribution in providing and acquiring the data used in this study, Dr. Susana Muñoz Maniega for LBC image data administration and processing, Dr. Mark E. Bastin for LBC image data protocol design and MRI quality assurance, Dr. Simon R. Cox and Professor Ian J. Deary - LBC study directors.

J.P. is funded by Medical Research Scotland [ref. PHD-50441-2021] and Canon Medical Research Europe. Funding from Row Fogo Charitable Trust (Ref No: AD.ROW4.35. BRO-D.FID3668413), and the UK Medical Research Council (UK Dementia Research Institute at the University of Edinburgh, award number UK DRI-4002; G0700704/84698) are also gratefully acknowledged. M.O.B. gratefully acknowledges funding from: EPSRC grant no. EP/X025705/1; British Heart Foundation and The Alan Turing Institute Cardiovascular Data Science Award (C-10180357); the SCONe projects funded by Chief Scientist Office, Edinburgh \& Lothians Health Foundation, Sight Scotland, the Royal College of Surgeons of Edinburgh, the RS Macdonald Charitable Trust, and Fight For Sight.

The data used in this project received funds from the Chief Scientist Office of the Scottish Executive (CZB/4/281), the Wellcome Trust (WT075611, WT088134/Z/09/A, Edinburgh Clinical Academic Track PhD Programme), the UK Dementia Research Institute funded by the UK MRC, Alzheimer’s Society and Alzheimer’s Research UK through the UK DRI, the Fondation Leducq Network for the Study of Perivascular Spaces in Small Vessel Disease (16 CVD 05), the Stroke Association (“Small Vessel Disease-Spotlight on Symptoms” (SAPG 19\textbackslash100068) and “LINCHPIN Study”), Age UK (the Disconnected Mind project), the UK Medical Research Council (G0701120, G1001245, MR/M013111/1, MR/R024065/1), joint funding from the Medical Research Council and the Biotechnology and Biological Sciences Research Council (MR/K026992/1 for the Centre for Cognitive Ageing and Cognitive Epidemiology), joint funding from the Biotechnology and Biological Sciences Research Council and the Economic and Social Research Council (BB/W008793/1), and the University of Edinburgh.

\appendix

\section{Region-of-interest (ROI)-based analysis of the pseudolabel model predictions} \label{appendix A} 

While the main text reports global segmentation metrics to evaluate and compare the overall efficacy of the proposed training strategies, this appendix provides a more granular, region-of-interest (ROI)-based analysis of the pseudolabels model, which showed the best overall performance. Here, we decompose the predictions for each test volume into clinically relevant sub-regions to assess spatially and morphologically conditioned model performance.

\subsection{White matter hyperintensities (WMH) analysis} \label{appendix A wmh}

\subsubsection{Region-of-interest (ROI) selection}

At the scan level, the test set was stratified into low, medium, and high \acrshort{wmh} burden subjects based on total lesion volume, utilising the tertiles of the complete dataset distribution.

Furthermore, ground truth and predicted \acrshort{wmh} masks were spatially partitioned into \acrfull{pvwmh} and \acrfull{dwmh}, given their distinct clinical implications \cite{kim2015periventricular}. \acrshort{pvwmh} typically originate adjacent to the ventricular walls and are canonically defined as residing within a 10 mm distance threshold from the ventricles \cite{chen2021bilateral,ottavi2023consensus}. However, advancing \acrshort{pvwmh} often exhibit confluent growth that extends beyond this boundary. Therefore, we initially classified any 6-connected \acrshort{wmh} component that partially intersected this 10 mm ventricular boundary as \acrshort{pvwmh}. 

\begin{figure}[htbp]
     \centering
     \begin{subfigure}[]{0.32\textwidth}
         \centering
         \includegraphics[width=\textwidth]{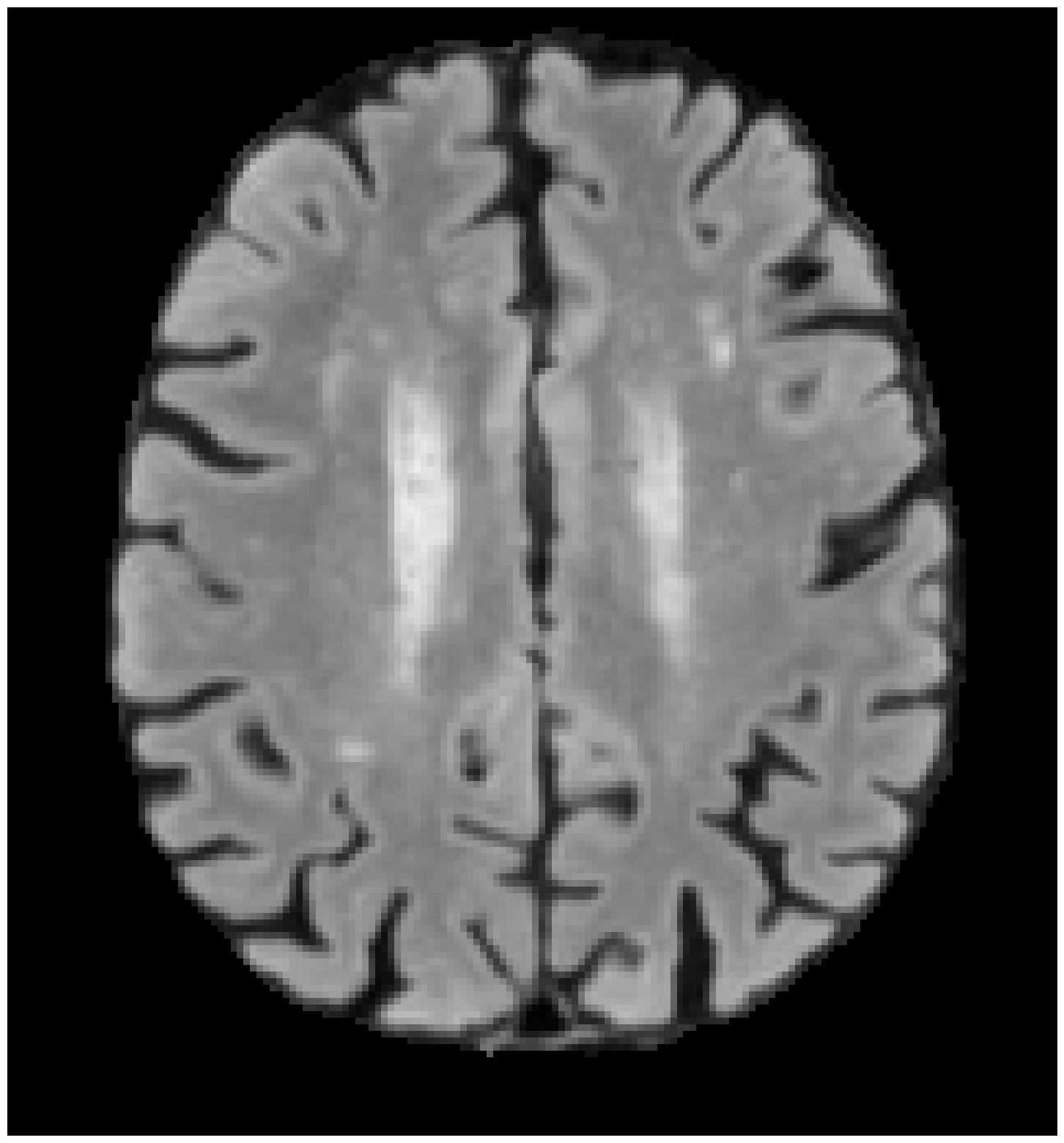}
         \caption{}
         \label{fig:PV vs D image}
     \end{subfigure}
     \hfill
     \begin{subfigure}[]{0.32\textwidth}
         \centering
         \includegraphics[width=\textwidth]{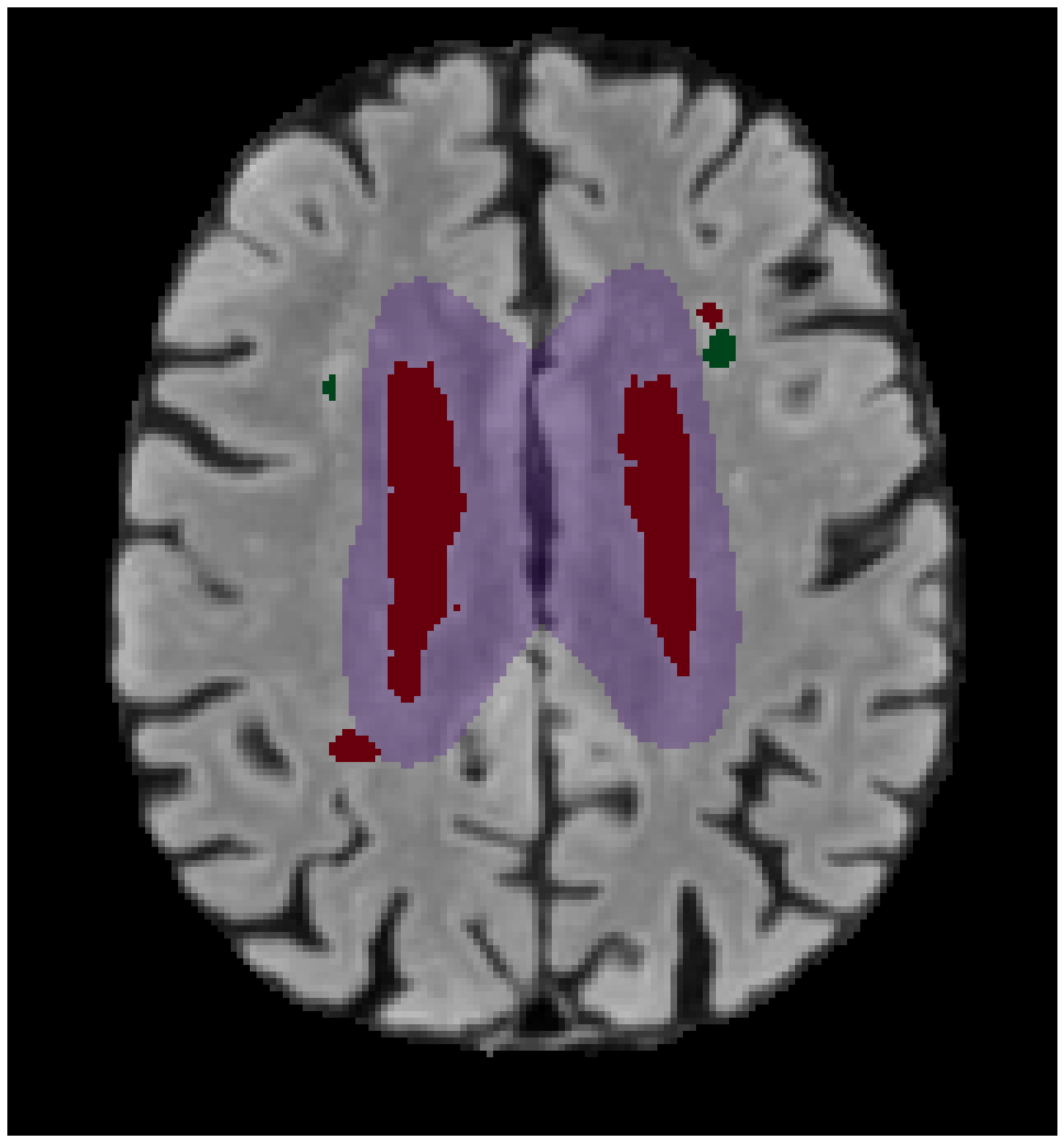}
         \caption{}
         \label{fig:PV vs D simple}
     \end{subfigure}
     \hfill
     \begin{subfigure}[]{0.32\textwidth}
         \centering
         \includegraphics[width=\textwidth]{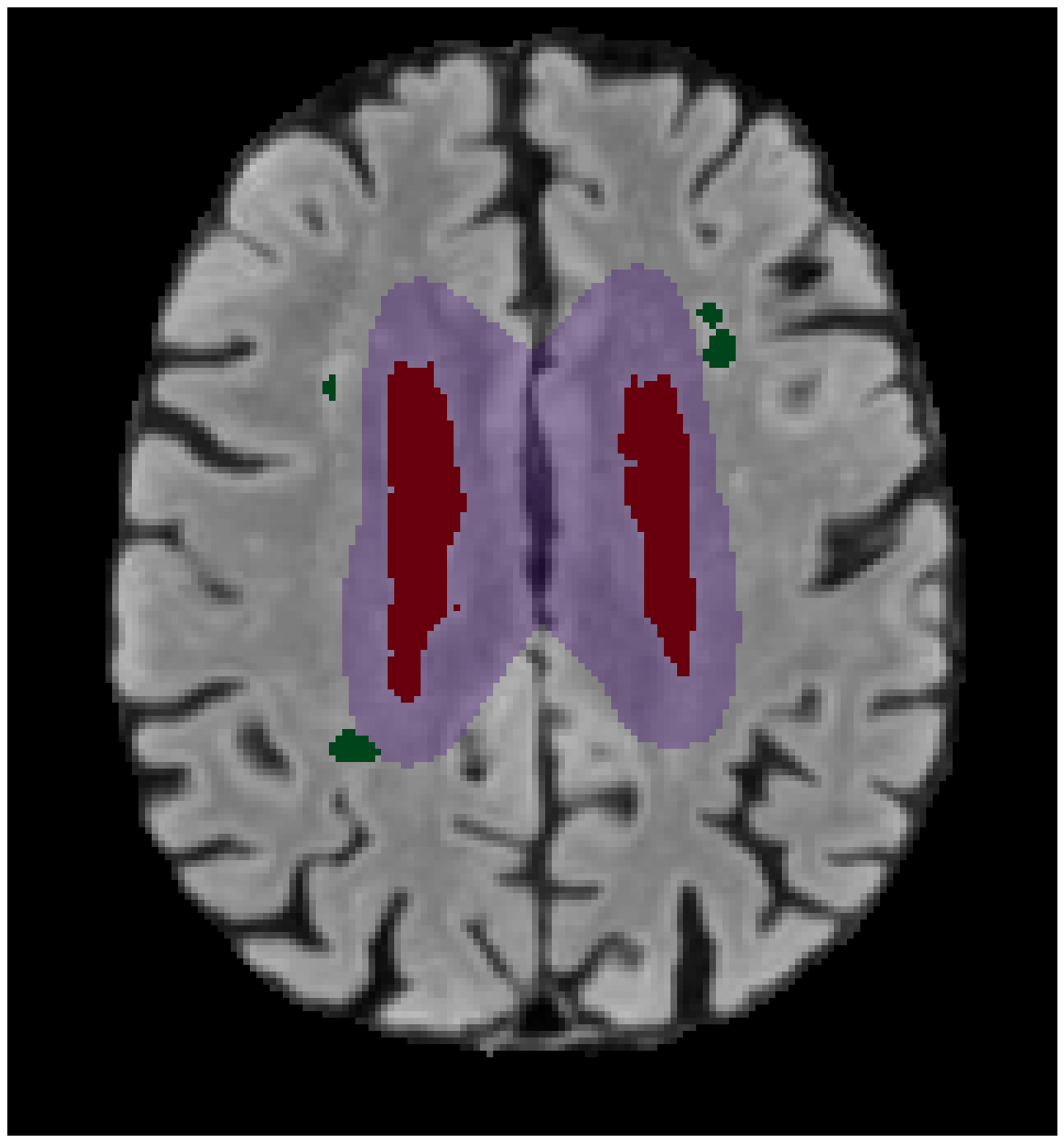}
         \caption{}
         \label{fig:PV vs D better}
     \end{subfigure}
        \caption{(a) FLAIR image; (b) initial periventricular white matter hyperintensity (PVWMH) and deep white matter hyperintensity (DWMH) masks; (c) final PVWMH and DWMH masks after removing easily disconnected components (disconnected with a single erosion) which do not fall within 10 mm from the ventricles from the PVWMH mask. Purple: 10 mm ventricle mask, red: PVWMH, green: DWMH.}
        \label{fig:PV vs D}
\end{figure}

Although this spatial rule generally aligns with expert consensus, visual inspection reveals instances where proximal but distinct \acrshort{dwmh} lesions merge with \acrshort{pvwmh} into a single connected component, leading to misclassification. \acrshort{dwmh} typically present as focal, non-confluent lesions. To resolve these topological mergers, a binary erosion morphological operation was applied to the initial \acrshort{pvwmh} mask. Any resulting newly disconnected component that fell entirely outside the 10 mm ventricular boundary was subsequently reclassified as \acrshort{dwmh}. The complete procedural logic is detailed in \textbf{Algorithm \ref{alg:WMH}}, with a visualisation provided in \textbf{Figure \ref{fig:PV vs D}}. Ventricular structures were automatically segmented using SynthSeg \cite{billot2023synthseg} to define the initial distance boundary.

\begin{algorithm}[htbp]
\caption{Splitting \acrshort{wmh} mask into \acrshort{pvwmh} and \acrshort{dwmh} masks}\label{alg:WMH}
\footnotesize
\begin{algorithmic}[1]
    \Require Total \acrshort{wmh} binary mask $M$, Ventricle binary mask $V$.
    \State \textbf{Define} $CC(\cdot)$ as 3D 6-connected component labeling.
    \State \textbf{Define} $Dil(\cdot, d)$ and $Ero(\cdot, d)$ as binary dilation and erosion by distance $d$.
    \State \textbf{Define} $Mask(L, \mathcal{S})$ as a function returning the binary mask of a set of labels $\mathcal{S}$ from labelled image $L$.
    \State \textbf{Define} $Labels(L, M)$ as a function returning the set of unique, non-zero labels in $L$ that intersect with binary mask $M$.

    \State $L_M \gets CC(M)$ \Comment{Label all \acrshort{wmh} components}
    \State $V_{10mm} \gets Dil(V, \text{10 mm})$ \Comment{Dilate ventricles to define the PV zone}
    
    \State $\mathcal{L}_{initial\_PV} \gets Labels(L_M, V_{10mm})$ \Comment{Find all \acrshort{wmh} touching the 10 mm PV zone}
    \State $M_{PV} \gets Mask(L_M, \mathcal{L}_{initial\_PV})$ \Comment{Form initial \acrshort{pvwmh} mask}
    
    \State $M_{PV\_ero} \gets Ero(M_{PV}, \text{1 voxel})$ \Comment{Erode to break tiny ``bridges'' to \acrshort{dwmh}}
    \State $L_{PV\_ero} \gets CC(M_{PV\_ero})$ \Comment{Label the eroded mask}
    
    \State $\mathcal{L}_{core} \gets Labels(L_{PV\_ero}, V_{10mm})$ \Comment{Find eroded WMH still in the PV zone}
    \State $\mathcal{L}_{disconnected} \gets \text{All labels in } L_{PV\_ero} \setminus \mathcal{L}_{core}$ \Comment{Isolate easily disconnected \acrshort{dwmh}}
    
    \State $M_{disconnected} \gets Mask(L_{PV\_ero}, \mathcal{L}_{disconnected})$ \Comment{Mask of eroded WMH}
    \State $M_{disconnected} \gets Dil(M_{disconnected}, \text{1 voxel})$ \Comment{Restore disconnected WMH size}
    
    \State $M_{PV} \gets M_{PV} \setminus M_{disconnected}$ \Comment{Remove disconnected WMH for final \acrshort{pvwmh} mask}
    \State $M_D \gets M \setminus M_{PV}$ \Comment{Final \acrshort{dwmh} mask is the remainder of the total \acrshort{wmh}}
    
    \Ensure $M = M_{PV} \cup M_D$ and $M_{PV} \cap M_D = \emptyset$ \Comment{Ensure masks exhaustive and exclusive}
    \State\Return $M_{PV}, M_D$
\end{algorithmic}
\normalsize
\end{algorithm}

\subsubsection{Metrics}

The average precision (\acrshort{ap}) metric, utilised in the primary analysis for threshold-independent model comparison, was omitted here, as we were no longer comparing models. Instead, we evaluated voxel-level \acrfull{pre} and \acrfull{rec}, alongside their distance-adjusted analogues: \acrfull{dpre} and \acrfull{drec}. These represent the precision and recall components of the \acrfull{ddsc} metric detailed in Section \ref{metrics description} of the main text.

\subsubsection{Results}

\textbf{Table \ref{tab:WMH ROIs}} presents the quantitative outcomes of the ROI-based \acrshort{wmh} evaluation. The results indicate that scans with a high \acrshort{wmh} burden consistently yield more favourable metric scores. This effect is largely attributable to the geometric properties of the lesions: the inter-rater subjectivity in \acrshort{wmh} segmentation predominantly occur at lesion boundaries and within small, punctate foci. Cases with high lesion burdens generally feature large, confluent \acrshort{wmh}, which possess a lower surface-area-to-volume ratio, thereby reducing the quantitative penalty of boundary ambiguities incurred by our metrics. 

Additionally, overlap metrics were markedly higher for \acrshort{pvwmh} compared to \acrshort{dwmh}. We attribute this mainly to the fact that \acrshort{pvwmh} are more confluent, predictable in their spatial distribution, and generally represent more total volume, than \acrshort{dwmh}. Drawing upon the qualitative observations presented in the main text---particularly the observation that the ground truth annotations do not consistently demonstrate superior quality or reliability compared to the model's predictions in these challenging deep white matter regions---we hypothesise that this performance disparity is also partially driven by ground truth errors in the \acrshort{dwmh} regions.

\begin{table}[htbp]
\caption{Results for white matter hyperintensity (WMH) segmentation of the pseudolabels model on different regions-of-interest (ROIs). Reported as mean {[}95\% CI{]}  with confidence intervals (CIs) calculated over a 10,000 sample bootstrap per ROI. N is the number of scans with at least one lesion in the ROI and hence the number of scans which most metric means are calculated over (except absolute volume difference (AVD) which can be calculated on empty ground truth cases---for AVD the results are reported as mean \{N\} {[}95\% CI{]} if this number is different). DSC: Dice Similarity Coefficient, PRE: voxel-level precision, REC: voxel-level recall, ASD: average surface distance, DSC$_\theta$: Distance-DSC, PRE$_\theta$: Distance-PRE, REC$_\theta$: distance-REC, AVD: absolute volume difference.}
\label{tab:WMH ROIs}
\resizebox{\textwidth}{!}{%
\begin{tabular}{l|cccccccc}
\hline
ROI (N)               & DSC (\%)                                                           & PRE (\%)                                                           & REC (\%)                                                           & ASD (mm)                                                                & DSC$_\theta$ (\%)                                                          & PRE$_\theta$ (\%)                                                          & REC$_\theta$ (\%)                                                          & AVD (\% ICV)                                                               \\ \hline
Overall (690)         & \begin{tabular}[c]{@{}c@{}}67.22\\[-5pt] \smalltext{{[}65.93, 68.51{]}}\end{tabular} & \begin{tabular}[c]{@{}c@{}}71.95\\[-5pt] \smalltext{{[}70.41, 73.46{]}}\end{tabular} & \begin{tabular}[c]{@{}c@{}}68.45\\[-5pt] \smalltext{{[}66.96, 69.86{]}}\end{tabular} & \begin{tabular}[c]{@{}c@{}}1.54\\[-5pt] \smalltext{{[}1.20, 1.96{]}}\end{tabular}         & \begin{tabular}[c]{@{}c@{}}94.67\\[-5pt] \smalltext{{[}93.76, 95.47{]}}\end{tabular} & \begin{tabular}[c]{@{}c@{}}91.91\\[-5pt] \smalltext{{[}90.80, 92.94{]}}\end{tabular} & \begin{tabular}[c]{@{}c@{}}99.36\\[-5pt] \smalltext{{[}99.16, 99.54{]}}\end{tabular} & \begin{tabular}[c]{@{}c@{}}0.2151 \{696\\[-5pt] \smalltext{{[}0.1928, 0.2389{]}}\end{tabular}   \\ \hline
Low vol (226)         & \begin{tabular}[c]{@{}c@{}}54.11\\[-5pt] \smalltext{{[}51.76, 56.40{]}}\end{tabular} & \begin{tabular}[c]{@{}c@{}}55.98\\[-5pt] \smalltext{{[}53.16, 58.72{]}}\end{tabular} & \begin{tabular}[c]{@{}c@{}}60.83\\[-5pt] \smalltext{{[}57.88, 63.71{]}}\end{tabular} & \begin{tabular}[c]{@{}c@{}}3.23\\[-5pt] \smalltext{{[}2.24, 4.47{]}}\end{tabular}         & \begin{tabular}[c]{@{}c@{}}88.90\\[-5pt] \smalltext{{[}86.51, 91.05{]}}\end{tabular} & \begin{tabular}[c]{@{}c@{}}83.31\\[-5pt] \smalltext{{[}80.56, 85.86{]}}\end{tabular} & \begin{tabular}[c]{@{}c@{}}99.66\\[-5pt] \smalltext{{[}99.27, 99.92{]}}\end{tabular} & \begin{tabular}[c]{@{}c@{}}0.0599 \{232\}\\[-5pt] \smalltext{{[}0.0527, 0.0673{]}}\end{tabular} \\
Medium vol (232)      & \begin{tabular}[c]{@{}c@{}}69.08\\[-5pt] \smalltext{{[}67.46, 70.6{]}}\end{tabular}  & \begin{tabular}[c]{@{}c@{}}76.21\\[-5pt] \smalltext{{[}74.33, 78.05{]}}\end{tabular} & \begin{tabular}[c]{@{}c@{}}67.41\\[-5pt] \smalltext{{[}65.17, 69.70{]}}\end{tabular} & \begin{tabular}[c]{@{}c@{}}0.76\\[-5pt] \smalltext{{[}0.66, 0.88{]}}\end{tabular}         & \begin{tabular}[c]{@{}c@{}}97.26\\[-5pt] \smalltext{{[}96.57, 97.85{]}}\end{tabular} & \begin{tabular}[c]{@{}c@{}}95.52\\[-5pt] \smalltext{{[}94.52, 96.40{]}}\end{tabular} & \begin{tabular}[c]{@{}c@{}}99.38\\[-5pt] \smalltext{{[}99.08, 99.63{]}}\end{tabular} & \begin{tabular}[c]{@{}c@{}}0.1597\\[-5pt] \smalltext{{[}0.1352, 0.1901{]}}\end{tabular}         \\
High vol (232)        & \begin{tabular}[c]{@{}c@{}}78.13\\[-5pt] \smalltext{{[}76.68, 79.51{]}}\end{tabular} & \begin{tabular}[c]{@{}c@{}}83.25\\[-5pt] \smalltext{{[}81.54, 84.90{]}}\end{tabular} & \begin{tabular}[c]{@{}c@{}}76.90\\[-5pt] \smalltext{{[}74.84, 78.87{]}}\end{tabular} & \begin{tabular}[c]{@{}c@{}}0.67\\[-5pt] \smalltext{{[}0.62, 0.73{]}}\end{tabular}         & \begin{tabular}[c]{@{}c@{}}97.68\\[-5pt] \smalltext{{[}97.04, 98.21{]}}\end{tabular} & \begin{tabular}[c]{@{}c@{}}96.68\\[-5pt] \smalltext{{[}95.73, 97.48{]}}\end{tabular} & \begin{tabular}[c]{@{}c@{}}99.05\\[-5pt] \smalltext{{[}98.67, 99.37{]}}\end{tabular} & \begin{tabular}[c]{@{}c@{}}0.4256\\[-5pt] \smalltext{{[}0.3750, 0.4803{]}}\end{tabular}         \\ \hline
Periventricular (687) & \begin{tabular}[c]{@{}c@{}}67.56\\[-5pt] \smalltext{{[}66.27, 68.82{]}}\end{tabular} & \begin{tabular}[c]{@{}c@{}}71.70\\[-5pt] \smalltext{{[}70.21, 73.22{]}}\end{tabular} & \begin{tabular}[c]{@{}c@{}}69.49\\[-5pt] \smalltext{{[}67.98, 70.97{]}}\end{tabular} & \begin{tabular}[c]{@{}c@{}}1.46\\[-5pt] \smalltext{{[}1.18, 1.81{]}}\end{tabular}         & \begin{tabular}[c]{@{}c@{}}94.33\\[-5pt] \smalltext{{[}93.46, 95.14{]}}\end{tabular} & \begin{tabular}[c]{@{}c@{}}91.43\\[-5pt] \smalltext{{[}90.30, 92.45{]}}\end{tabular} & \begin{tabular}[c]{@{}c@{}}99.21\\[-5pt] \smalltext{{[}98.97, 99.42{]}}\end{tabular} & \begin{tabular}[c]{@{}c@{}}0.1996 \{694\}\\[-5pt] \smalltext{{[}0.1786, 0.2219{]}}\end{tabular} \\
Deep (677)            & \begin{tabular}[c]{@{}c@{}}41.11\\[-5pt] \smalltext{{[}39.35, 42.89{]}}\end{tabular} & \begin{tabular}[c]{@{}c@{}}56.44\\[-5pt] \smalltext{{[}54.43, 58.44{]}}\end{tabular} & \begin{tabular}[c]{@{}c@{}}39.47\\[-5pt] \smalltext{{[}37.62, 41.35{]}}\end{tabular} & \begin{tabular}[c]{@{}c@{}}5.26 \{645\}\\[-5pt] \smalltext{{[}4.48, 6.10{]}}\end{tabular} & \begin{tabular}[c]{@{}c@{}}82.98\\[-5pt] \smalltext{{[}81.12, 84.71{]}}\end{tabular} & \begin{tabular}[c]{@{}c@{}}76.68\\[-5pt] \smalltext{{[}74.66, 78.57{]}}\end{tabular} & \begin{tabular}[c]{@{}c@{}}98.75\\[-5pt] \smalltext{{[}98.18, 99.23{]}}\end{tabular} & \begin{tabular}[c]{@{}c@{}}0.0276 \{694\}\\[-5pt] \smalltext{{[}0.0245, 0.0310{]}}\end{tabular} \\ \hline
\end{tabular}%
}
\end{table}

\subsection{Ischaemic stroke lesion (ISL) analysis} \label{appendix A isl}

\subsubsection{Region-of-interest (ROI) selection}

The assignment of \acrshort{isl} to specific ROIs was executed via a two-stage pipeline, comprising an initial spatial allocation followed by an overlap-based assignment correction. We evaluated model performance across three distinct ROI categories: (1) lesion volume, (2) cortical versus subcortical location, and (3) arterial territory. Every 26-connected ground truth \acrshort{isl} component was exclusively assigned to one ROI within each of these three groups (with the exception of the lesion volume group, detailed below).

\paragraph{Initial assignment}
First, we established volume thresholds using the tertiles of the lesion-level rather than scan-level \acrshort{isl} volume distribution across the entire dataset. Each ground truth \acrshort{isl} component in each volume was subsequently categorised into a low, medium, or high-volume ROI. During this initial assignment phase, all predicted lesions were naively assumed to belong to the predicted mask for every volume ROI.

Cortical and subcortical anatomical regions were automatically delineated using SynthSeg. Any ground truth or predicted \acrshort{isl} intersecting with the cortical mask was assigned to the cortical ROI; otherwise, it was designated as subcortical. The clinical rationale for this intersection-based rule is that lesions spanning the cortical-subcortical boundary predominantly originate from occlusions in larger cortical vessels, which consequently induce ischaemia in the smaller, downstream penetrating vessels of the affected subcortical regions.

\begin{figure}[htbp]
     \centering
     \begin{subfigure}[]{0.47\textwidth}
         \centering
         \includegraphics[width=\textwidth]{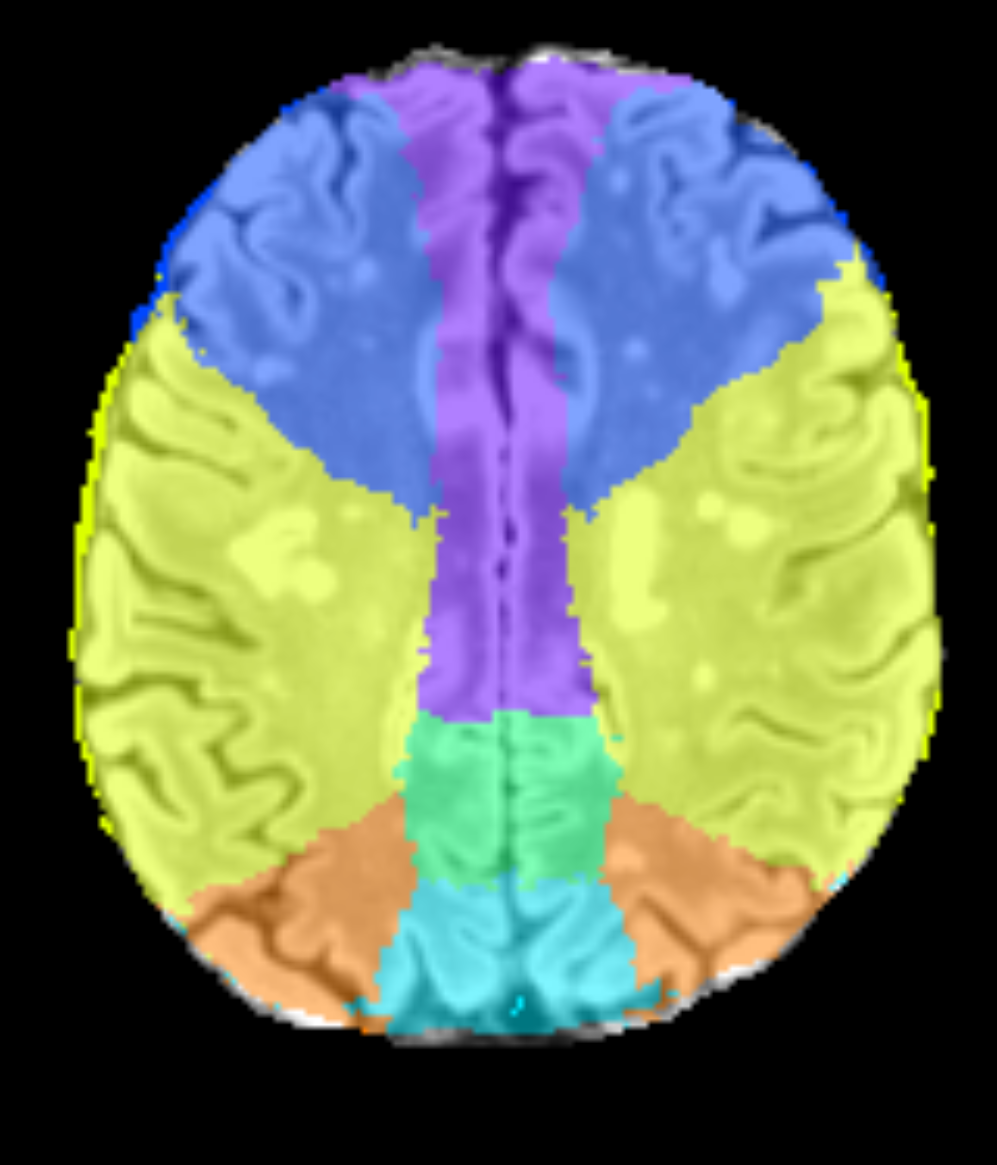}
         \caption{}
         \label{fig:BAT lower}
     \end{subfigure}
     \hfill
     \begin{subfigure}[]{0.47\textwidth}
         \centering
         \includegraphics[width=\textwidth]{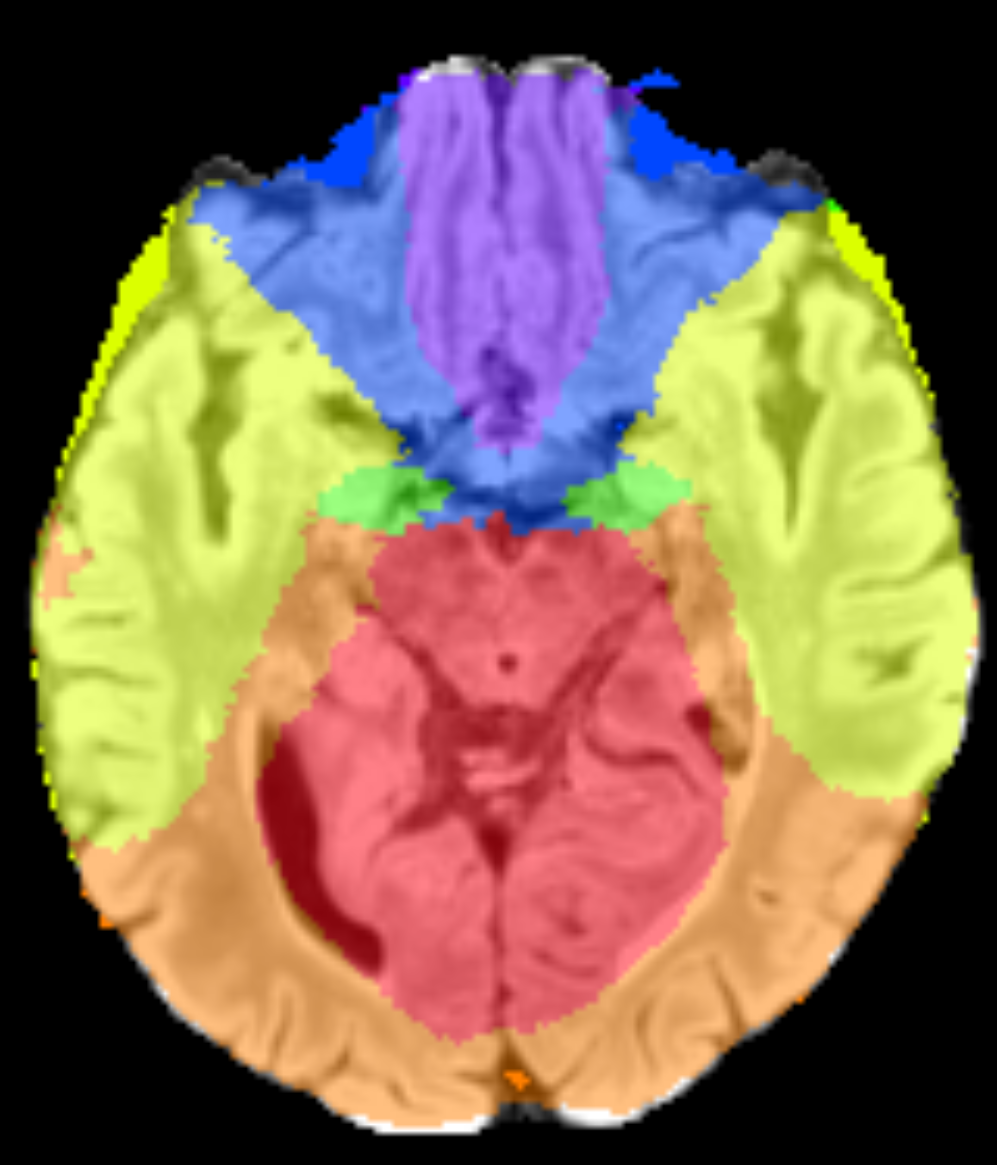}
         \caption{}
         \label{fig:BAT upper}
     \end{subfigure}
        \caption{Brain arterial territory mask registered to one of our test images, shown at two different axial slices. Purple: ACA, Blue: ACAMCA, Turqoise: ACAMCAPCA, Pale green (left image): ACAPCA, Lime green (right image): AChA, Yellow: MCA, Orange: MCAPCA, Red: PCA.}
        \label{fig:BAT}
\end{figure}

For the final ROI category, each \acrshort{isl} was allocated to one of eight arterial territories: the anterior (ACA), middle (MCA), and posterior (PCA) cerebral arteries, alongside their respective watershed borderzones (ACA-MCA, ACA-MCA-PCA, ACA-PCA, MChA, and MCA-PCA). Given the high inter-subject variability of arterial territories and the impossibility of determining exact vascular boundaries from structural \acrshort{mri} alone, we utilised a probabilistic arterial territory atlas. This atlas was constructed from manual delineations of 21 subjects presenting with \acrshort{isl} in clinically confirmed territories, adapting the methodology in \cite{liu2023digital} to explicitly include watershed regions. The atlas was non-linearly co-registered to each test subject's native space using NiftyReg \cite{modat2010fast} (illustrated in \textbf{Figure \ref{fig:BAT}}). Both ground truth and predicted \acrshort{isl} were assigned to the territory encompassing the majority of their respective voxels.

\paragraph{Assignment correction}

The initial spatial assignment rules are susceptible to misclassification caused by minor inaccuracies in the model's predictions. For instance, the model may successfully segment a subcortical lesion near the cortical boundary with high overlap, but erroneously extend the predicted mask slightly into the cortex. Under the strict initial intersection rule, this prediction would be allocated to the cortical ROI, while its corresponding ground truth remains in the subcortical ROI. To resolve these spatial mismatches, we implemented a correction mechanism based on prediction-to-ground-truth overlap. If the \acrlong{pre} of a predicted \acrshort{isl} relative to a ground truth \acrshort{isl} exceeds a threshold of 0.5, and no alternative prediction yields a higher \acrlong{pre} for that specific ground truth lesion, we infer that the model successfully targeted the subcortical ground truth. Consequently, the prediction's ROI assignment is overridden to match the subcortical ROI of its corresponding ground truth.

To formalise this correction, we framed the lesion matching as a bipartite matching problem. Predicted \acrshort{isl} components were optimally paired with ground truth \acrshort{isl} components using a linear sum assignment algorithm \cite{crouse2016implementing} (implemented via the SciPy library \cite{2020SciPy-NMeth}), utilising \acrlong{pre} as the edge weights. This process generates a mapping dictionary, $D$, linking predictions to their corresponding ground truth targets. \textbf{Algorithm \ref{alg:ISL}} is subsequently applied independently to each of the three ROI categories. It should be noted that, within the volume-based ROI group, if the number of predicted \acrshort{isl} exceeds the number of ground truth lesions, the (surplus) unassigned predictions remain present across all volume ROIs.

\begin{algorithm}[htbp]
\footnotesize
\caption{\acrshort{isl} assignment correction for one ROI group}\label{alg:ISL}
\begin{algorithmic}[1] 
    \Require Ground truth $\{Y_n\}_{n=1}^N$ and predicted $\{\hat{Y}_n\}_{n=1}^N$ binary masks for $N$ ROIs; ground truth $L_{gt}$ and predicted $L_{pred}$ global labelled images considering all \acrshort{isl}; dictionary $D$ mapping predicted labels to ground truth labels; precision threshold $\tau$.
    \State \textbf{Define} $Mask(L, l)$ as a function returning the binary mask of label $l$ from labelled image $L$.
    \State \textbf{Define} $Labels(L, M)$ as a function returning the set of unique, non-zero labels in $L$ that intersect with binary mask $M$.
    
    \For{$a = 1 \dots N$} \Comment{Loop over all ROIs}
        \State $\mathcal{L}_{gt\_other} \gets Labels(L_{gt}, \neg Y_a)$ \Comment{GT labels located outside current ROI $a$}
        \State $\mathcal{L}_{pred} \gets Labels(L_{pred}, \hat{Y}_a)$ \Comment{Predicted labels currently assigned to ROI $a$}
        
        \For{each $l_{pred} \in \mathcal{L}_{pred}$} \Comment{Check each prediction in current ROI}
            \If{$l_{pred} \in \text{keys}(D)$} \Comment{If prediction was successfully matched}
                \State $l_{gt} \gets D[l_{pred}]$ \Comment{Retrieve the matched GT label}
                
                \If{$l_{gt} \in \mathcal{L}_{gt\_other}$} \Comment{Mismatch: GT target belongs to a different ROI}
                    \State $v \gets Precision(Mask(L_{pred}, l_{pred}), Mask(L_{gt}, l_{gt}))$
                    
                    \If{$v > \tau$} \Comment{Precision above threshold (we set $\tau = 0.5$)}
                        \State $M \gets Mask(L_{pred}, l_{pred})$ \Comment{Binary mask of \acrshort{isl} to be reassigned}
                        \State $\hat{Y}_a \gets \hat{Y}_a \setminus M$ \Comment{Remove \acrshort{isl} from the incorrect pred mask}
                        
                        \For{$b = 1 \dots N$, where $b \neq a$} \Comment{Search other ROIs for the GT}
                            \State $\mathcal{L}_{gt\_target} \gets Labels(L_{gt}, Y_b)$ 
                            \If{$l_{gt} \in \mathcal{L}_{gt\_target}$} \Comment{Found the ROI containing the GT}
                                \State $\hat{Y}_b \gets \hat{Y}_b \cup M$ \Comment{Reassign \acrshort{isl} to the correct pred mask}
                            \EndIf
                        \EndFor
                        
                    \EndIf
                \EndIf
            \EndIf
        \EndFor
    \EndFor
    \State \Return $\{Y_n\}_{n=1}^N, \{\hat{Y}_n\}_{n=1}^N$
\end{algorithmic}
\normalsize
\end{algorithm}

\subsubsection{Metrics}

In addition to omitting the \acrshort{ap} metric, we excluded the scan-level false-positive (FP) metric (which measures the percentage of lesion-free scans exhibiting false-positive predictions), as these empty scans cannot be logically assigned to spatial ROIs. Given that the number and spatial location of \acrshort{isl} are of particular clinical relevance, this analysis prioritises lesion-level metrics. Specifically, we introduce the \acrfull{ldsc} and replace the \acrshort{avd} with \acrfull{lcwa}. The \acrshort{lcwa} metric specifically rewards predicting the correct number of lesions with high precision. It is computed by constructing a weighted bipartite graph between predicted and ground truth lesions---using \acrlong{pre} scores as edge weights---and subsequently dividing the mean precision across all predicted lesions by the total number of ground truth lesions. This metric yields a normalised score bounded between 0 and 1, where 1 indicates optimal performance.

\subsubsection{Results}

\textbf{Table \ref{tab:ISL ROIs}} summarises the quantitative outcomes of the ROI-based \acrshort{isl} analysis. The results indicate a pronounced performance degradation for low-volume lesions, which frequently escape detection entirely. Morphologically, these small lesions often manifest as punctate, disconnected satellite components surrounding a larger, successfully segmented primary infarct. Furthermore, the analysis reveals that cortical lesions are segmented with greater accuracy than their subcortical counterparts in our datasets. 

Regarding arterial distribution, \acrshort{isl} frequency is not consistent across vascular territories. Lesions within the MCA territory are the most prevalent and demonstrate the highest segmentation performance. This robust performance is likely driven both by the high representation of MCA lesions in the training distribution and the physiological reality that the MCA supplies a vast anatomical region, typically resulting in larger infarcts. Conversely, the model exhibits its poorest performance on infarcts located within the rare ACAMCAPCA and ACAPCA watershed borderzones.

\begin{table}[htbp]
\caption{Results for Ischaemic Stroke Lesion (ISL) segmentation of the pseudolabels model on different regions-of-interest (ROIs). Reported as mean {[}95\% CI{]}  with confidence intervals (CIs) calculated over a 10,000 sample bootstrap per ROI. N is the number of scans with at least one lesion in the ROI and hence the number of scans which most metric means are calculated over (except average surface distance (ASD) which can only be calculate for cases where there is a least one predicted voxel as well---for ASD the results are reported as mean \{N\} {[}95\% CI{]} if this number if different). DSC: Dice Similarity Coefficient, PRE: voxel-level precision, REC: voxel-level recall, ASD: average surface distance, LDSC: lesion-level DSC, LPRE: lesion-level PRE, LREC: lesion-level REC, LCWA: lesion count by weighted assignment.}
\label{tab:ISL ROIs}
\resizebox{\textwidth}{!}{%
\begin{tabular}{l|cccccccc}
\hline
ROI (N)           & DSC (\%)                                                            & PRE (\%)                                                            & REC (\%)                                                           & ASD (mm)                                                                   & LDSC (\%)                                                          & LPRE (\%)                                                             & LREC (\%)                                                          & LCWA (\%)                                                          \\ \hline
Overall (575)     & \begin{tabular}[c]{@{}c@{}}46.95\\[-5pt] \smalltext{{[}44.24, 49.58{]}}\end{tabular}  & \begin{tabular}[c]{@{}c@{}}69.51\\[-5pt] \smalltext{{[}66.84, 72.16{]}}\end{tabular}  & \begin{tabular}[c]{@{}c@{}}45.49\\[-5pt] \smalltext{{[}42.77, 48.24{]}}\end{tabular} & \begin{tabular}[c]{@{}c@{}}9.96 \{503\}\\[-5pt] \smalltext{{[}8.43, 11.55{]}}\end{tabular}   & \begin{tabular}[c]{@{}c@{}}50.47\\[-5pt] \smalltext{{[}47.56, 53.46{]}}\end{tabular} & \begin{tabular}[c]{@{}c@{}}70.60\\[-5pt] \smalltext{{[}67.62, 73.52{]}}\end{tabular}    & \begin{tabular}[c]{@{}c@{}}57.82\\[-5pt] \smalltext{{[}54.38, 61.08{]}}\end{tabular} & \begin{tabular}[c]{@{}c@{}}62.74\\[-5pt] \smalltext{{[}59.60, 65.87{]}}\end{tabular} \\ \hline
Low vol (159)     & \begin{tabular}[c]{@{}c@{}}1.18\\[-5pt] \smalltext{{[}0.15, 2.58{]}}\end{tabular}     & \begin{tabular}[c]{@{}c@{}}36.57\\[-5pt] \smalltext{{[}29.36, 44.08{]}}\end{tabular}  & \begin{tabular}[c]{@{}c@{}}2.65\\[-5pt] \smalltext{{[}1.18, 4.47{]}}\end{tabular}    & \begin{tabular}[c]{@{}c@{}}39.95 \{103\}\\[-5pt] \smalltext{{[}35.37, 44.68{]}}\end{tabular} & \begin{tabular}[c]{@{}c@{}}4.30\\[-5pt] \smalltext{{[}2.10, 6.82{]}}\end{tabular}    & \begin{tabular}[c]{@{}c@{}}40.20\\[-5pt] \smalltext{{[}32.86, 47.63{]}}\end{tabular}    & \begin{tabular}[c]{@{}c@{}}4.51\\[-5pt] \smalltext{{[}2.28, 7.15{]}}\end{tabular}    & \begin{tabular}[c]{@{}c@{}}36.60\\[-5pt] \smalltext{{[}30.53, 42.67{]}}\end{tabular} \\
Medium vol (288)  & \begin{tabular}[c]{@{}c@{}}14.98\\[-5pt] \smalltext{{[}12.04, 18.09{]}}\end{tabular}  & \begin{tabular}[c]{@{}c@{}}48.50\\[-5pt] \smalltext{{[}43.41, 53.81{]}}\end{tabular}  & \begin{tabular}[c]{@{}c@{}}17.05\\[-5pt] \smalltext{{[}13.79, 20.43{]}}\end{tabular} & \begin{tabular}[c]{@{}c@{}}27.00 \{203\}\\[-5pt] \smalltext{{[}23.49, 30.58{]}}\end{tabular} & \begin{tabular}[c]{@{}c@{}}25.20\\[-5pt] \smalltext{{[}20.91, 29.61{]}}\end{tabular} & \begin{tabular}[c]{@{}c@{}}56.17\\[-5pt] \smalltext{{[}50.91, 61.66{]}}\end{tabular}    & \begin{tabular}[c]{@{}c@{}}28.81\\[-5pt] \smalltext{{[}23.90, 33.77{]}}\end{tabular} & \begin{tabular}[c]{@{}c@{}}52.73\\[-5pt] \smalltext{{[}47.73, 57.69{]}}\end{tabular} \\
High vol (408)    & \begin{tabular}[c]{@{}c@{}}45.99\\[-5pt] \smalltext{{[}42.73, 49.34{]}}\end{tabular}  & \begin{tabular}[c]{@{}c@{}}68.03\\[-5pt] \smalltext{{[}64.67, 71.31{]}}\end{tabular}  & \begin{tabular}[c]{@{}c@{}}44.14\\[-5pt] \smalltext{{[}40.87, 47.51{]}}\end{tabular} & \begin{tabular}[c]{@{}c@{}}10.08 \{363\}\\[-5pt] \smalltext{{[}8.35, 11.91{]}}\end{tabular}  & \begin{tabular}[c]{@{}c@{}}52.10\\[-5pt] \smalltext{{[}48.46, 55.79{]}}\end{tabular} & \begin{tabular}[c]{@{}c@{}}62.86\\[-5pt] \smalltext{{[}59.06, 66.69{]}}\end{tabular}    & \begin{tabular}[c]{@{}c@{}}65.15\\[-5pt] \smalltext{{[}61.07, 69.18{]}}\end{tabular} & \begin{tabular}[c]{@{}c@{}}76.11\\[-5pt] \smalltext{{[}72.64, 79.57{]}}\end{tabular} \\ \hline
Cortical (362)    & \begin{tabular}[c]{@{}c@{}}48.31 \\[-5pt] \smalltext{{[}44.83, 51.78{]}}\end{tabular} & \begin{tabular}[c]{@{}c@{}}76.13\\[-5pt] \smalltext{{[}73.17, 79.01{]}}\end{tabular}  & \begin{tabular}[c]{@{}c@{}}46.46\\[-5pt] \smalltext{{[}42.97, 49.89{]}}\end{tabular} & \begin{tabular}[c]{@{}c@{}}8.35 \{301\}\\[-5pt] \smalltext{{[}6.40, 10.51{]}}\end{tabular}   & \begin{tabular}[c]{@{}c@{}}54.06\\[-5pt] \smalltext{{[}50.13, 57.87{]}}\end{tabular} & \begin{tabular}[c]{@{}c@{}}78.72\\[-5pt] \smalltext{{[}75.35, 82.01{]}}\end{tabular}    & \begin{tabular}[c]{@{}c@{}}57.59\\[-5pt] \smalltext{{[}53.16, 61.81{]}}\end{tabular} & \begin{tabular}[c]{@{}c@{}}60.39\\[-5pt] \smalltext{{[}56.26, 64.55{]}}\end{tabular} \\
Subcortical (367) & \begin{tabular}[c]{@{}c@{}}31.34\\[-5pt] \smalltext{{[}27.94, 34.86{]}}\end{tabular}  & \begin{tabular}[c]{@{}c@{}}71.82\\[-5pt] \smalltext{{[}67.95, 75.40{]}}\end{tabular}  & \begin{tabular}[c]{@{}c@{}}29.72\\[-5pt] \smalltext{{[}26.33, 33.21{]}}\end{tabular} & \begin{tabular}[c]{@{}c@{}}11.53 \{250\}\\[-5pt] \smalltext{{[}9.14, 14.03{]}}\end{tabular}  & \begin{tabular}[c]{@{}c@{}}41.76\\[-5pt] \smalltext{{[}37.41, 46.23{]}}\end{tabular} & \begin{tabular}[c]{@{}c@{}}75.32\\[-5pt] \smalltext{{[}71.23, 79.25{]}}\end{tabular}    & \begin{tabular}[c]{@{}c@{}}45.02\\[-5pt] \smalltext{{[}40.39, 49.88{]}}\end{tabular} & \begin{tabular}[c]{@{}c@{}}56.73\\[-5pt] \smalltext{{[}52.26, 61.27{]}}\end{tabular} \\ \hline
ACA (27)          & \begin{tabular}[c]{@{}c@{}}20.22\\[-5pt] \smalltext{{[}9.06, 32.61{]}}\end{tabular}   & \begin{tabular}[c]{@{}c@{}}80.99\\[-5pt] \smalltext{{[}67.50, 92.04{]}}\end{tabular}  & \begin{tabular}[c]{@{}c@{}}19.18\\[-5pt] \smalltext{{[}8.06, 32.00{]}}\end{tabular}  & \begin{tabular}[c]{@{}c@{}}8.84 \{13\}\\[-5pt] \smalltext{{[}1.13, 17.96{]}}\end{tabular}    & \begin{tabular}[c]{@{}c@{}}30.19\\[-5pt] \smalltext{{[}14.44, 46.85{]}}\end{tabular} & \begin{tabular}[c]{@{}c@{}}86.11\\[-5pt] \smalltext{{[}72.22, 97.22{]}}\end{tabular}    & \begin{tabular}[c]{@{}c@{}}31.39\\[-5pt] \smalltext{{[}15.34, 48.68{]}}\end{tabular} & \begin{tabular}[c]{@{}c@{}}40.64\\[-5pt] \smalltext{{[}23.45, 58.63{]}}\end{tabular} \\
ACAMCA (121)      & \begin{tabular}[c]{@{}c@{}}27.81\\[-5pt] \smalltext{{[}21.64, 34.24{]}}\end{tabular}  & \begin{tabular}[c]{@{}c@{}}84.88\\[-5pt] \smalltext{{[}79.80, 89.59{]}}\end{tabular}  & \begin{tabular}[c]{@{}c@{}}26.99\\[-5pt] \smalltext{{[}20.81, 33.31{]}}\end{tabular} & \begin{tabular}[c]{@{}c@{}}6.23 \{62\}\\[-5pt] \smalltext{{[}3.06, 9.97{]}}\end{tabular}     & \begin{tabular}[c]{@{}c@{}}38.40\\[-5pt] \smalltext{{[}30.60, 46.40{]}}\end{tabular} & \begin{tabular}[c]{@{}c@{}}90.36\\[-5pt] \smalltext{{[}85.40, 94.63{]}}\end{tabular}    & \begin{tabular}[c]{@{}c@{}}38.93\\[-5pt] \smalltext{{[}30.89, 47.11{]}}\end{tabular} & \begin{tabular}[c]{@{}c@{}}41.39\\[-5pt] \smalltext{{[}33.45, 49.56{]}}\end{tabular} \\
ACAMCAPCA (6)     & \begin{tabular}[c]{@{}c@{}}13.87\\[-5pt] \smalltext{{[}0.00, 41.61{]}}\end{tabular}   & \begin{tabular}[c]{@{}c@{}}78.80\\[-5pt] \smalltext{{[}45.47, 100.00{]}}\end{tabular} & \begin{tabular}[c]{@{}c@{}}16.18\\[-5pt] \smalltext{{[}0.00, 48.55{]}}\end{tabular}  & \begin{tabular}[c]{@{}c@{}}7.49 \{2\}\\[-5pt] \smalltext{{[}0.38, 14.60{]}}\end{tabular}     & \begin{tabular}[c]{@{}c@{}}16.67\\[-5pt] \smalltext{{[}0.00, 50.00{]}}\end{tabular}  & \begin{tabular}[c]{@{}c@{}}83.33\\[-5pt] \smalltext{{[}50.00, 100.00{]}}\end{tabular}   & \begin{tabular}[c]{@{}c@{}}16.67\\[-5pt] \smalltext{{[}0.00, 50.00{]}}\end{tabular}  & \begin{tabular}[c]{@{}c@{}}33.33\\[-5pt] \smalltext{{[}0.00, 66.67{]}}\end{tabular}  \\
ACAPCA (10)       & \begin{tabular}[c]{@{}c@{}}15.36\\[-5pt] \smalltext{{[}0.00, 34.13{]}}\end{tabular}   & \begin{tabular}[c]{@{}c@{}}89.50\\[-5pt] \smalltext{{[}76.39, 100.00{]}}\end{tabular} & \begin{tabular}[c]{@{}c@{}}13.28\\[-5pt] \smalltext{{[}0.00, 30.63{]}}\end{tabular}  & \begin{tabular}[c]{@{}c@{}}1.45 \{3\}\\[-5pt] \smalltext{{[}0.57, 2.64{]}}\end{tabular}      & \begin{tabular}[c]{@{}c@{}}28.00\\[-5pt] \smalltext{{[}0.00, 56.00{]}}\end{tabular}  & \begin{tabular}[c]{@{}c@{}}100.00\\[-5pt] \smalltext{{[}100.00, 100.00{]}}\end{tabular} & \begin{tabular}[c]{@{}c@{}}26.67\\[-5pt] \smalltext{{[}0.00, 53.33{]}}\end{tabular}  & \begin{tabular}[c]{@{}c@{}}25.54\\[-5pt] \smalltext{{[}0.00, 51.09{]}}\end{tabular}  \\
AChA (25)         & \begin{tabular}[c]{@{}c@{}}27.76\\[-5pt] \smalltext{{[}14.15, 41.86{]}}\end{tabular}  & \begin{tabular}[c]{@{}c@{}}90.58\\[-5pt] \smalltext{{[}84.44, 95.88{]}}\end{tabular}  & \begin{tabular}[c]{@{}c@{}}26.87\\[-5pt] \smalltext{{[}13.43, 40.91{]}}\end{tabular} & \begin{tabular}[c]{@{}c@{}}1.00 \{10\}\\[-5pt] \smalltext{{[}0.63, 1.40{]}}\end{tabular}     & \begin{tabular}[c]{@{}c@{}}40.00\\[-5pt] \smalltext{{[}20.00, 60.00{]}}\end{tabular} & \begin{tabular}[c]{@{}c@{}}100.00\\[-5pt] \smalltext{{[}100.00, 100.00{]}}\end{tabular} & \begin{tabular}[c]{@{}c@{}}40.00\\[-5pt] \smalltext{{[}20.00, 60.00{]}}\end{tabular} & \begin{tabular}[c]{@{}c@{}}40.00\\[-5pt] \smalltext{{[}20.00, 60.00{]}}\end{tabular} \\
MCA (338)         & \begin{tabular}[c]{@{}c@{}}44.71\\[-5pt] \smalltext{{[}40.96, 48.42{]}}\end{tabular}  & \begin{tabular}[c]{@{}c@{}}76.63\\[-5pt] \smalltext{{[}73.43, 79.74{]}}\end{tabular}  & \begin{tabular}[c]{@{}c@{}}43.59\\[-5pt] \smalltext{{[}39.86, 47.37{]}}\end{tabular} & \begin{tabular}[c]{@{}c@{}}5.69 \{266\}\\[-5pt] \smalltext{{[}4.33, 7.20{]}}\end{tabular}    & \begin{tabular}[c]{@{}c@{}}53.38\\[-5pt] \smalltext{{[}50.15, 58.69{]}}\end{tabular} & \begin{tabular}[c]{@{}c@{}}82.23\\[-5pt] \smalltext{{[}78.80, 85.65{]}}\end{tabular}    & \begin{tabular}[c]{@{}c@{}}57.14\\[-5pt] \smalltext{{[}52.52, 61.72{]}}\end{tabular} & \begin{tabular}[c]{@{}c@{}}60.49\\[-5pt] \smalltext{{[}56.03, 64.88{]}}\end{tabular} \\
MCAPCA (126)      & \begin{tabular}[c]{@{}c@{}}25.55\\[-5pt] \smalltext{{[}19.93, 31.30{]}}\end{tabular}  & \begin{tabular}[c]{@{}c@{}}81.09\\[-5pt] \smalltext{{[}75.31, 86.39{]}}\end{tabular}  & \begin{tabular}[c]{@{}c@{}}24.42\\[-5pt] \smalltext{{[}18.95, 30.07{]}}\end{tabular} & \begin{tabular}[c]{@{}c@{}}6.57 \{71\}\\[-5pt] \smalltext{{[}3.91, 9.71{]}}\end{tabular}     & \begin{tabular}[c]{@{}c@{}}37.85\\[-5pt] \smalltext{{[}30.33, 45.24{]}}\end{tabular} & \begin{tabular}[c]{@{}c@{}}88.17\\[-5pt] \smalltext{{[}82.71, 93.06{]}}\end{tabular}    & \begin{tabular}[c]{@{}c@{}}37.56\\[-5pt] \smalltext{{[}29.95, 45.14{]}}\end{tabular} & \begin{tabular}[c]{@{}c@{}}42.88\\[-5pt] \smalltext{{[}34.97, 50.61{]}}\end{tabular} \\
PCA (157)         & \begin{tabular}[c]{@{}c@{}}40.02\\[-5pt] \smalltext{{[}34.64, 45.52{]}}\end{tabular}  & \begin{tabular}[c]{@{}c@{}}83.16\\[-5pt] \smalltext{{[}79.06, 87.02{]}}\end{tabular}  & \begin{tabular}[c]{@{}c@{}}36.72\\[-5pt] \smalltext{{[}31.57, 41.97{]}}\end{tabular} & \begin{tabular}[c]{@{}c@{}}5.29 \{104\}\\[-5pt] \smalltext{{[}2.83, 8.10{]}}\end{tabular}    & \begin{tabular}[c]{@{}c@{}}50.47\\[-5pt] \smalltext{{[}43.38, 57.47{]}}\end{tabular} & \begin{tabular}[c]{@{}c@{}}88.21\\[-5pt] \smalltext{{[}83.65, 92.38{]}}\end{tabular}    & \begin{tabular}[c]{@{}c@{}}50.92\\[-5pt] \smalltext{{[}43.61, 58.04{]}}\end{tabular} & \begin{tabular}[c]{@{}c@{}}55.20\\[-5pt] \smalltext{{[}48.06, 62.11{]}}\end{tabular} \\ \hline
\end{tabular}%
}
\end{table}

\clearpage

\section{Probabilistic lesion maps} 

All FLAIR volumes and their corresponding label maps were non-linearly co-registered to the ICBM 152 reference template \cite{fonov2009unbiased,fonov2011unbiased} utilising NiftyReg \cite{modat2010fast}. Voxel-wise spatial probability distributions were subsequently generated by computing the mean occurrence across all registered masks. This procedure was applied to derive spatial probability maps for the ground truth and predicted segmentations, as well as the spatial distributions of false-positive and false-negative predictions. To maximise visual contrast, colourmaps were normalised independently for each rendered volume; consequently, the accompanying colour bars must be referenced to interpret the absolute probability values.

\subsection{Cohort-level ground truth distributions} \label{appendix B gt}

The following probabilistic maps were computed across the entirety of each respective cohort (encompassing the training, validation, and test partitions) to visually characterise the distinct spatial disease phenotypes inherent to each dataset.

% MSS1
\begin{figure}[htbp]
     \centering
     \begin{subfigure}[]{1.0\textwidth}
         \centering
         \includegraphics[width=\textwidth]{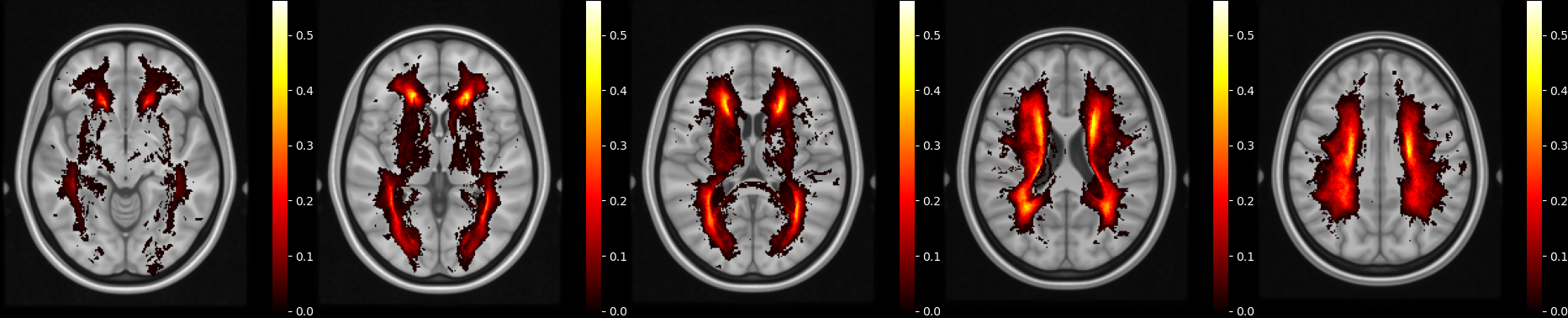}
         \caption{White Matter Hyperintensities (WMH)}
         \label{fig:mss1_prob_maps_full_wmh_gt}
     \end{subfigure}
     \hfill
     \begin{subfigure}[]{1.0\textwidth}
         \centering
         \includegraphics[width=\textwidth]{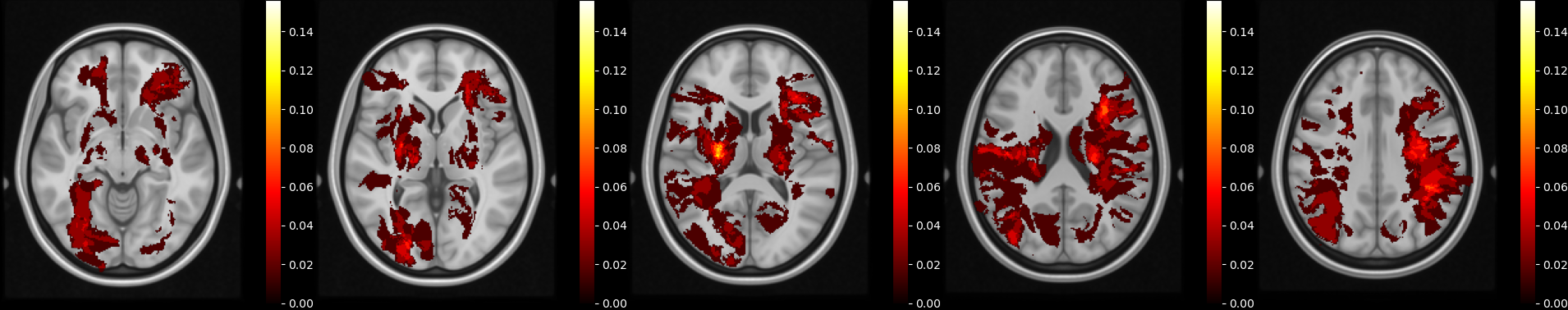}
         \caption{Ischaemic Stroke Lesions (ISL)}
         \label{fig:mss1_prob_maps_full_isl_gt}
    \end{subfigure}
    \caption{MSS1 lesions distributions (slices: 140, 150, 160, 170, 180)}
    \label{fig:mss1_prob_maps_full_gt}
\end{figure}

% MSS2
\begin{figure}[htbp]
     \centering
     \begin{subfigure}[]{1.0\textwidth}
         \centering
         \includegraphics[width=\textwidth]{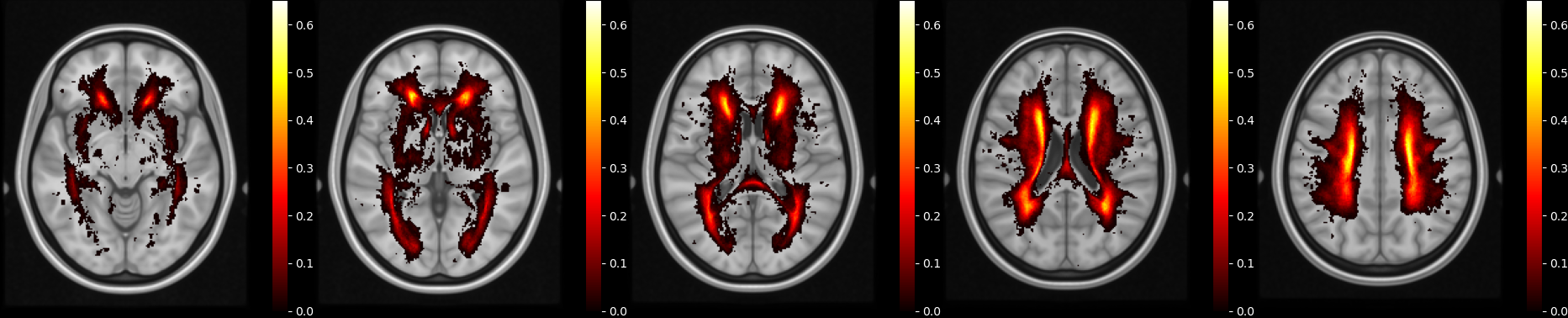}
         \caption{White Matter Hyperintensities (WMH)}
         \label{fig:mss2_prob_maps_full_wmh_gt}
     \end{subfigure}
     \hfill
     \begin{subfigure}[]{1.0\textwidth}
         \centering
         \includegraphics[width=\textwidth]{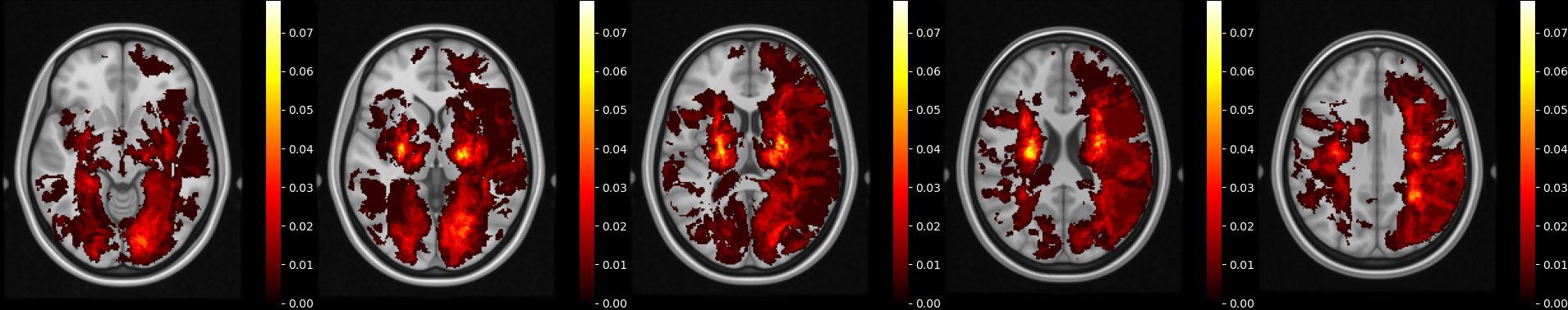}
         \caption{Ischaemic Stroke Lesions (ISL)}
         \label{fig:mss2_prob_maps_full_isl_gt}
    \end{subfigure}
    \caption{MSS2 lesions distributions (slices: 140, 150, 160, 170, 180)}
    \label{fig:mss2_prob_maps_full_gt}
\end{figure}

% MSS3
\begin{figure}[htbp]
     \centering
     \begin{subfigure}[]{1.0\textwidth}
         \centering
         \includegraphics[width=\textwidth]{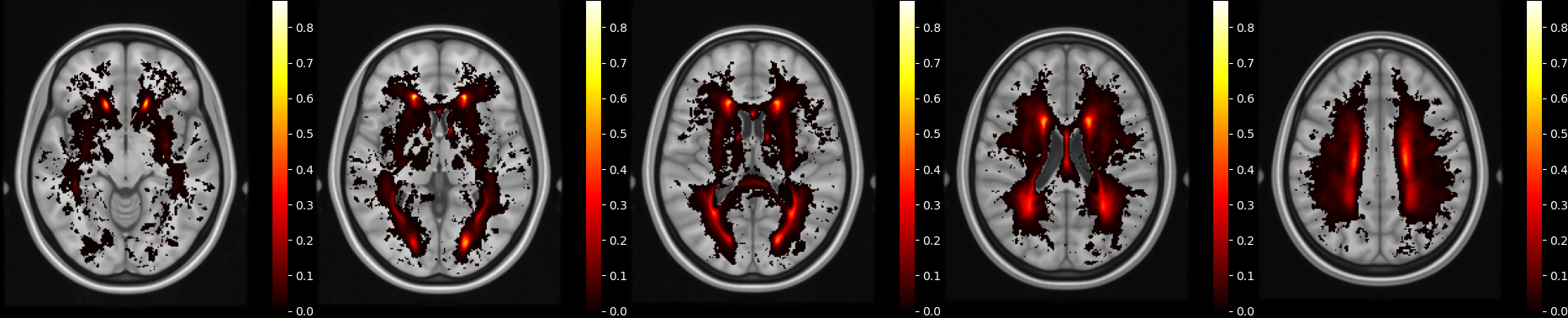}
         \caption{White Matter Hyperintensities (WMH)}
         \label{fig:mss2_prob_maps_full_wmh_gt}
     \end{subfigure}
     \hfill
     \begin{subfigure}[]{1.0\textwidth}
         \centering
         \includegraphics[width=\textwidth]{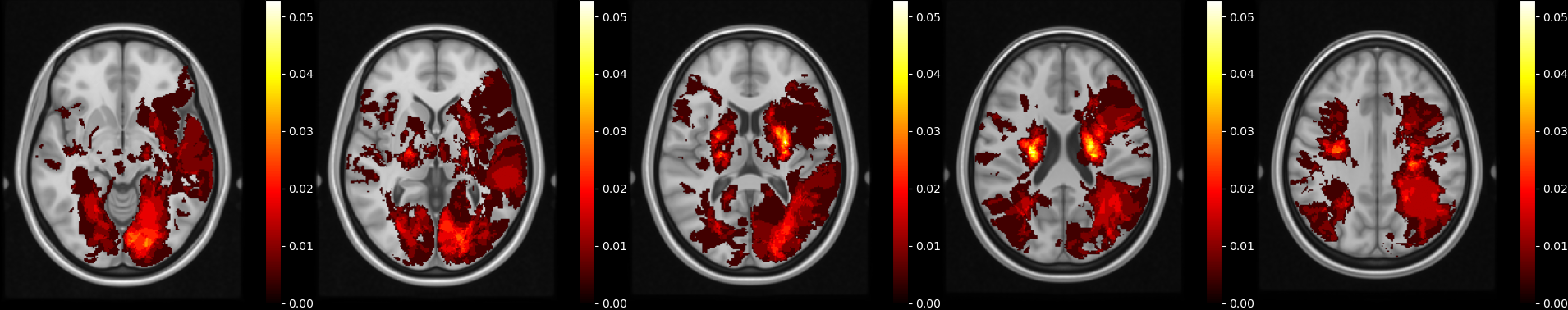}
         \caption{Ischaemic Stroke Lesions (ISL)}
         \label{fig:mss2_prob_maps_full_isl_gt}
    \end{subfigure}
    \caption{MSS3 lesions distributions (slices: 140, 150, 160, 170, 180)}
    \label{fig:mss2_prob_maps_full_gt}
\end{figure}

% LBC1936
\begin{figure}[htbp]
     \centering
     \begin{subfigure}[]{1.0\textwidth}
         \centering
         \includegraphics[width=\textwidth]{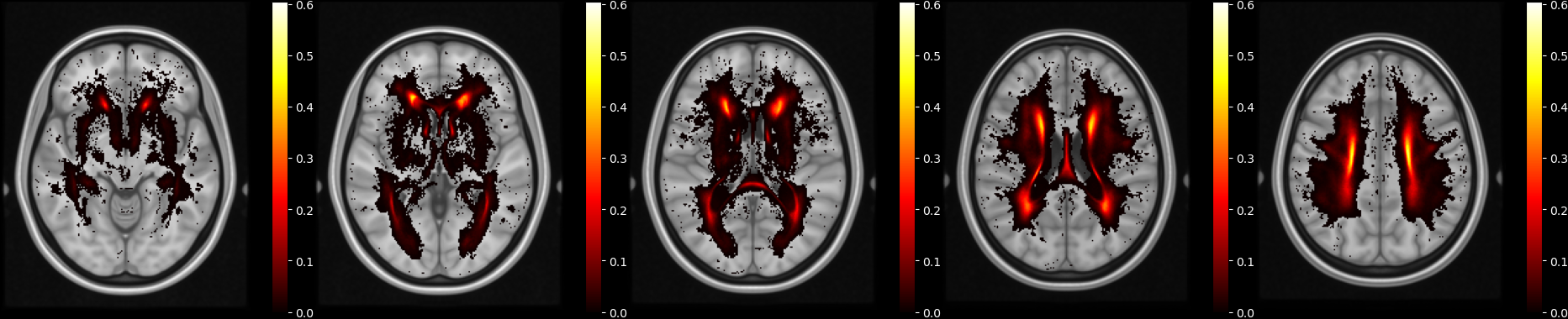}
         \caption{White Matter Hyperintensities (WMH)}
         \label{fig:lbc1936_prob_maps_full_wmh_gt}
     \end{subfigure}
     \hfill
     \begin{subfigure}[]{1.0\textwidth}
         \centering
         \includegraphics[width=\textwidth]{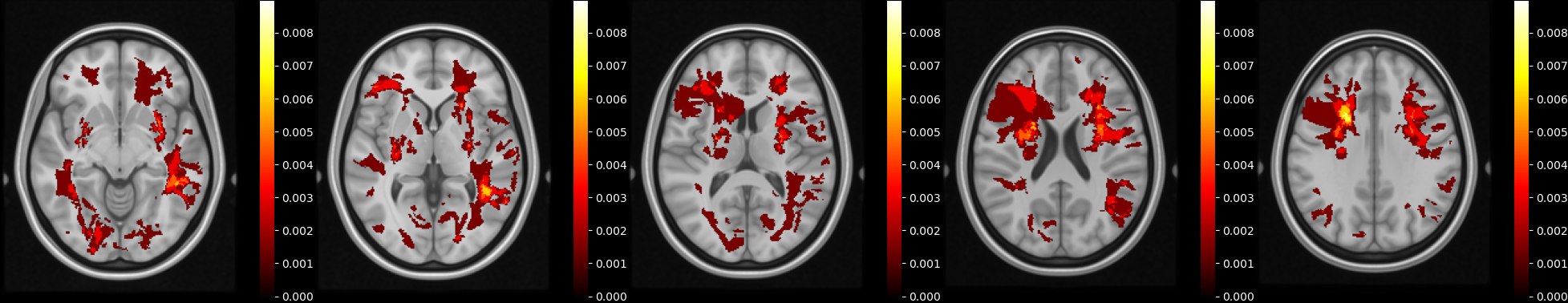}
         \caption{Ischaemic Stroke Lesions (ISL)}
         \label{fig:lbc1936_prob_maps_full_isl_gt}
    \end{subfigure}
    \caption{LBC1936 lesions distributions (slices: 140, 150, 160, 170, 180)}
    \label{fig:lbc1936_prob_maps_full_gt}
\end{figure}

% LBC1921
\begin{figure}[htbp]
     \centering
     \begin{subfigure}[]{1.0\textwidth}
         \centering
         \includegraphics[width=\textwidth]{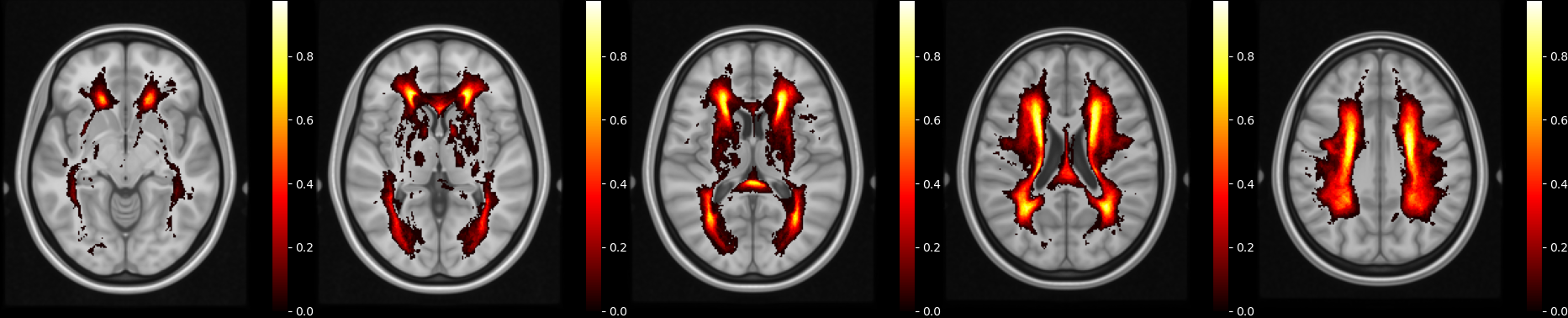}
         \caption{White Matter Hyperintensities (WMH)}
         \label{fig:lbc19321_prob_maps_full_wmh_gt}
     \end{subfigure}
     \hfill
     \begin{subfigure}[]{1.0\textwidth}
         \centering
         \includegraphics[width=\textwidth]{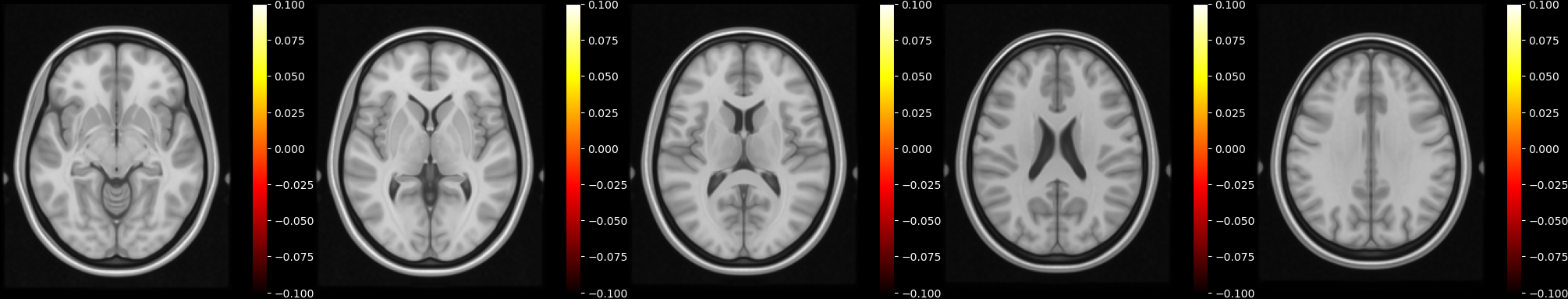}
         \caption{Ischaemic Stroke Lesions (ISL) (NO GROUND TRUTH)}
         \label{fig:lbc19321_prob_maps_full_isl_gt}
    \end{subfigure}
    \caption{LBC1921 lesions distributions (slices: 140, 150, 160, 170, 180)}
    \label{fig:lbc19321_prob_maps_full_gt}
\end{figure}

% WMH-ch
\begin{figure}[htbp]
     \centering
     \begin{subfigure}[]{1.0\textwidth}
         \centering
         \includegraphics[width=\textwidth]{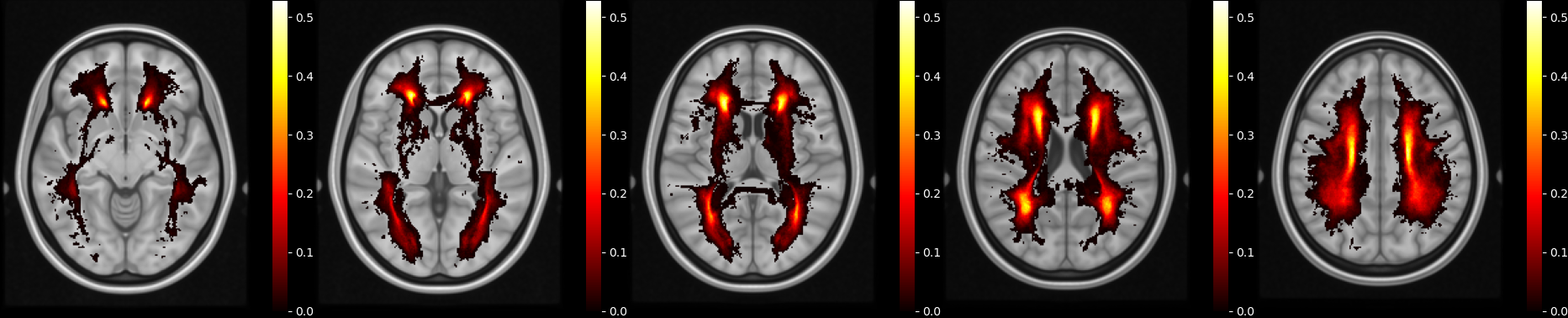}
         \caption{White Matter Hyperintensities (WMH)}
         \label{fig:wmh_prob_maps_full_wmh_gt}
     \end{subfigure}
     \hfill
     \begin{subfigure}[]{1.0\textwidth}
         \centering
         \includegraphics[width=\textwidth]{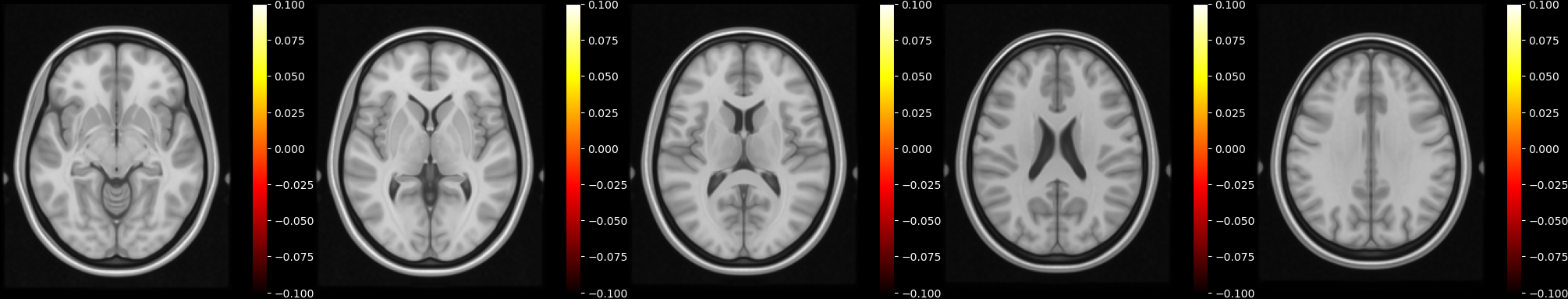}
         \caption{Ischaemic Stroke Lesions (ISL) (NO GROUND TRUTH)}
         \label{fig:wmh_prob_maps_full_isl_gt}
    \end{subfigure}
    \caption{WMH-ch lesions distributions (slices: 140, 150, 160, 170, 180)}
    \label{fig:wmh_prob_maps_full_gt}
\end{figure}

% ISLES
\begin{figure}[htbp]
     \centering
     \begin{subfigure}[]{1.0\textwidth}
         \centering
         \includegraphics[width=\textwidth]{f_a_p_None_LBC1921_ISL_gt_slc_140-180.png}
         \caption{White Matter Hyperintensities (WMH) (NO GROUND TRUTH)}
         \label{fig:isles_prob_maps_full_wmh_gt}
     \end{subfigure}
     \hfill
     \begin{subfigure}[]{1.0\textwidth}
         \centering
         \includegraphics[width=\textwidth]{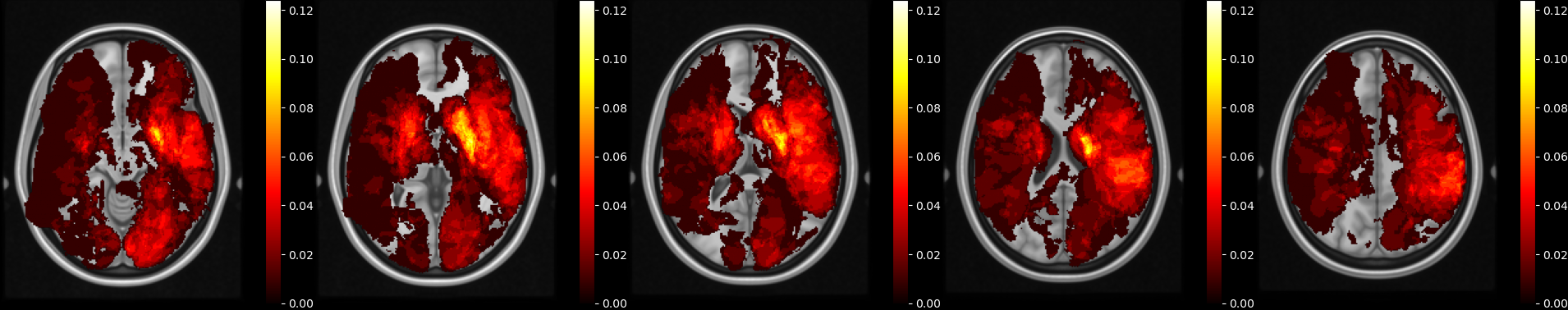}
         \caption{Ischaemic Stroke Lesions (ISL)}
         \label{fig:isles_prob_maps_full_isl_gt}
    \end{subfigure}
    \caption{ISLES lesions distributions (slices: 140, 150, 160, 170, 180)}
    \label{fig:isles_prob_maps_full_gt}
\end{figure}

% SOOP
\begin{figure}[htbp]
     \centering
     \begin{subfigure}[]{1.0\textwidth}
         \centering
         \includegraphics[width=\textwidth]{f_a_p_None_LBC1921_ISL_gt_slc_140-180.png}
         \caption{White Matter Hyperintensities (WMH) (NO GROUND TRUTH)}
         \label{fig:soop_prob_maps_full_wmh_gt}
     \end{subfigure}
     \hfill
     \begin{subfigure}[]{1.0\textwidth}
         \centering
         \includegraphics[width=\textwidth]{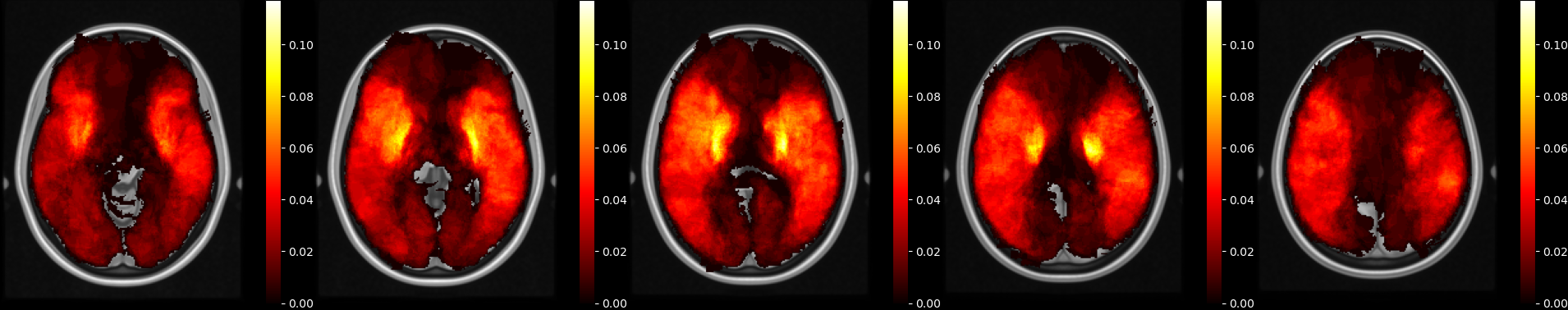}
         \caption{Ischaemic Stroke Lesions (ISL)}
         \label{fig:soop_prob_maps_full_isl_gt}
    \end{subfigure}
    \caption{SOOP lesions distributions (slices: 140, 150, 160, 170, 180)}
    \label{fig:soop_prob_maps_full_gt}
\end{figure}

% WSS
\begin{figure}[htbp]
     \centering
     \begin{subfigure}[]{1.0\textwidth}
         \centering
         \includegraphics[width=\textwidth]{f_a_p_None_LBC1921_ISL_gt_slc_140-180.png}
         \caption{White Matter Hyperintensities (WMH) (NO GROUND TRUTH)}
         \label{fig:wss_prob_maps_full_wmh_gt}
     \end{subfigure}
     \hfill
     \begin{subfigure}[]{1.0\textwidth}
         \centering
         \includegraphics[width=\textwidth]{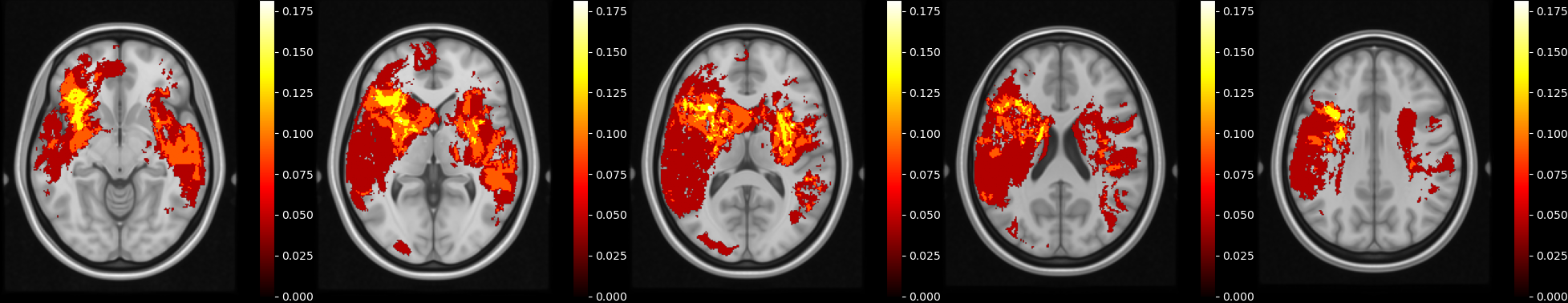}
         \caption{Ischaemic Stroke Lesions (ISL)}
         \label{fig:wss_prob_maps_full_isl_gt}
    \end{subfigure}
    \caption{WSS lesions distributions (slices: 140, 150, 160, 170, 180)}
    \label{fig:wss_prob_maps_full_gt}
\end{figure}

% ESS
\begin{figure}[htbp]
     \centering
     \begin{subfigure}[]{1.0\textwidth}
         \centering
         \includegraphics[width=\textwidth]{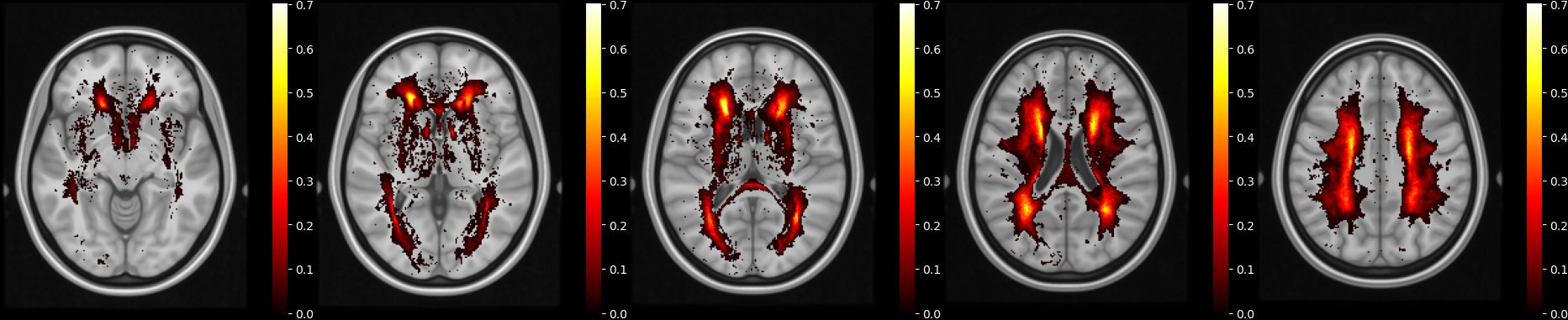}
         \caption{White Matter Hyperintensities (WMH)}
         \label{fig:ess_prob_maps_full_wmh_gt}
     \end{subfigure}
     \hfill
     \begin{subfigure}[]{1.0\textwidth}
         \centering
         \includegraphics[width=\textwidth]{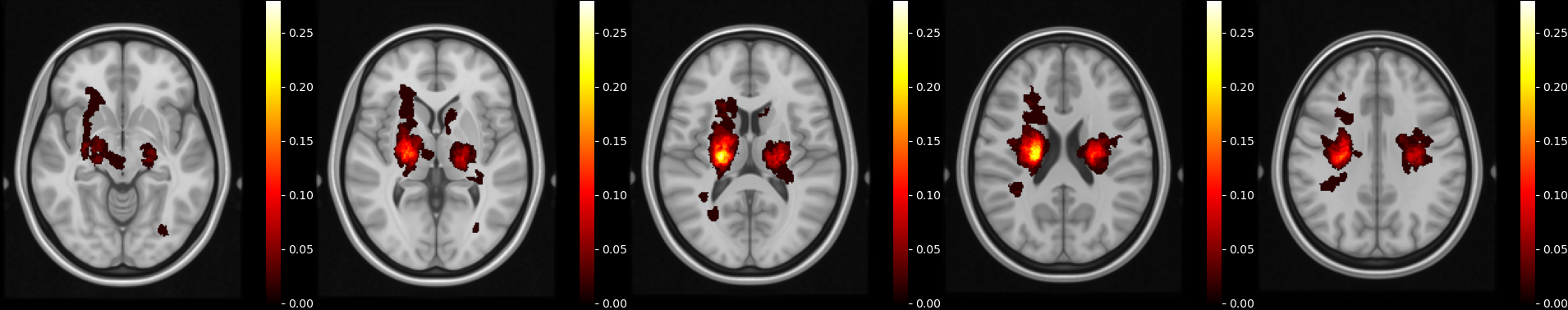}
         \caption{Ischaemic Stroke Lesions (ISL)}
         \label{fig:ess_prob_maps_full_isl_gt}
    \end{subfigure}
    \caption{ESS lesions distributions (slices: 140, 150, 160, 170, 180)}
    \label{fig:ess_prob_maps_full_gt}
\end{figure}

\clearpage

\subsection{Ground truth and predicted distributions in the test set} \label{appendix B fp fn}

The probabilistic distributions presented in this section are derived exclusively from the test set, with the objective of visually characterising systematic spatial discrepancies between the ground truth annotations and the pseudolabel-trained model predictions. Alongside the probabilistic maps of the ground truth and predicted segmentations, we provide the voxel-wise spatial distributions of both false-positive and false-negative predictions. Readers are advised to rigorously consult the absolute scale bars when interpreting these error maps; because volumetric segmentation errors are inherently sparse, regions of high relative intensity on the colourmap may represent only a small absolute number of misclassified voxels aggregated across the entire test cohort.

% MSS1
\begin{figure}[htbp]
     \centering
     \begin{subfigure}[]{1.0\textwidth}
         \centering
         \includegraphics[width=\textwidth]{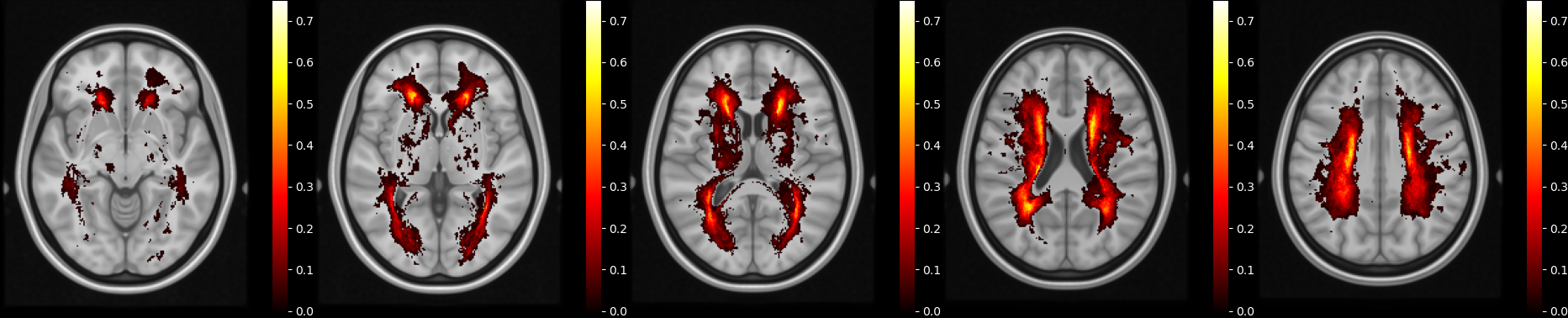}
         \caption{Ground truth}
         \label{fig:mss1_prob_maps_test_wmh_gt}
     \end{subfigure}
     \hfill
     \begin{subfigure}[]{1.0\textwidth}
         \centering
         \includegraphics[width=\textwidth]{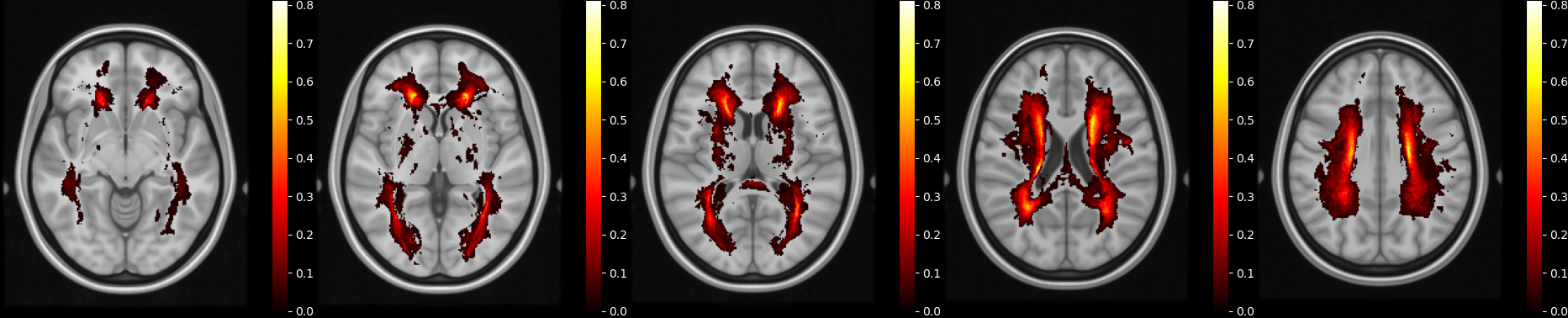}
         \caption{Predicted}
         \label{fig:mss1_prob_maps_test_wmh_pred}
    \end{subfigure}
    \hfill
    \begin{subfigure}[]{1.0\textwidth}
         \centering
         \includegraphics[width=\textwidth]{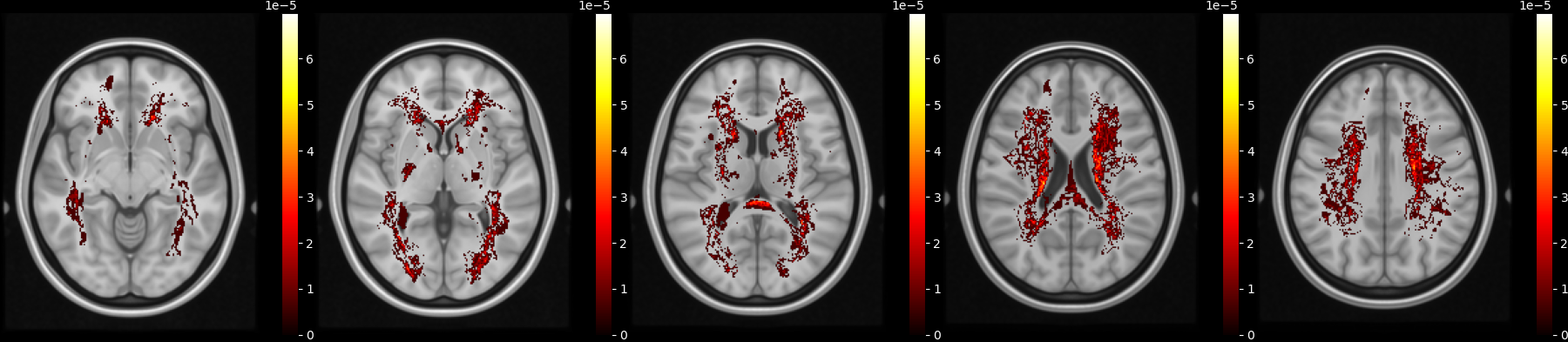}
         \caption{False-positives}
         \label{fig:mss1_prob_maps_test_wmh_fp}
    \end{subfigure}
    \hfill
    \begin{subfigure}[]{1.0\textwidth}
         \centering
         \includegraphics[width=\textwidth]{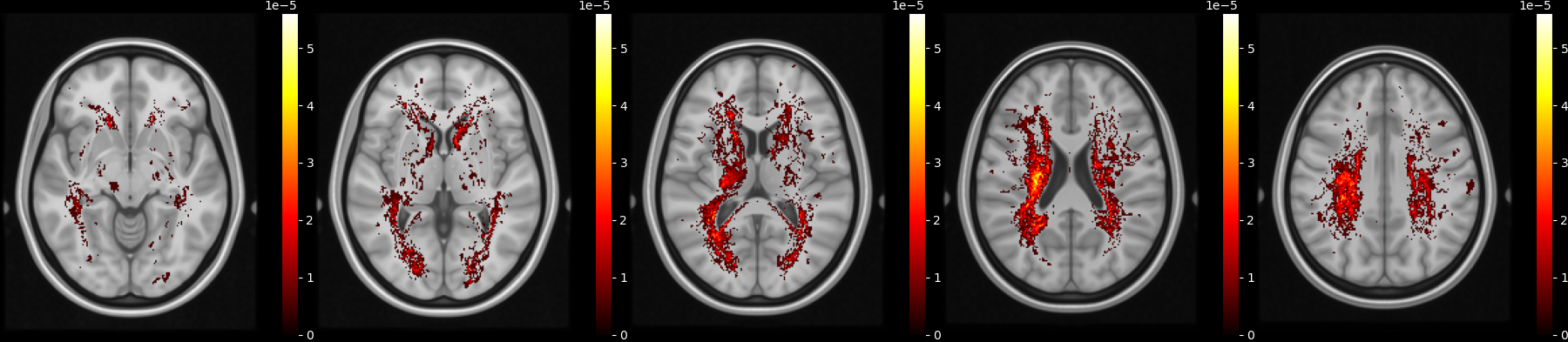}
         \caption{False-negatives}
         \label{fig:mss1_prob_maps_test_wmh_fn}
    \end{subfigure}
    \caption{MSS1 White Matter Hyperintensities (WMH) distributions (slices: 140, 150, 160, 170, 180)}
    \label{fig:mss1_prob_maps_full_gt}
\end{figure}

% MSS1
\begin{figure}[htbp]
     \centering
     \begin{subfigure}[]{1.0\textwidth}
         \centering
         \includegraphics[width=\textwidth]{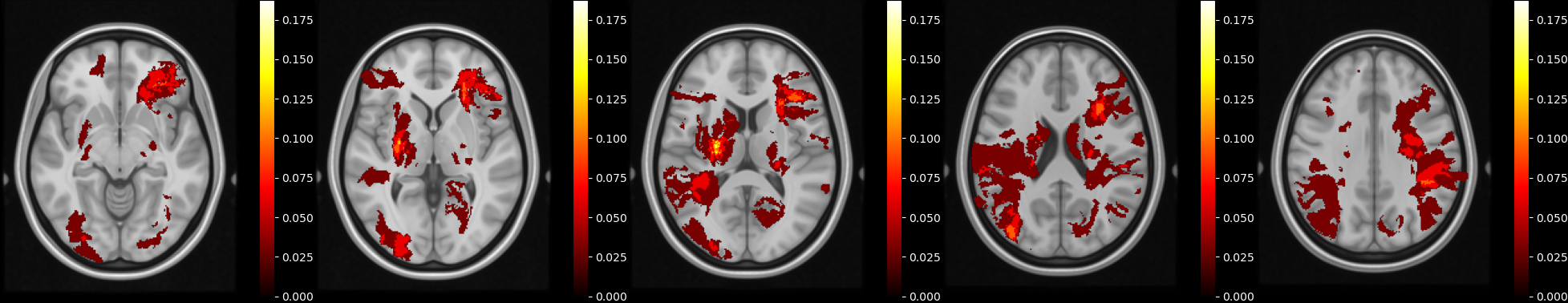}
         \caption{Ground truth}
         \label{fig:mss1_prob_maps_test_isl_gt}
     \end{subfigure}
     \hfill
     \begin{subfigure}[]{1.0\textwidth}
         \centering
         \includegraphics[width=\textwidth]{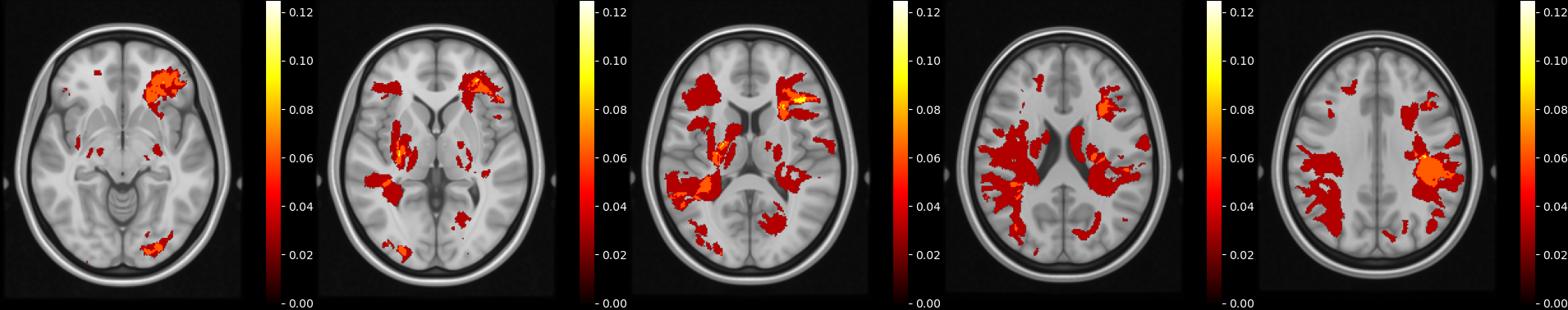}
         \caption{Predicted}
         \label{fig:mss1_prob_maps_test_isl_pred}
    \end{subfigure}
    \hfill
    \begin{subfigure}[]{1.0\textwidth}
         \centering
         \includegraphics[width=\textwidth]{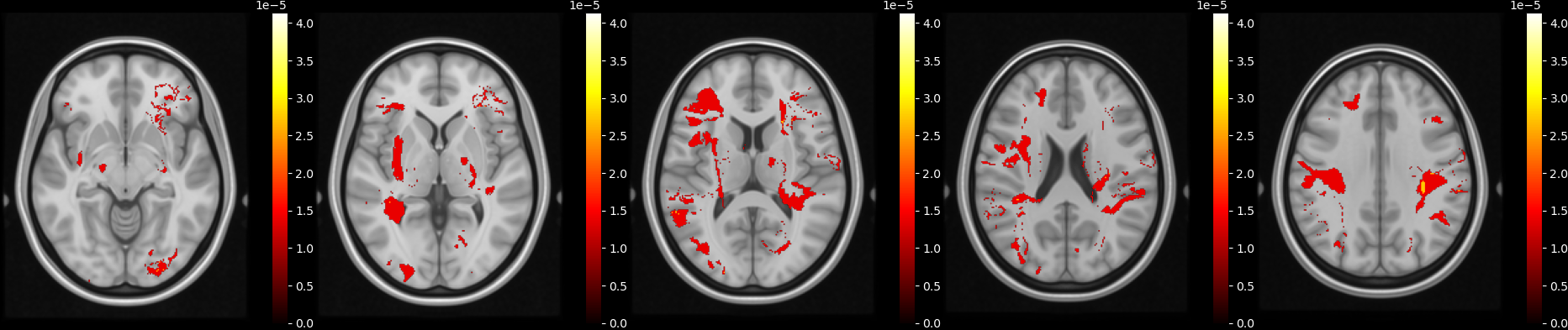}
         \caption{False-positives}
         \label{fig:mss1_prob_maps_test_isl_fp}
    \end{subfigure}
    \hfill
    \begin{subfigure}[]{1.0\textwidth}
         \centering
         \includegraphics[width=\textwidth]{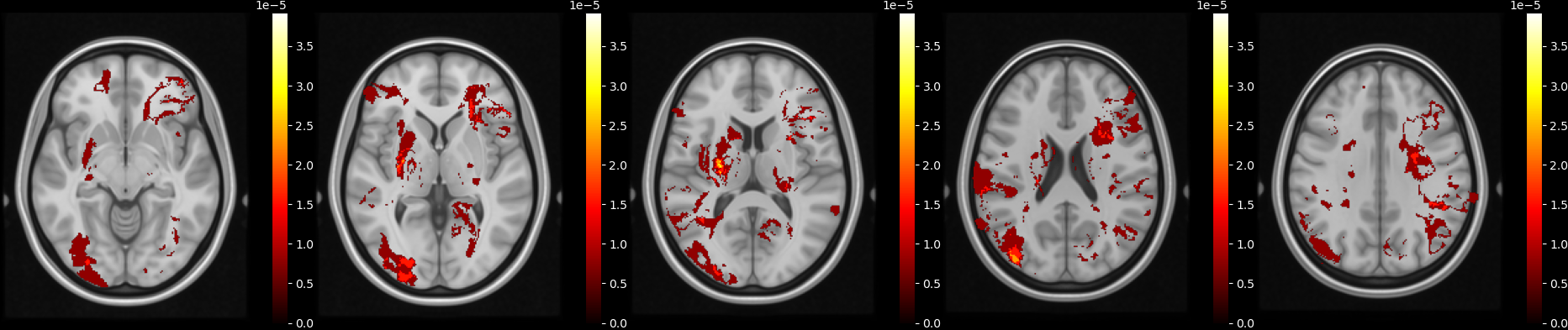}
         \caption{False-negatives}
         \label{fig:mss1_prob_maps_test_isl_fn}
    \end{subfigure}
    \caption{MSS1 Ischaemic Stroke Lesion (ISL) distributions (slices: 140, 150, 160, 170, 180)}
    \label{fig:mss1_prob_maps_full_gt}
\end{figure}

% MSS2
\begin{figure}[htbp]
     \centering
     \begin{subfigure}[]{1.0\textwidth}
         \centering
         \includegraphics[width=\textwidth]{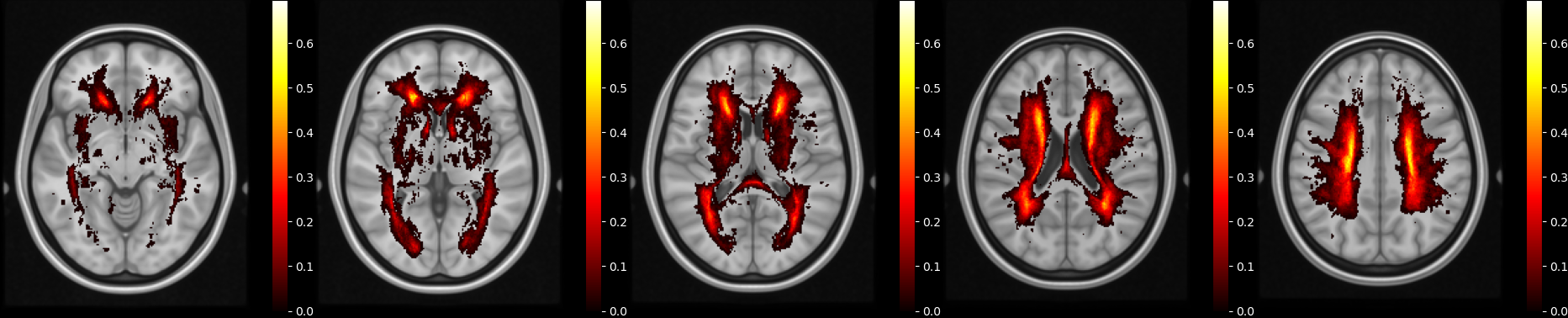}
         \caption{Ground truth}
         \label{fig:mss2_prob_maps_test_wmh_gt}
     \end{subfigure}
     \hfill
     \begin{subfigure}[]{1.0\textwidth}
         \centering
         \includegraphics[width=\textwidth]{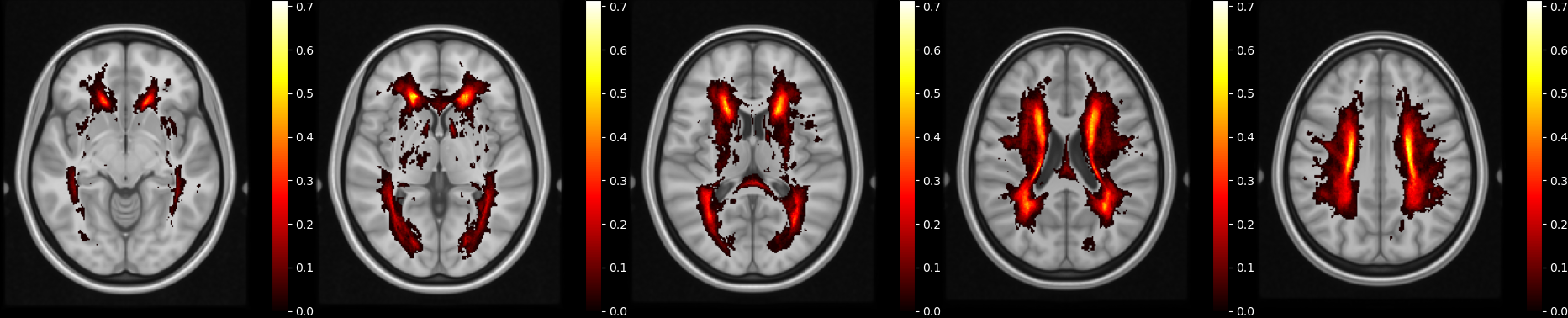}
         \caption{Predicted}
         \label{fig:mss2_prob_maps_test_wmh_pred}
    \end{subfigure}
    \hfill
    \begin{subfigure}[]{1.0\textwidth}
         \centering
         \includegraphics[width=\textwidth]{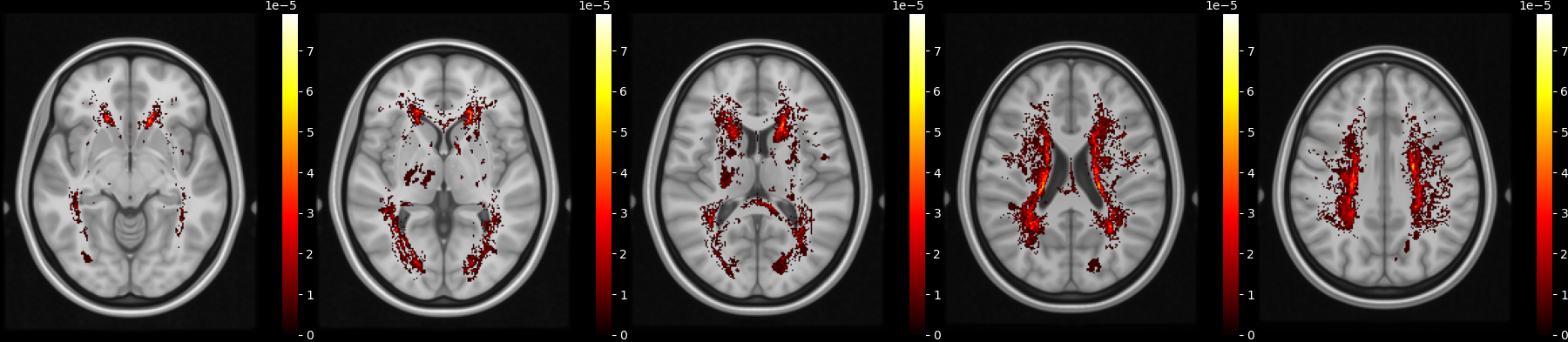}
         \caption{False-positives}
         \label{fig:mss2_prob_maps_test_wmh_fp}
    \end{subfigure}
    \hfill
    \begin{subfigure}[]{1.0\textwidth}
         \centering
         \includegraphics[width=\textwidth]{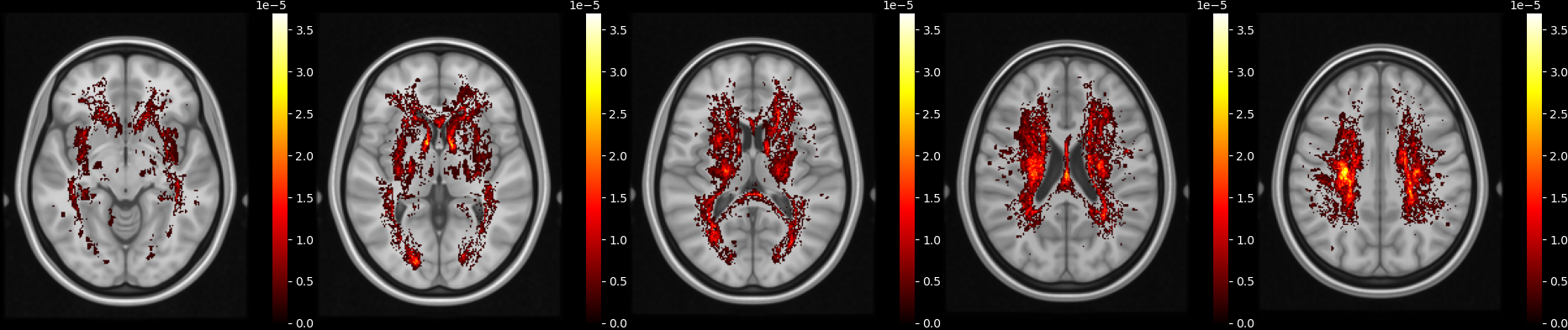}
         \caption{False-negatives}
         \label{fig:mss2_prob_maps_test_wmh_fn}
    \end{subfigure}
    \caption{MSS2 White Matter Hyperintensities (WMH) distributions (slices: 140, 150, 160, 170, 180)}
    \label{fig:mss2_prob_maps_full_gt}
\end{figure}

% MSS2
\begin{figure}[htbp]
     \centering
     \begin{subfigure}[]{1.0\textwidth}
         \centering
         \includegraphics[width=\textwidth]{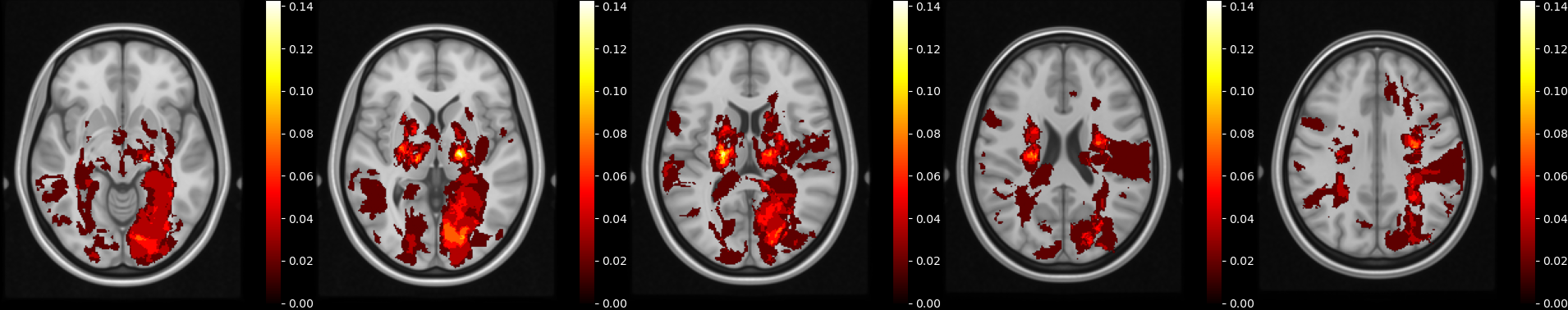}
         \caption{Ground truth}
         \label{fig:mss2_prob_maps_test_isl_gt}
     \end{subfigure}
     \hfill
     \begin{subfigure}[]{1.0\textwidth}
         \centering
         \includegraphics[width=\textwidth]{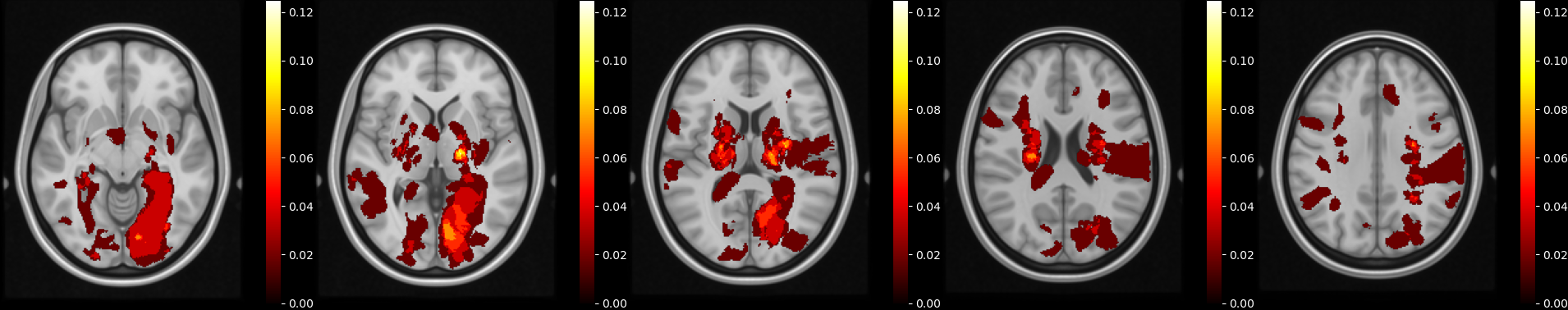}
         \caption{Predicted}
         \label{fig:mss2_prob_maps_test_isl_pred}
    \end{subfigure}
    \hfill
    \begin{subfigure}[]{1.0\textwidth}
         \centering
         \includegraphics[width=\textwidth]{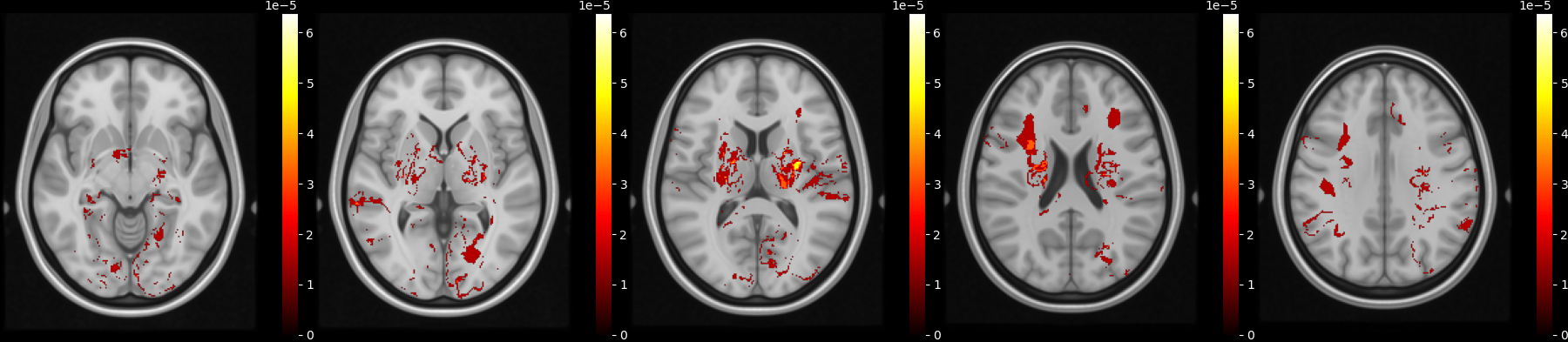}
         \caption{False-positives}
         \label{fig:mss2_prob_maps_test_isl_fp}
    \end{subfigure}
    \hfill
    \begin{subfigure}[]{1.0\textwidth}
         \centering
         \includegraphics[width=\textwidth]{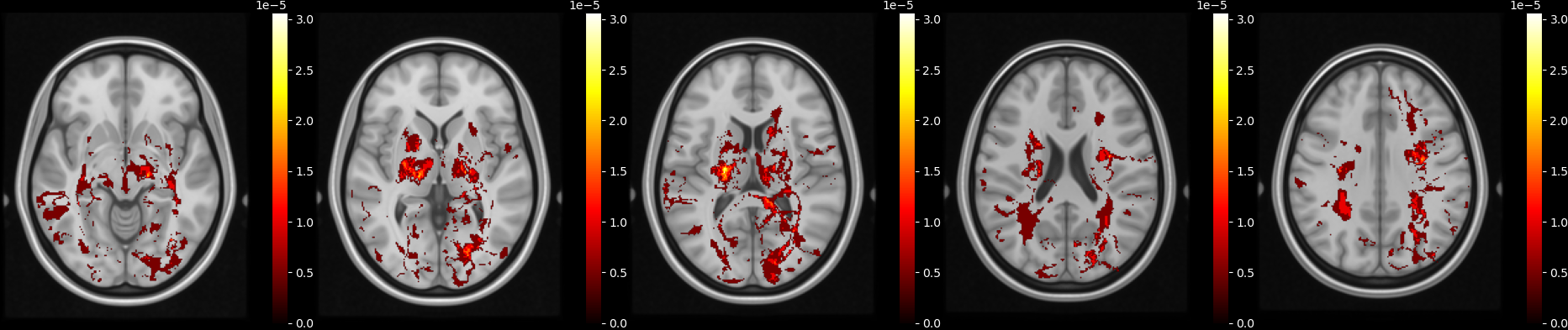}
         \caption{False-negatives}
         \label{fig:mss2_prob_maps_test_isl_fn}
    \end{subfigure}
    \caption{MSS2 Ischaemic Stroke Lesion (ISL) distributions (slices: 140, 150, 160, 170, 180)}
    \label{fig:mss2_prob_maps_full_gt}
\end{figure}

% MSS3
\begin{figure}[htbp]
     \centering
     \begin{subfigure}[]{1.0\textwidth}
         \centering
         \includegraphics[width=\textwidth]{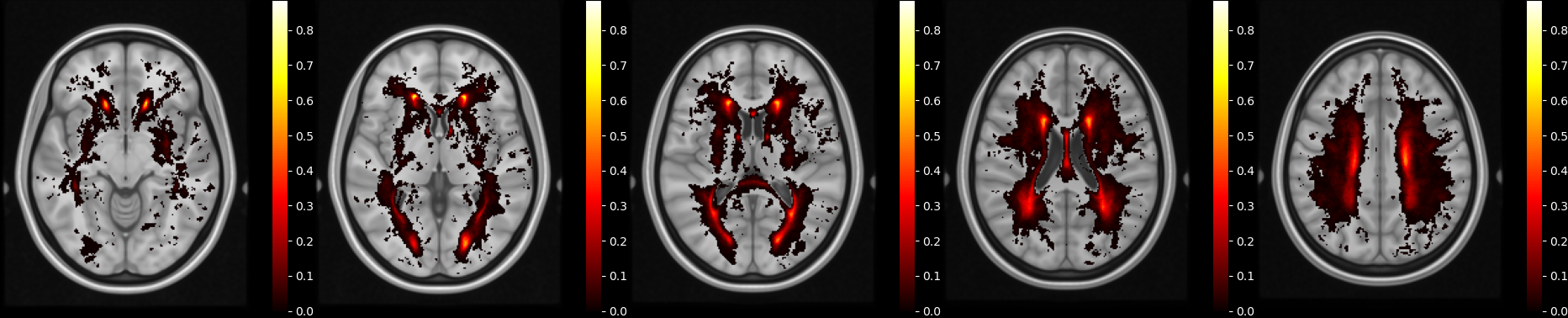}
         \caption{Ground truth}
         \label{fig:mss3_prob_maps_test_wmh_gt}
     \end{subfigure}
     \hfill
     \begin{subfigure}[]{1.0\textwidth}
         \centering
         \includegraphics[width=\textwidth]{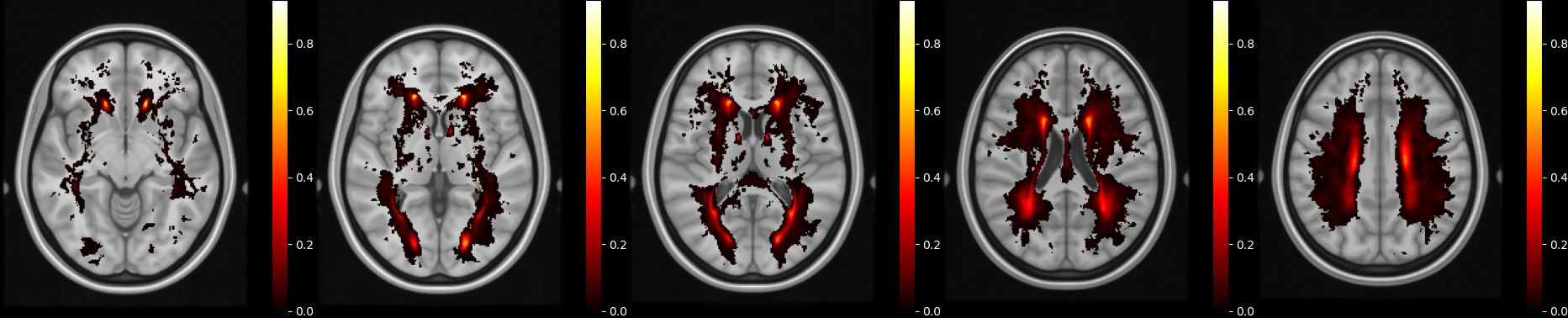}
         \caption{Predicted}
         \label{fig:mss3_prob_maps_test_wmh_pred}
    \end{subfigure}
    \hfill
    \begin{subfigure}[]{1.0\textwidth}
         \centering
         \includegraphics[width=\textwidth]{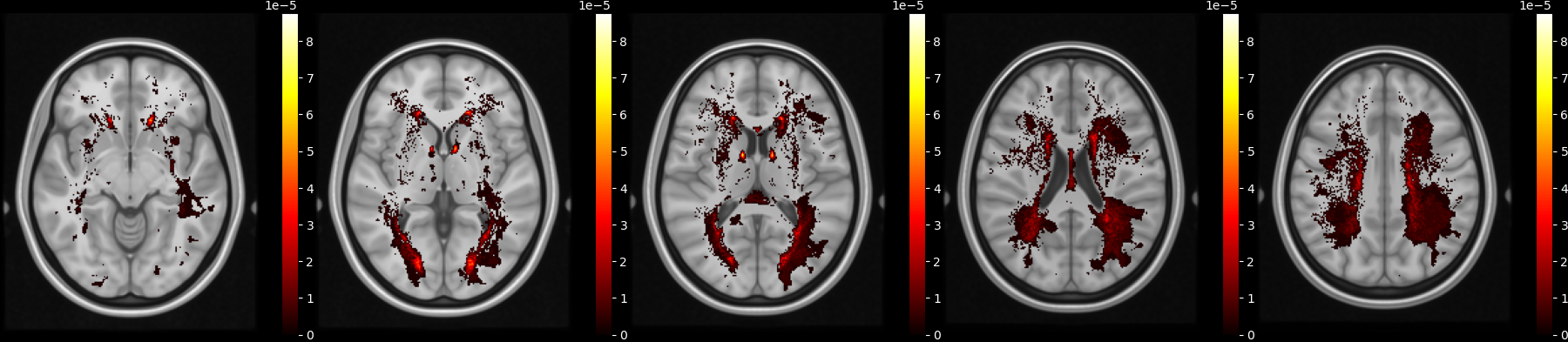}
         \caption{False-positives}
         \label{fig:mss3_prob_maps_test_wmh_fp}
    \end{subfigure}
    \hfill
    \begin{subfigure}[]{1.0\textwidth}
         \centering
         \includegraphics[width=\textwidth]{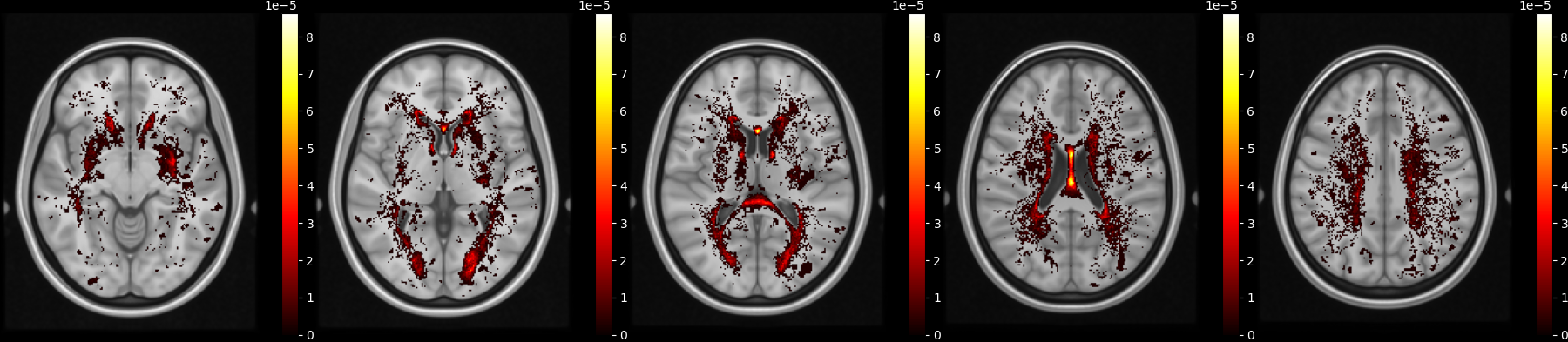}
         \caption{False-negatives}
         \label{fig:mss3_prob_maps_test_wmh_fn}
    \end{subfigure}
    \caption{MSS3 White Matter Hyperintensities (WMH) distributions (slices: 140, 150, 160, 170, 180)}
    \label{fig:mss3_prob_maps_full_gt}
\end{figure}

% MSS3
\begin{figure}[htbp]
     \centering
     \begin{subfigure}[]{1.0\textwidth}
         \centering
         \includegraphics[width=\textwidth]{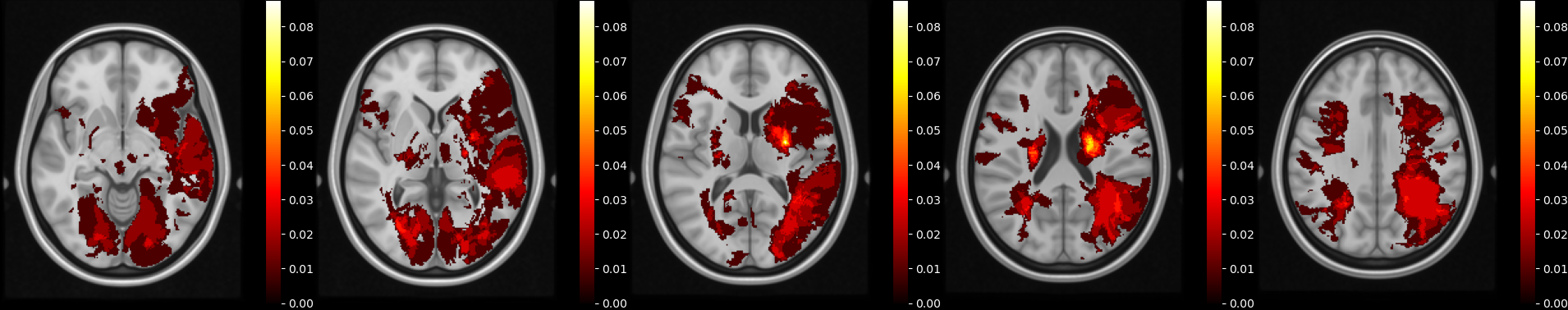}
         \caption{Ground truth}
         \label{fig:mss3_prob_maps_test_isl_gt}
     \end{subfigure}
     \hfill
     \begin{subfigure}[]{1.0\textwidth}
         \centering
         \includegraphics[width=\textwidth]{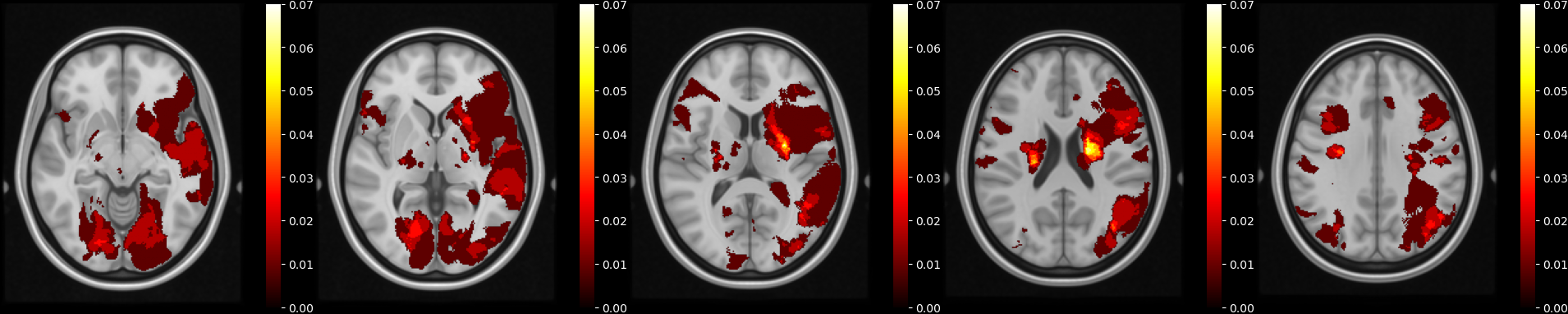}
         \caption{Predicted}
         \label{fig:mss3_prob_maps_test_isl_pred}
    \end{subfigure}
    \hfill
    \begin{subfigure}[]{1.0\textwidth}
         \centering
         \includegraphics[width=\textwidth]{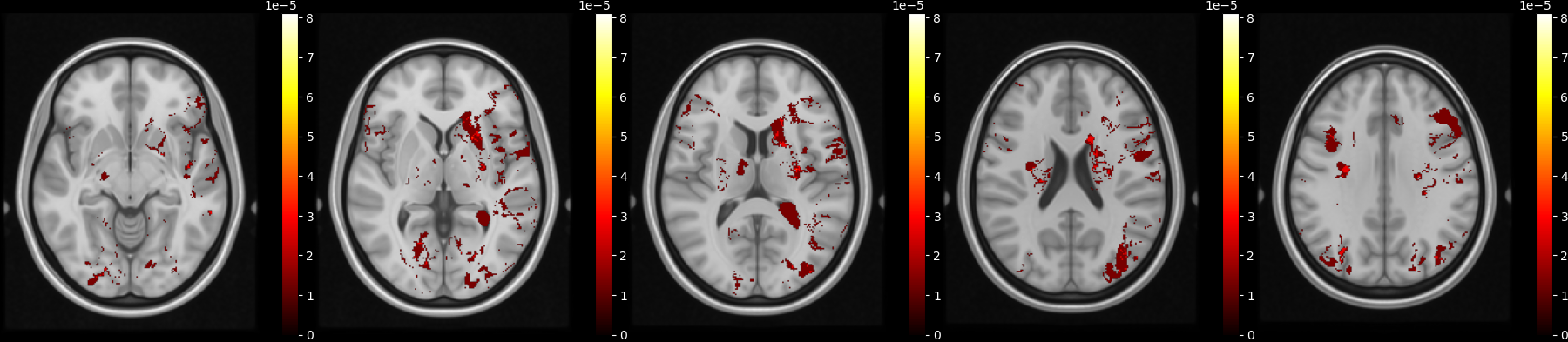}
         \caption{False-positives}
         \label{fig:mss3_prob_maps_test_isl_fp}
    \end{subfigure}
    \hfill
    \begin{subfigure}[]{1.0\textwidth}
         \centering
         \includegraphics[width=\textwidth]{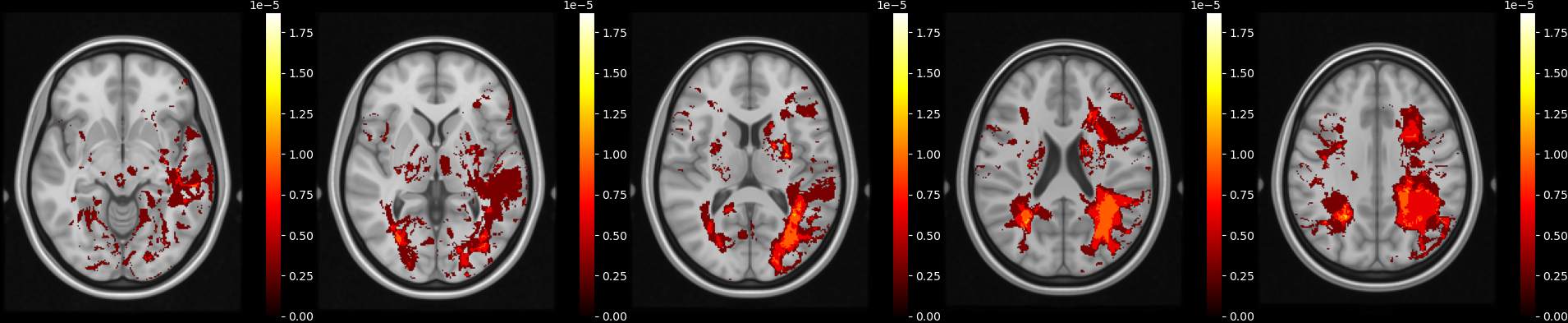}
         \caption{False-negatives}
         \label{fig:mss3_prob_maps_test_isl_fn}
    \end{subfigure}
    \caption{MSS3 Ischaemic Stroke Lesion (ISL) distributions (slices: 140, 150, 160, 170, 180)}
    \label{fig:mss3_prob_maps_full_gt}
\end{figure}

% LBC1936
\begin{figure}[htbp]
     \centering
     \begin{subfigure}[]{1.0\textwidth}
         \centering
         \includegraphics[width=\textwidth]{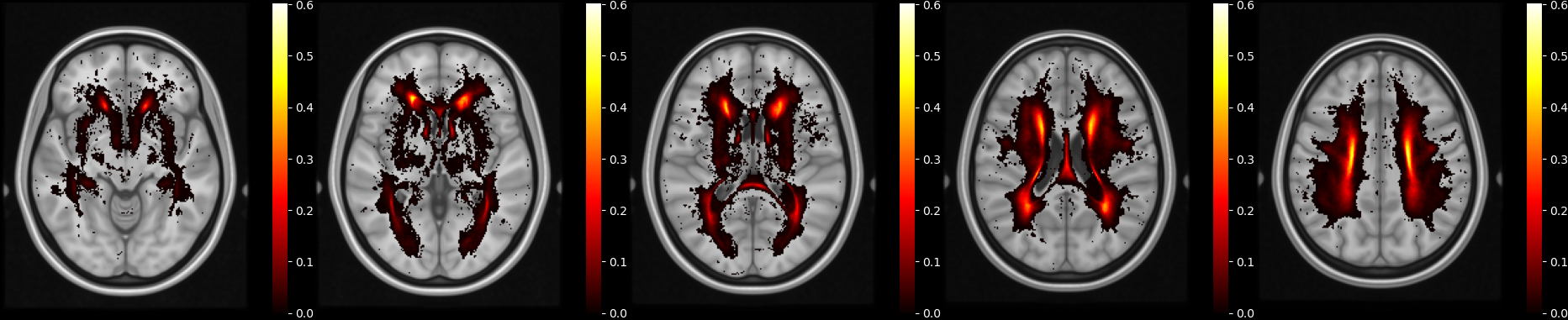}
         \caption{Ground truth}
         \label{fig:lbc1936_prob_maps_test_wmh_gt}
     \end{subfigure}
     \hfill
     \begin{subfigure}[]{1.0\textwidth}
         \centering
         \includegraphics[width=\textwidth]{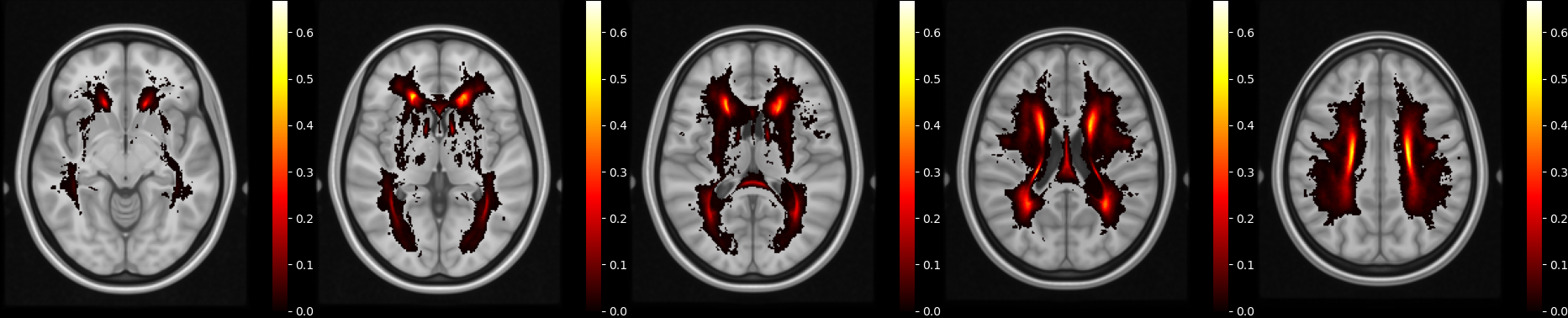}
         \caption{Predicted}
         \label{fig:lbc1936_prob_maps_test_wmh_pred}
    \end{subfigure}
    \hfill
    \begin{subfigure}[]{1.0\textwidth}
         \centering
         \includegraphics[width=\textwidth]{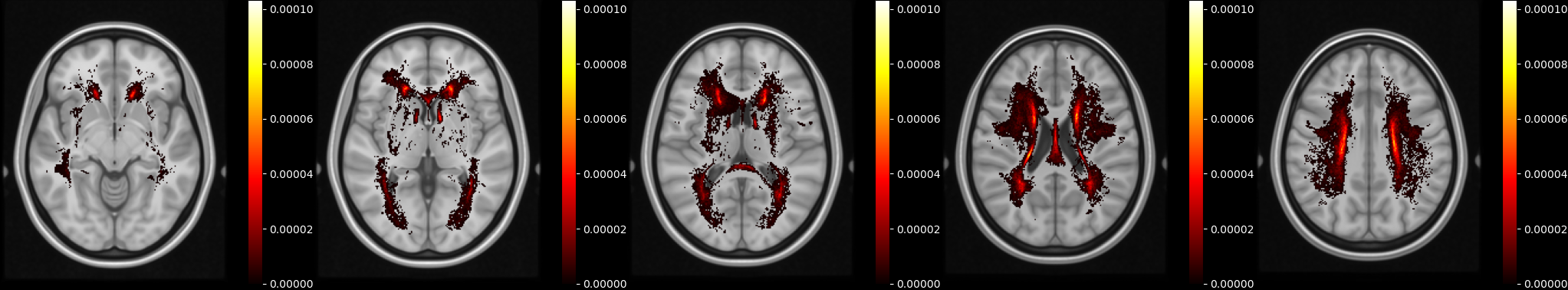}
         \caption{False-positives}
         \label{fig:lbc1936_prob_maps_test_wmh_fp}
    \end{subfigure}
    \hfill
    \begin{subfigure}[]{1.0\textwidth}
         \centering
         \includegraphics[width=\textwidth]{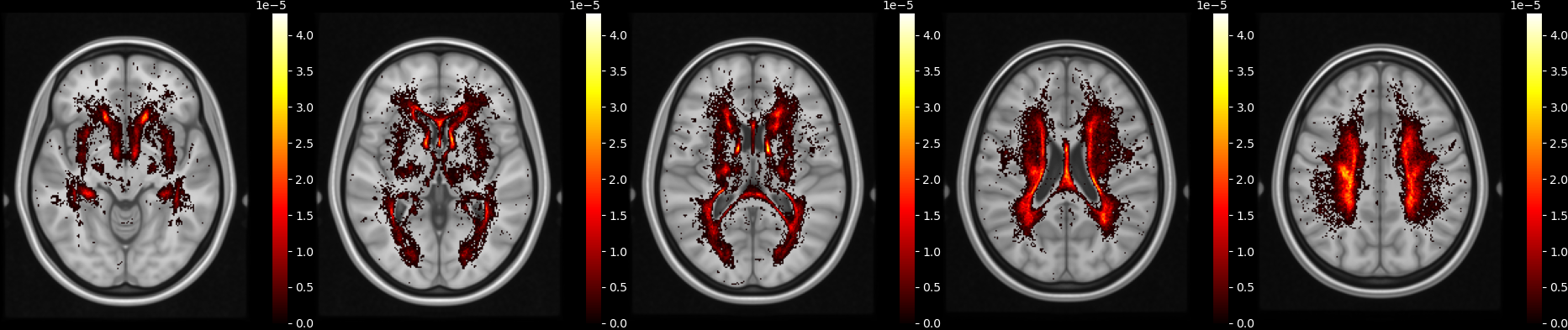}
         \caption{False-negatives}
         \label{fig:lbc1936_prob_maps_test_wmh_fn}
    \end{subfigure}
    \caption{LBC1936 White Matter Hyperintensities (WMH) distributions (slices: 140, 150, 160, 170, 180)}
    \label{fig:lbc1936_prob_maps_full_gt}
\end{figure}

% LBC1936
\begin{figure}[htbp]
     \centering
     \begin{subfigure}[]{1.0\textwidth}
         \centering
         \includegraphics[width=\textwidth]{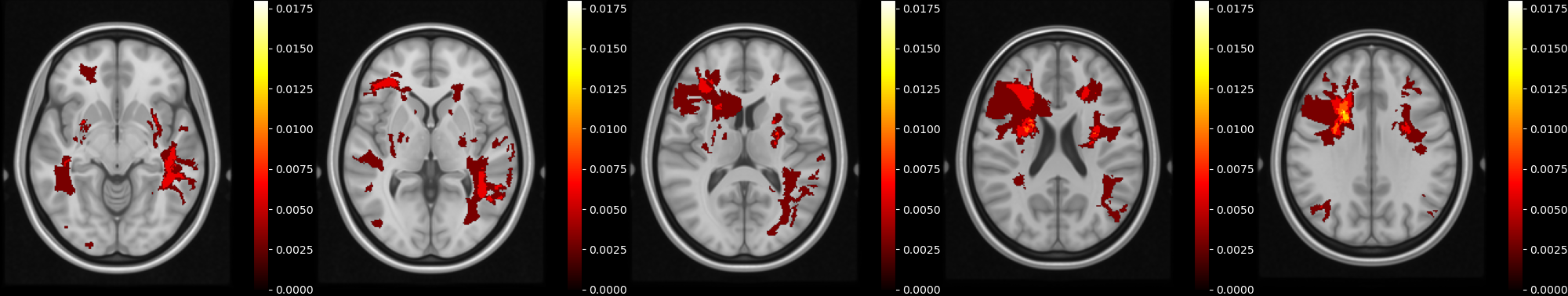}
         \caption{Ground truth}
         \label{fig:lbc1936_prob_maps_test_isl_gt}
     \end{subfigure}
     \hfill
     \begin{subfigure}[]{1.0\textwidth}
         \centering
         \includegraphics[width=\textwidth]{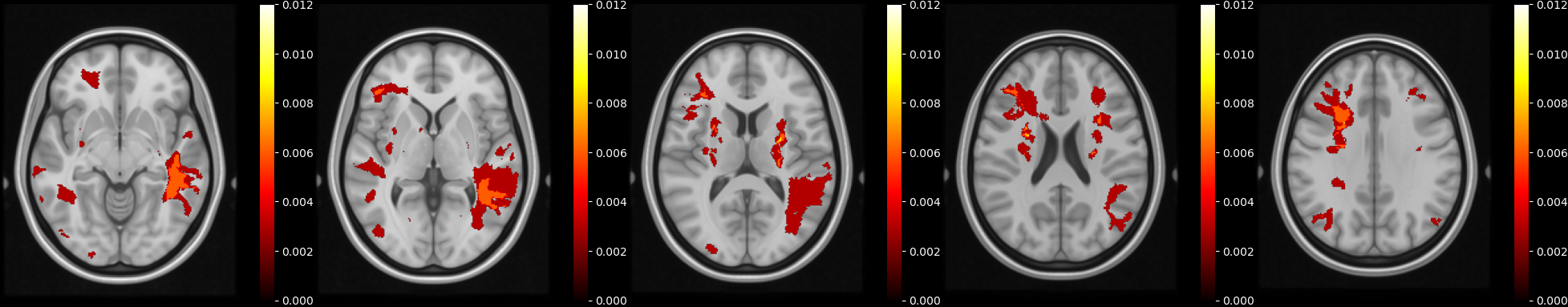}
         \caption{Predicted}
         \label{fig:lbc1936_prob_maps_test_isl_pred}
    \end{subfigure}
    \hfill
    \begin{subfigure}[]{1.0\textwidth}
         \centering
         \includegraphics[width=\textwidth]{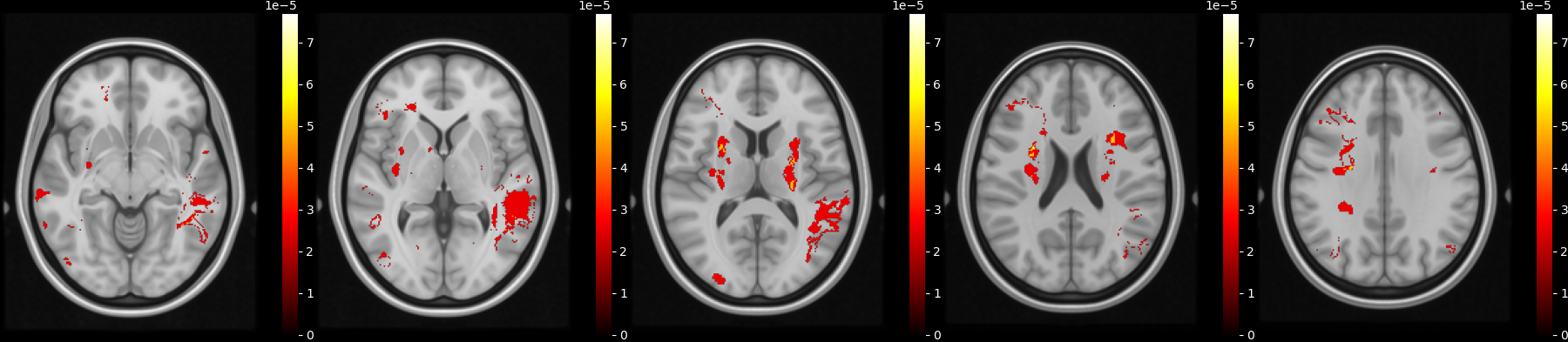}
         \caption{False-positives}
         \label{fig:lbc1936_prob_maps_test_isl_fp}
    \end{subfigure}
    \hfill
    \begin{subfigure}[]{1.0\textwidth}
         \centering
         \includegraphics[width=\textwidth]{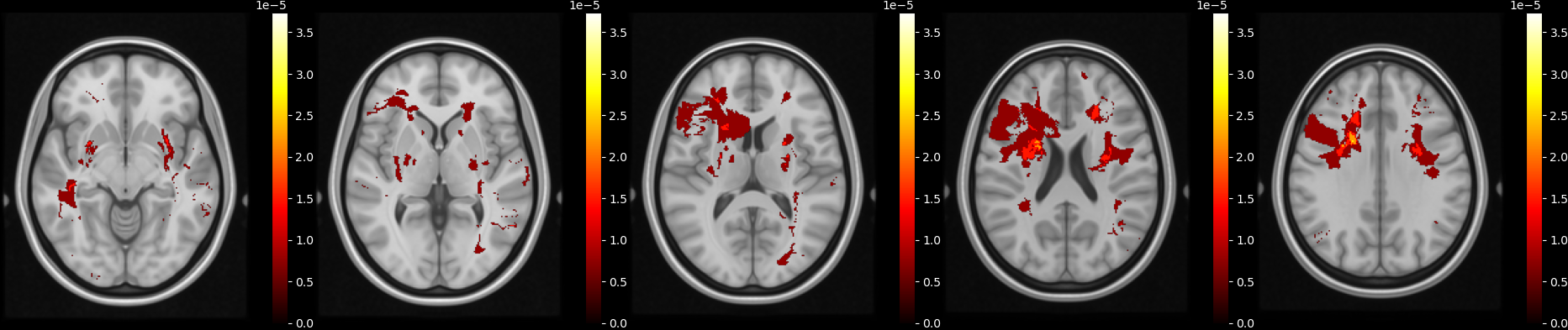}
         \caption{False-negatives}
         \label{fig:lbc1936_prob_maps_test_isl_fn}
    \end{subfigure}
    \caption{LBC1936 Ischaemic Stroke Lesion (ISL) distributions (slices: 140, 150, 160, 170, 180)}
    \label{fig:lbc1936_prob_maps_full_gt}
\end{figure}

% LBC1921
\begin{figure}[htbp]
     \centering
     \begin{subfigure}[]{1.0\textwidth}
         \centering
         \includegraphics[width=\textwidth]{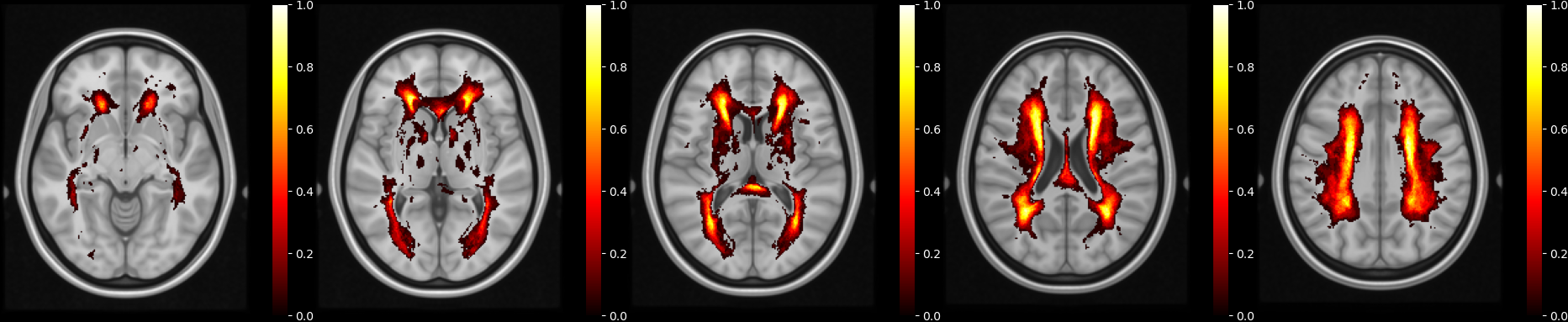}
         \caption{Ground truth}
         \label{fig:lbc1921_prob_maps_test_wmh_gt}
     \end{subfigure}
     \hfill
     \begin{subfigure}[]{1.0\textwidth}
         \centering
         \includegraphics[width=\textwidth]{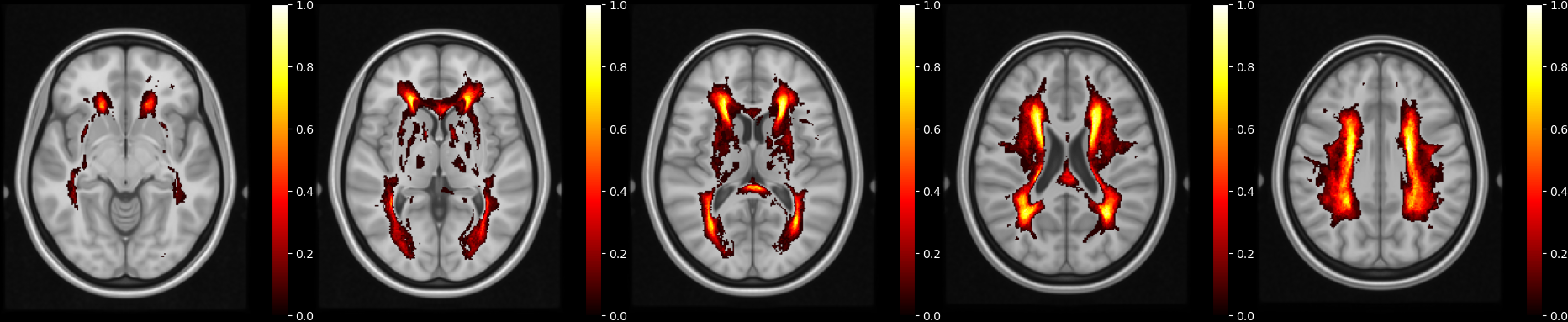}
         \caption{Predicted}
         \label{fig:lbc1921_prob_maps_test_wmh_pred}
    \end{subfigure}
    \hfill
    \begin{subfigure}[]{1.0\textwidth}
         \centering
         \includegraphics[width=\textwidth]{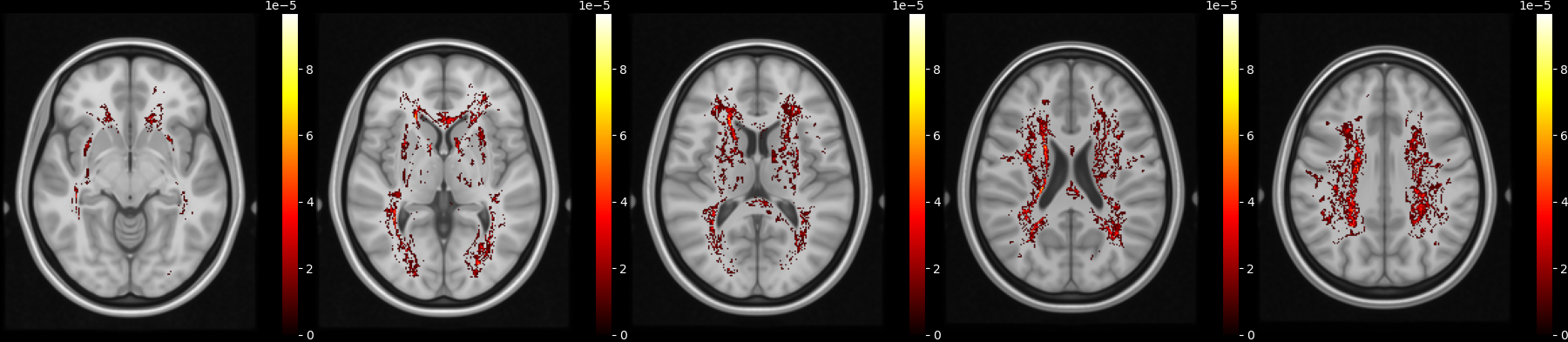}
         \caption{False-positives}
         \label{fig:lbc1921_prob_maps_test_wmh_fp}
    \end{subfigure}
    \hfill
    \begin{subfigure}[]{1.0\textwidth}
         \centering
         \includegraphics[width=\textwidth]{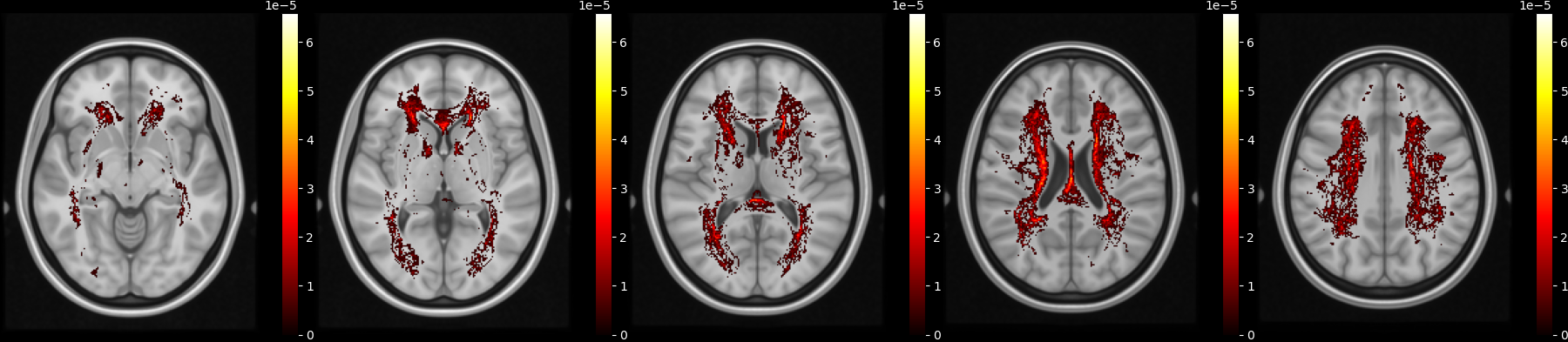}
         \caption{False-negatives}
         \label{fig:lbc1921_prob_maps_test_wmh_fn}
    \end{subfigure}
    \caption{LBC1921 White Matter Hyperintensities (WMH) distributions (slices: 140, 150, 160, 170, 180)}
    \label{fig:lbc1921_prob_maps_full_gt}
\end{figure}

% WMH-ch
\begin{figure}[htbp]
     \centering
     \begin{subfigure}[]{1.0\textwidth}
         \centering
         \includegraphics[width=\textwidth]{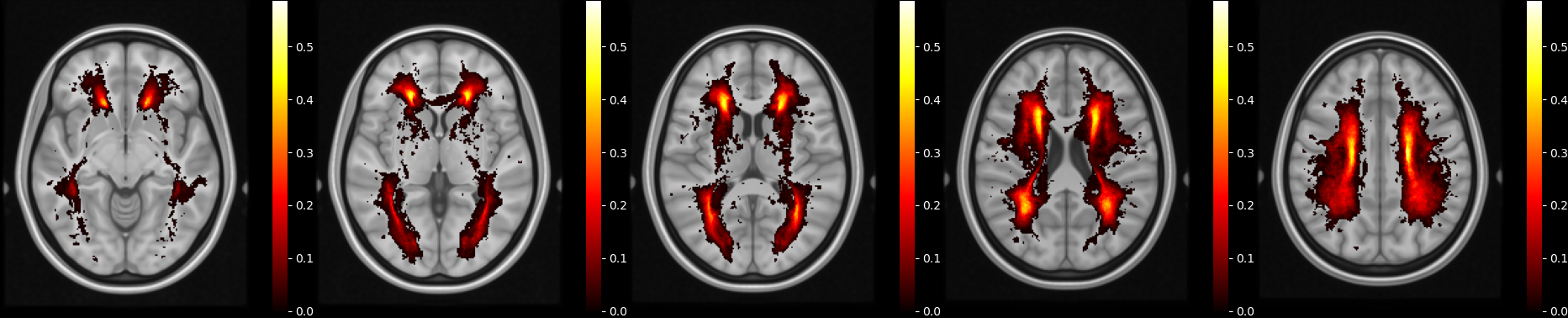}
         \caption{Ground truth}
         \label{fig:wmh_prob_maps_test_wmh_gt}
     \end{subfigure}
     \hfill
     \begin{subfigure}[]{1.0\textwidth}
         \centering
         \includegraphics[width=\textwidth]{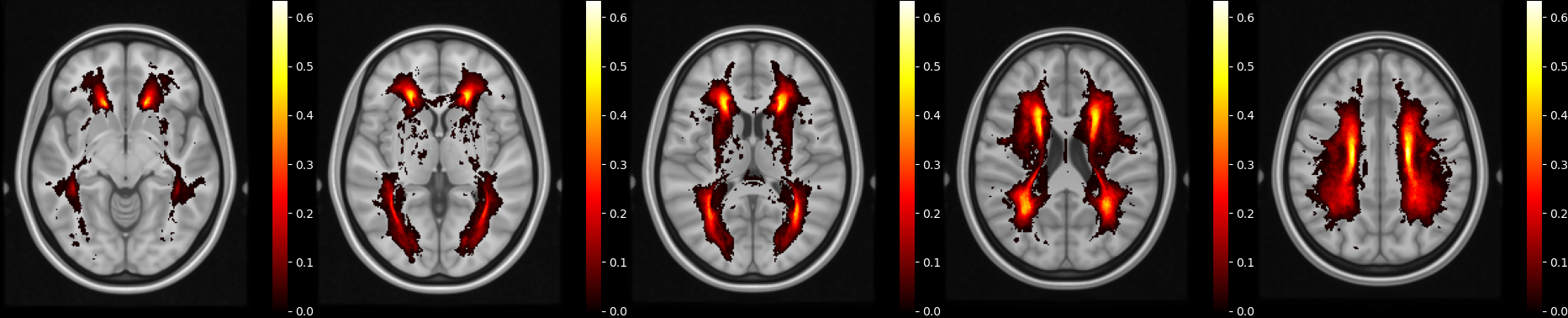}
         \caption{Predicted}
         \label{fig:wmh_prob_maps_test_wmh_pred}
    \end{subfigure}
    \hfill
    \begin{subfigure}[]{1.0\textwidth}
         \centering
         \includegraphics[width=\textwidth]{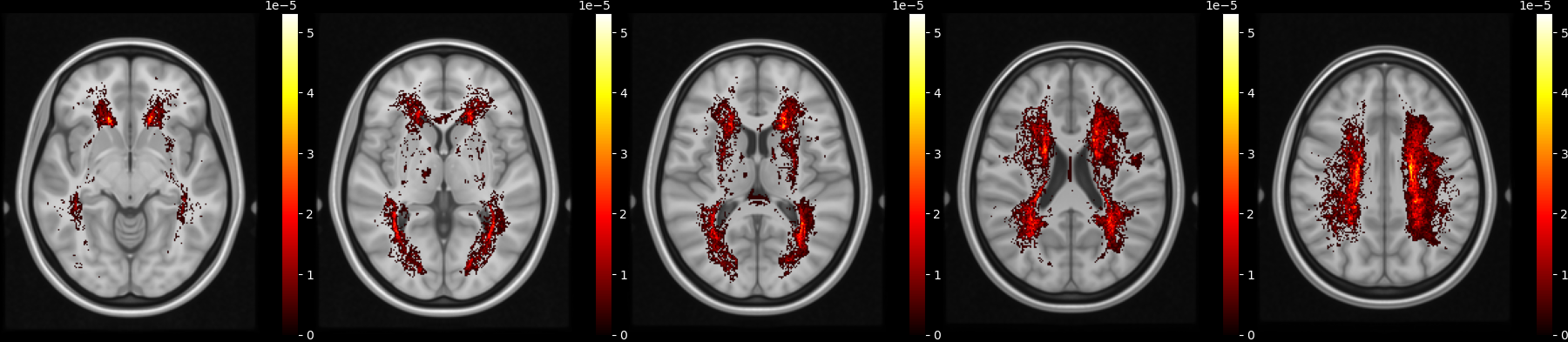}
         \caption{False-positives}
         \label{fig:wmh_prob_maps_test_wmh_fp}
    \end{subfigure}
    \hfill
    \begin{subfigure}[]{1.0\textwidth}
         \centering
         \includegraphics[width=\textwidth]{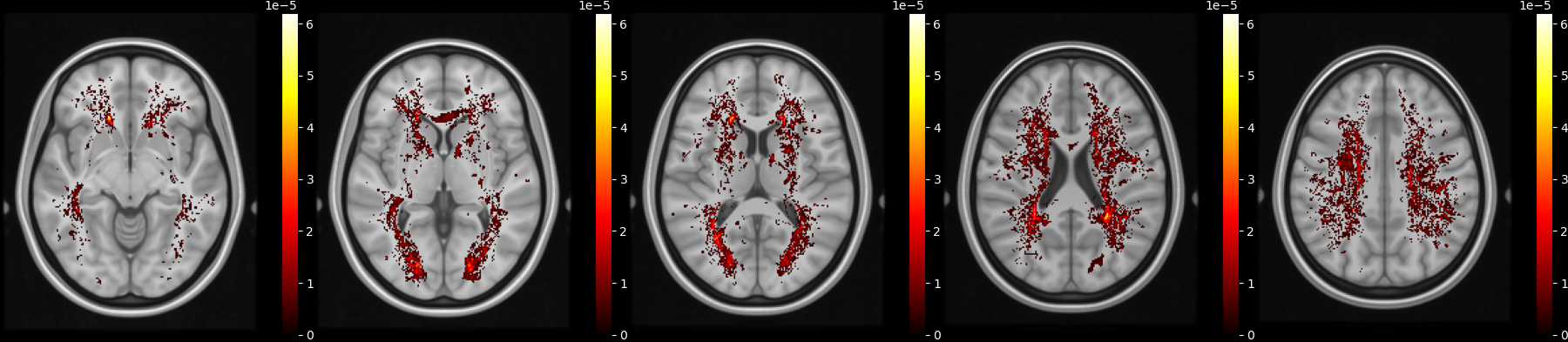}
         \caption{False-negatives}
         \label{fig:wmh_prob_maps_test_wmh_fn}
    \end{subfigure}
    \caption{WMH-ch White Matter Hyperintensities (WMH) distributions (slices: 140, 150, 160, 170, 180)}
    \label{fig:wmh_prob_maps_full_gt}
\end{figure}

% ISLES
\begin{figure}[htbp]
     \centering
     \begin{subfigure}[]{1.0\textwidth}
         \centering
         \includegraphics[width=\textwidth]{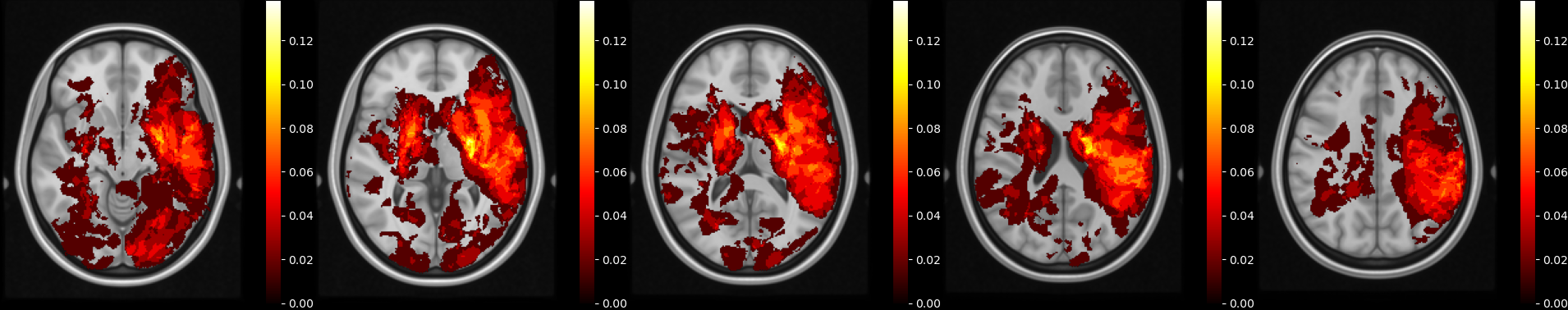}
         \caption{Ground truth}
         \label{fig:isles_prob_maps_test_isl_gt}
     \end{subfigure}
     \hfill
     \begin{subfigure}[]{1.0\textwidth}
         \centering
         \includegraphics[width=\textwidth]{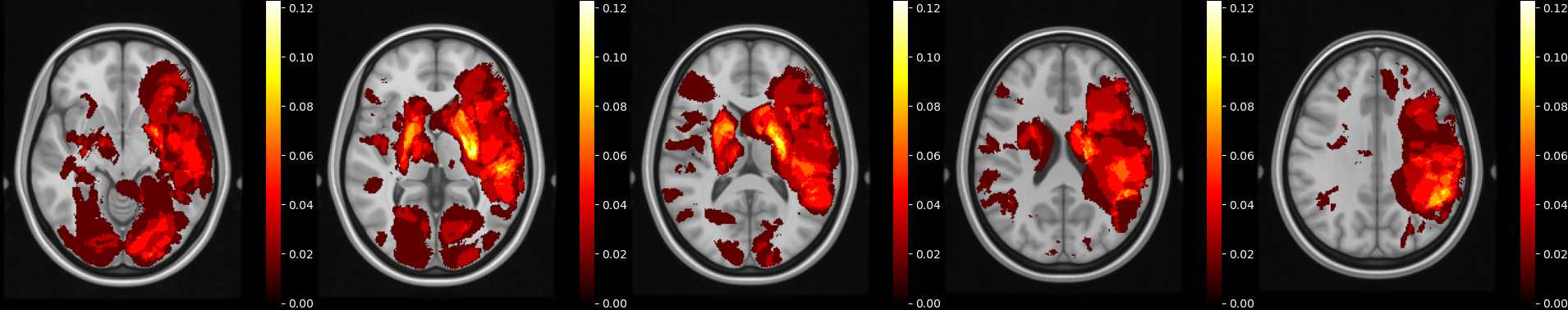}
         \caption{Predicted}
         \label{fig:isles_prob_maps_test_isl_pred}
    \end{subfigure}
    \hfill
    \begin{subfigure}[]{1.0\textwidth}
         \centering
         \includegraphics[width=\textwidth]{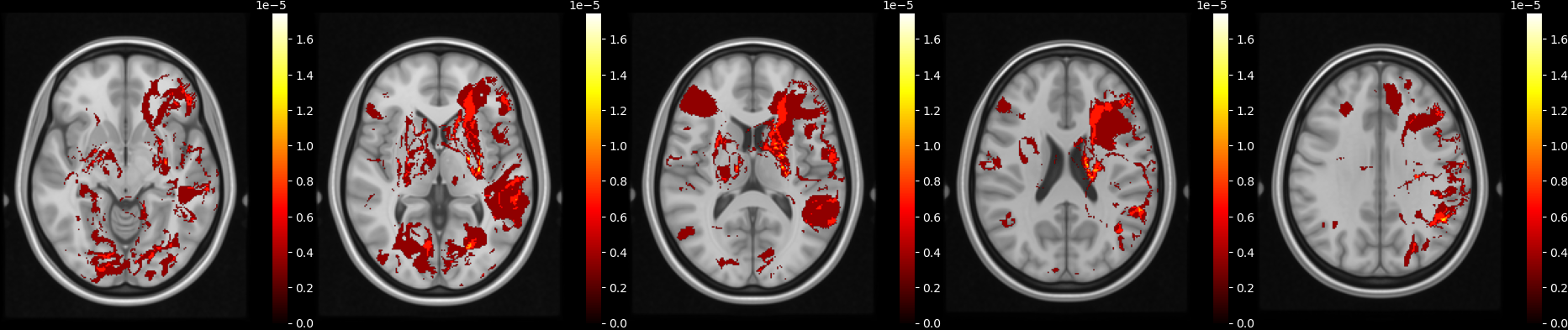}
         \caption{False-positives}
         \label{fig:isles_prob_maps_test_isl_fp}
    \end{subfigure}
    \hfill
    \begin{subfigure}[]{1.0\textwidth}
         \centering
         \includegraphics[width=\textwidth]{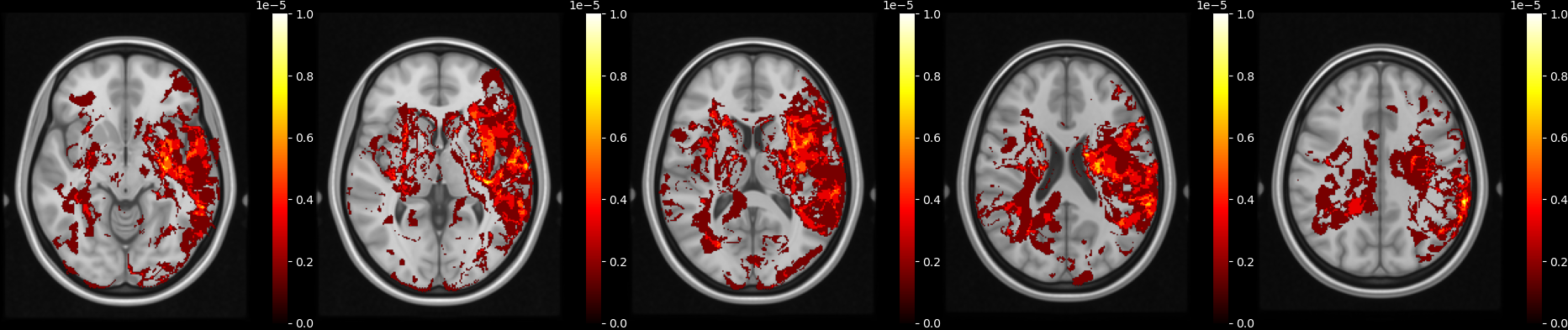}
         \caption{False-negatives}
         \label{fig:isles_prob_maps_test_isl_fn}
    \end{subfigure}
    \caption{ISLES Ischaemic Stroke Lesion (ISL) distributions (slices: 140, 150, 160, 170, 180)}
    \label{fig:isles_prob_maps_full_gt}
\end{figure}

% SOOP
\begin{figure}[htbp]
     \centering
     \begin{subfigure}[]{1.0\textwidth}
         \centering
         \includegraphics[width=\textwidth]{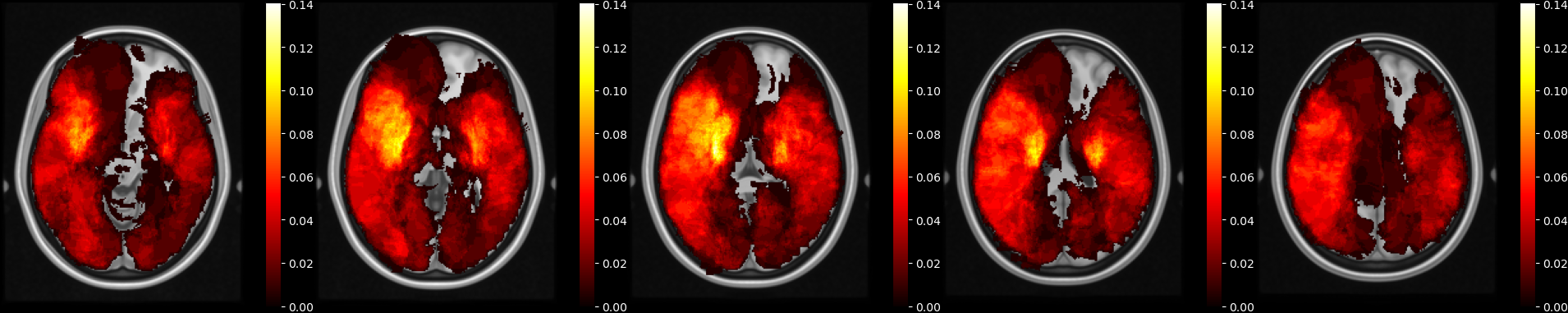}
         \caption{Ground truth}
         \label{fig:soop_prob_maps_test_isl_gt}
     \end{subfigure}
     \hfill
     \begin{subfigure}[]{1.0\textwidth}
         \centering
         \includegraphics[width=\textwidth]{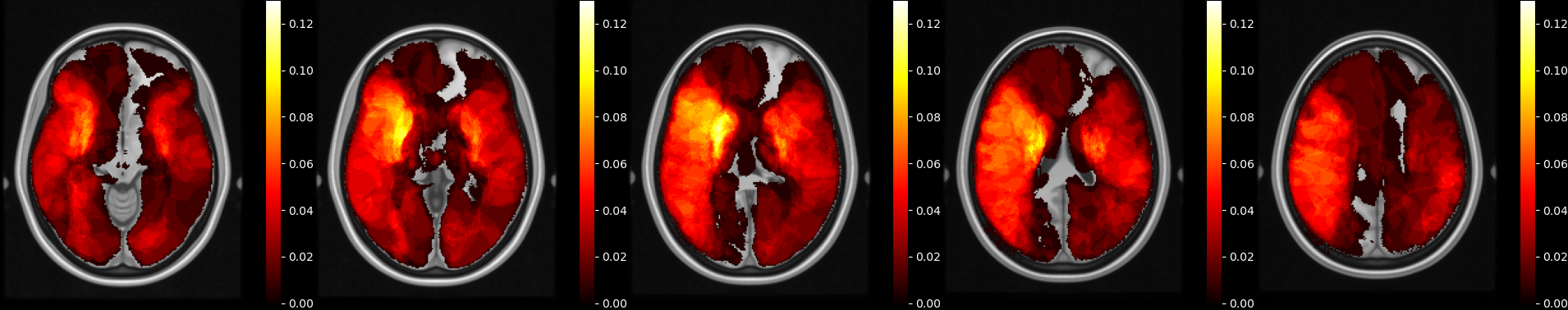}
         \caption{Predicted}
         \label{fig:soop_prob_maps_test_isl_pred}
    \end{subfigure}
    \hfill
    \begin{subfigure}[]{1.0\textwidth}
         \centering
         \includegraphics[width=\textwidth]{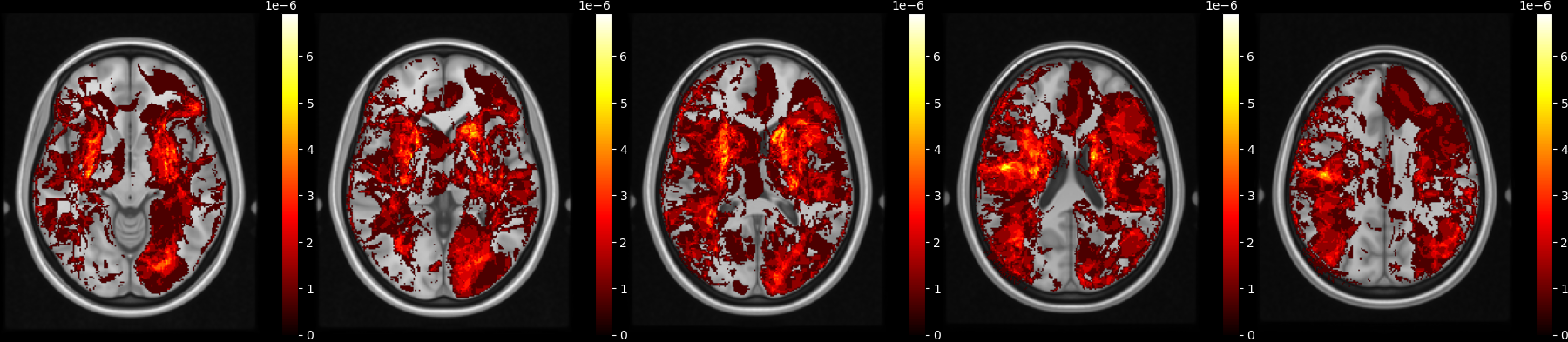}
         \caption{False-positives}
         \label{fig:soop_prob_maps_test_isl_fp}
    \end{subfigure}
    \hfill
    \begin{subfigure}[]{1.0\textwidth}
         \centering
         \includegraphics[width=\textwidth]{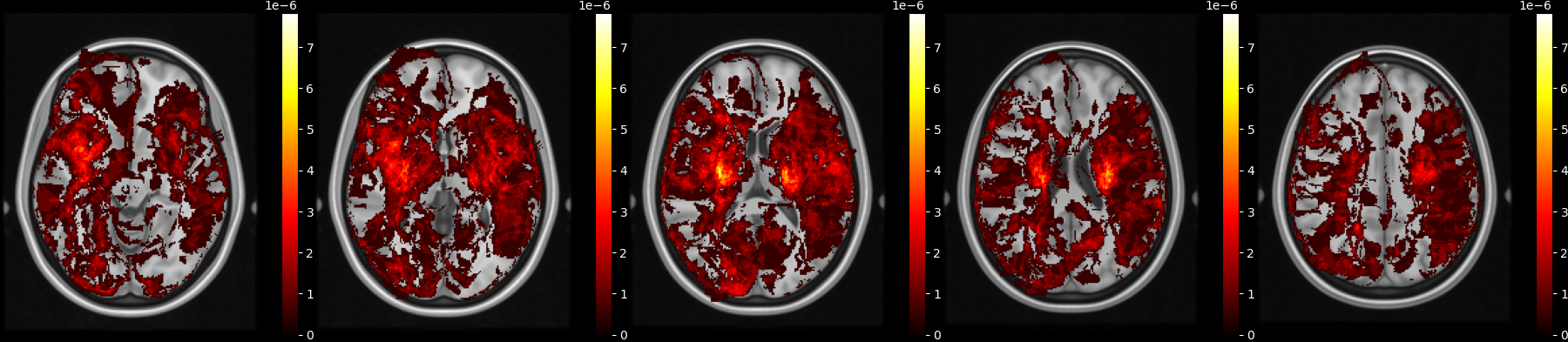}
         \caption{False-negatives}
         \label{fig:soop_prob_maps_test_isl_fn}
    \end{subfigure}
    \caption{SOOP Ischaemic Stroke Lesion (ISL) distributions (slices: 140, 150, 160, 170, 180)}
    \label{fig:soop_prob_maps_full_gt}
\end{figure}

% WSS
\begin{figure}[htbp]
     \centering
     \begin{subfigure}[]{1.0\textwidth}
         \centering
         \includegraphics[width=\textwidth]{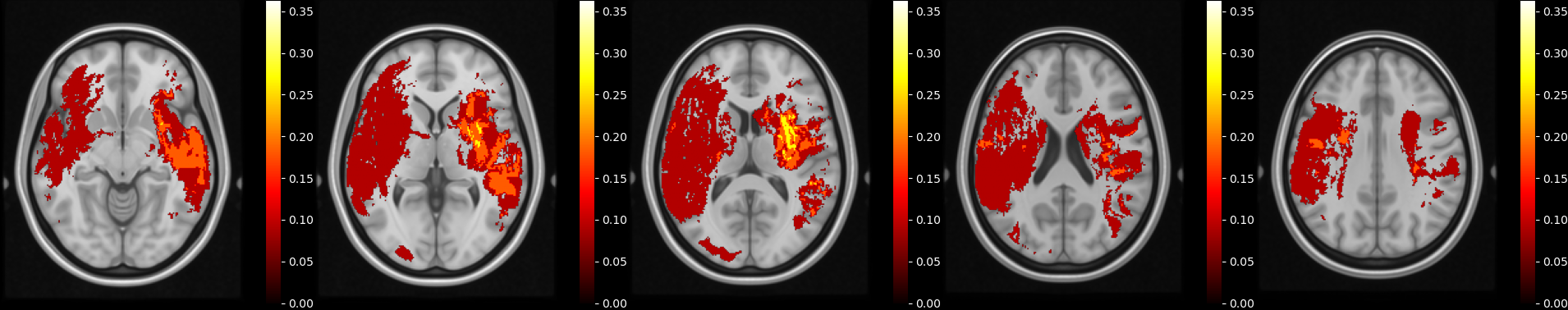}
         \caption{Ground truth}
         \label{fig:wss_prob_maps_test_isl_gt}
     \end{subfigure}
     \hfill
     \begin{subfigure}[]{1.0\textwidth}
         \centering
         \includegraphics[width=\textwidth]{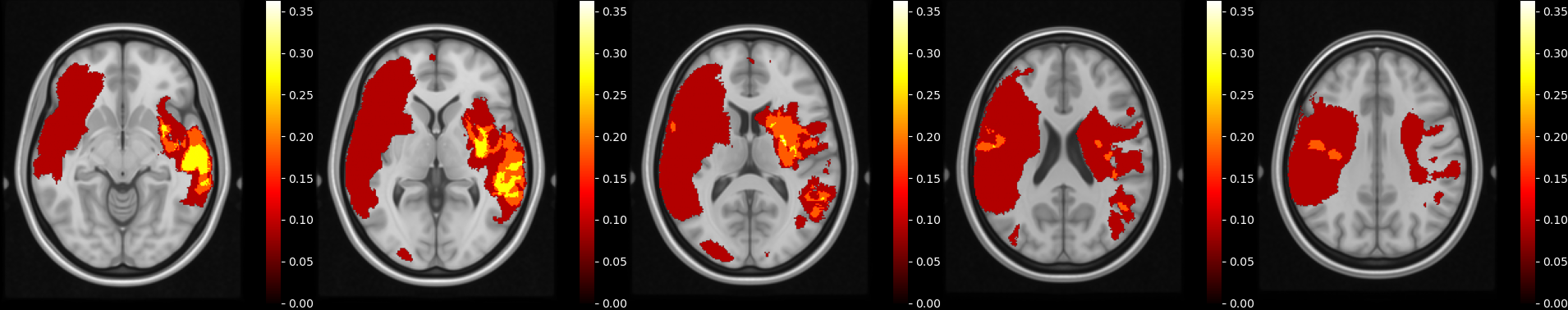}
         \caption{Predicted}
         \label{fig:wss_prob_maps_test_isl_pred}
    \end{subfigure}
    \hfill
    \begin{subfigure}[]{1.0\textwidth}
         \centering
         \includegraphics[width=\textwidth]{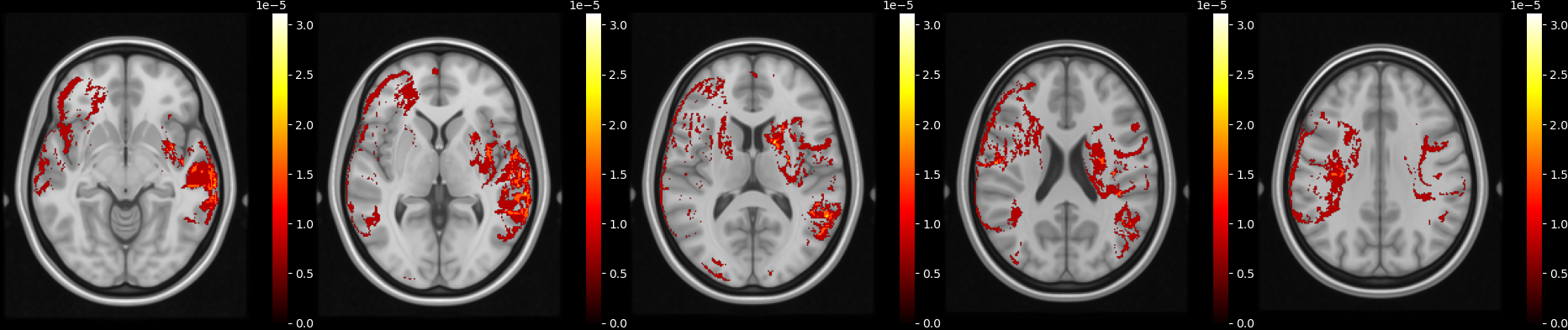}
         \caption{False-positives}
         \label{fig:wss_prob_maps_test_isl_fp}
    \end{subfigure}
    \hfill
    \begin{subfigure}[]{1.0\textwidth}
         \centering
         \includegraphics[width=\textwidth]{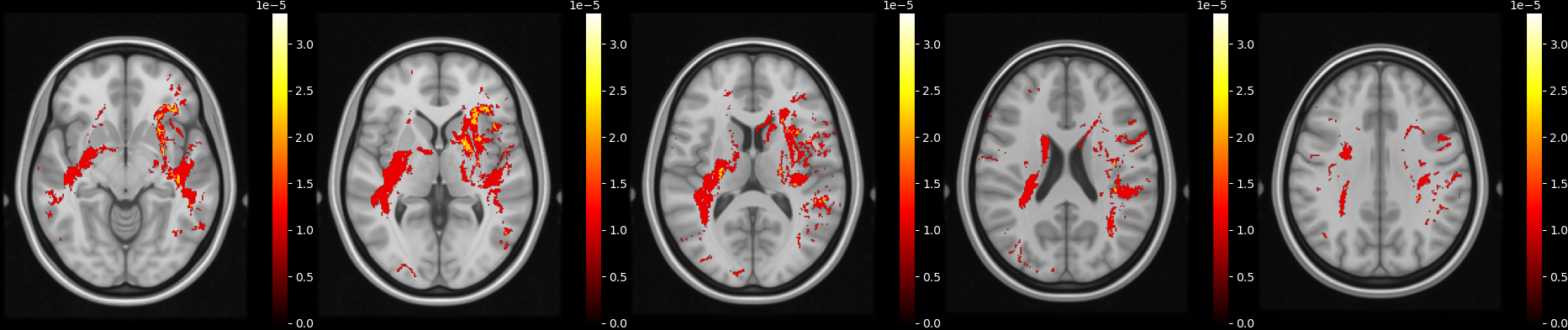}
         \caption{False-negatives}
         \label{fig:wss_prob_maps_test_isl_fn}
    \end{subfigure}
    \caption{WSS Ischaemic Stroke Lesion (ISL) distributions (slices: 140, 150, 160, 170, 180)}
    \label{fig:wss_prob_maps_full_gt}
\end{figure}

% ESS
\begin{figure}[htbp]
     \centering
     \begin{subfigure}[]{1.0\textwidth}
         \centering
         \includegraphics[width=\textwidth]{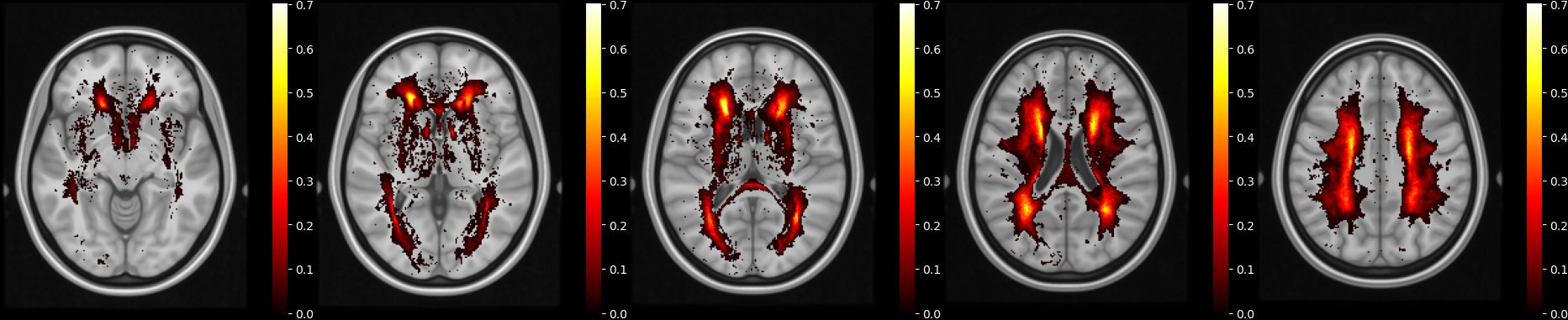}
         \caption{Ground truth}
         \label{fig:ess_prob_maps_test_wmh_gt}
     \end{subfigure}
     \hfill
     \begin{subfigure}[]{1.0\textwidth}
         \centering
         \includegraphics[width=\textwidth]{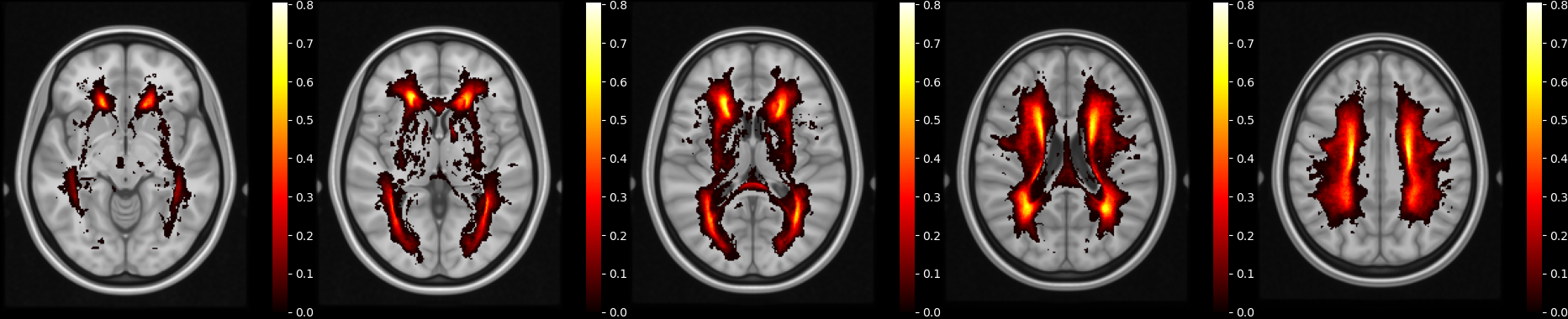}
         \caption{Predicted}
         \label{fig:ess_prob_maps_test_wmh_pred}
    \end{subfigure}
    \hfill
    \begin{subfigure}[]{1.0\textwidth}
         \centering
         \includegraphics[width=\textwidth]{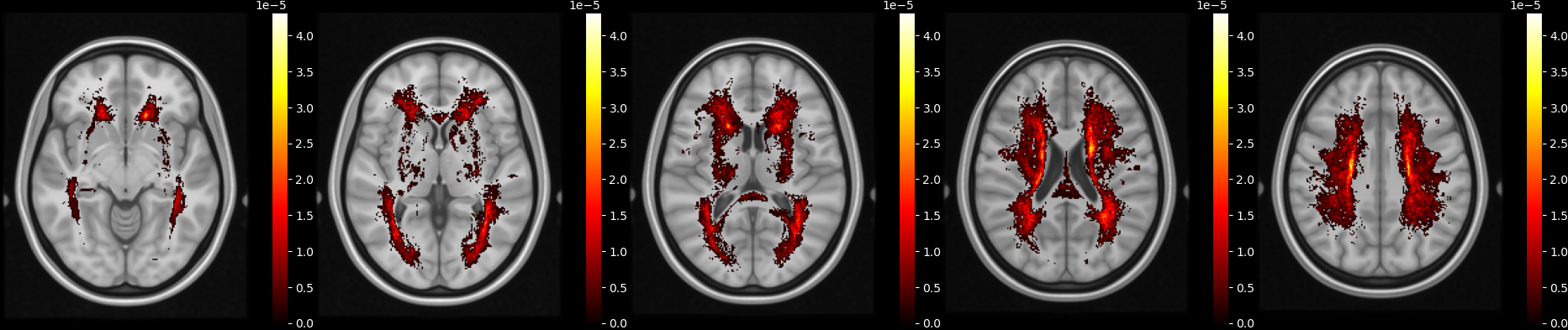}
         \caption{False-positives}
         \label{fig:ess_prob_maps_test_wmh_fp}
    \end{subfigure}
    \hfill
    \begin{subfigure}[]{1.0\textwidth}
         \centering
         \includegraphics[width=\textwidth]{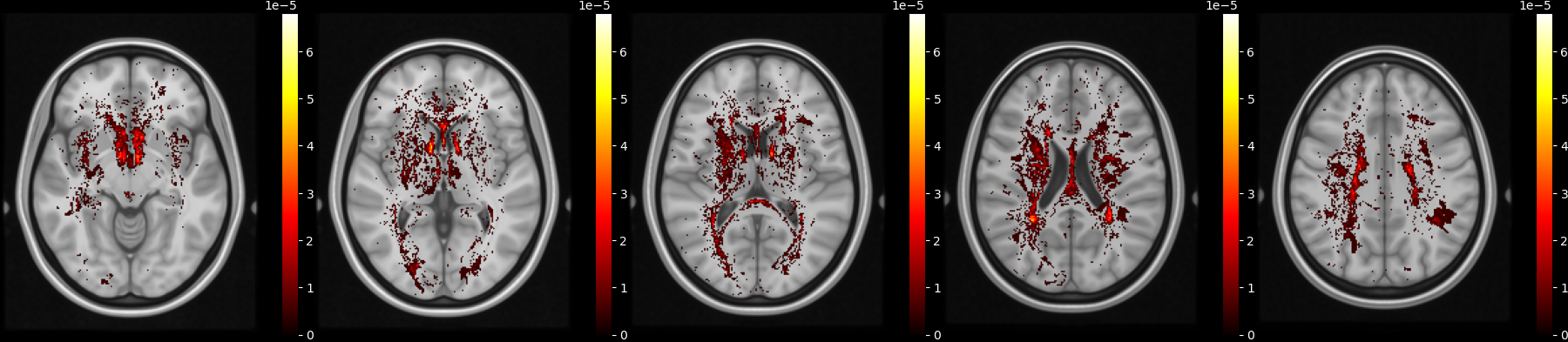}
         \caption{False-negatives}
         \label{fig:ess_prob_maps_test_wmh_fn}
    \end{subfigure}
    \caption{ESS White Matter Hyperintensities (WMH) distributions (slices: 140, 150, 160, 170, 180)}
    \label{fig:ess_prob_maps_full_gt}
\end{figure}

% ESS
\begin{figure}[htbp]
     \centering
     \begin{subfigure}[]{1.0\textwidth}
         \centering
         \includegraphics[width=\textwidth]{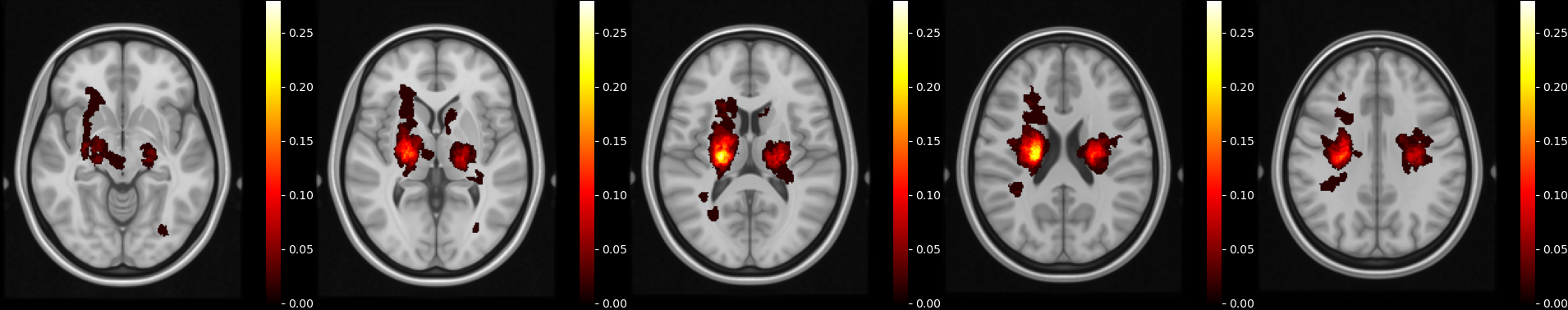}
         \caption{Ground truth}
         \label{fig:ess_prob_maps_test_isl_gt}
     \end{subfigure}
     \hfill
     \begin{subfigure}[]{1.0\textwidth}
         \centering
         \includegraphics[width=\textwidth]{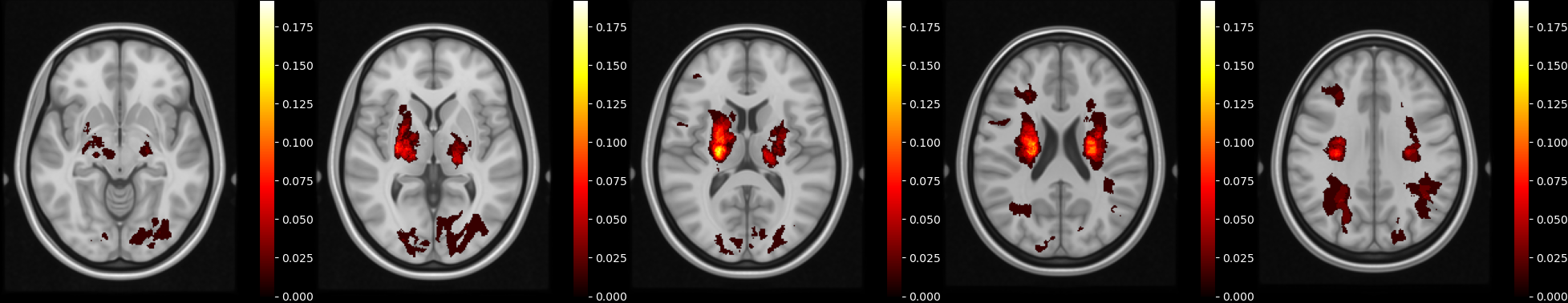}
         \caption{Predicted}
         \label{fig:ess_prob_maps_test_isl_pred}
    \end{subfigure}
    \hfill
    \begin{subfigure}[]{1.0\textwidth}
         \centering
         \includegraphics[width=\textwidth]{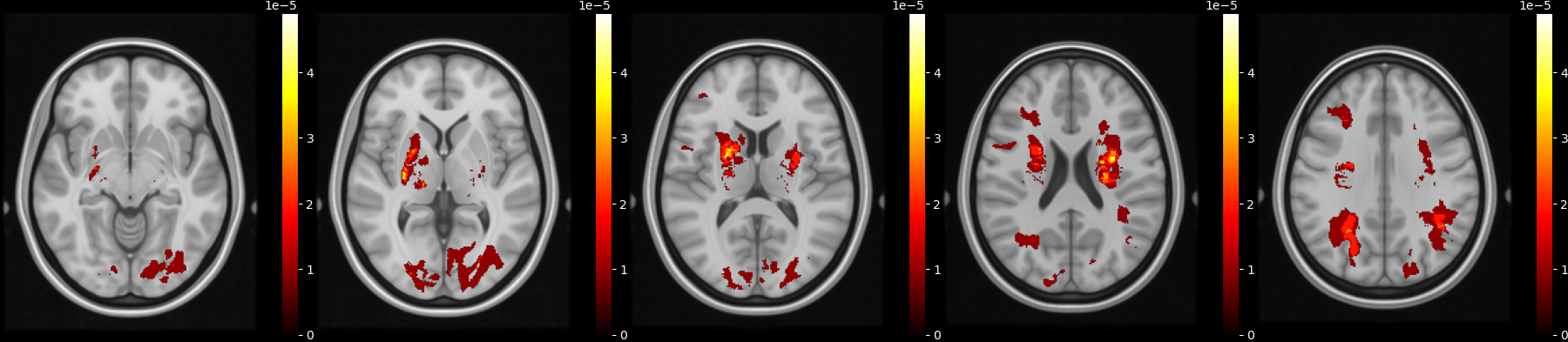}
         \caption{False-positives}
         \label{fig:ess_prob_maps_test_isl_fp}
    \end{subfigure}
    \hfill
    \begin{subfigure}[]{1.0\textwidth}
         \centering
         \includegraphics[width=\textwidth]{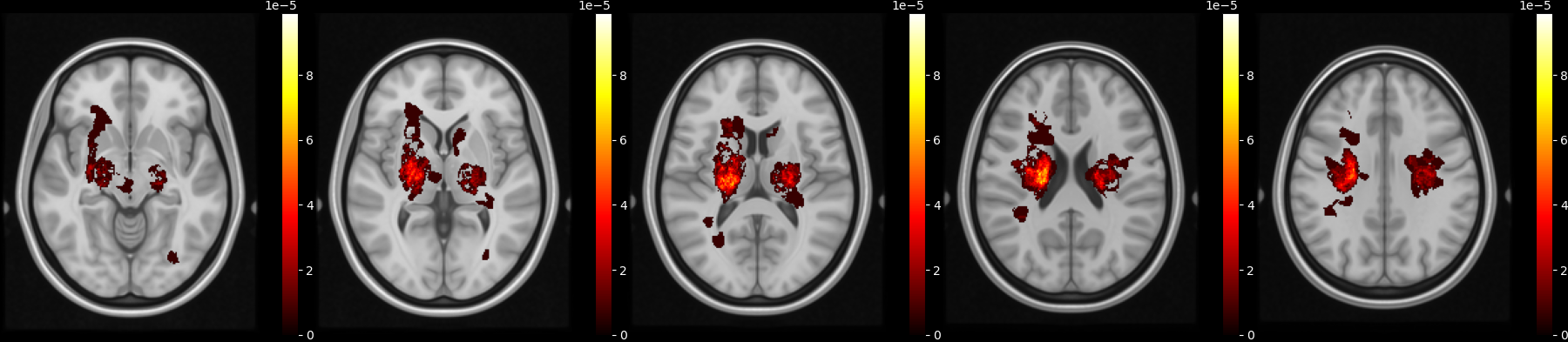}
         \caption{False-negatives}
         \label{fig:ess_prob_maps_test_isl_fn}
    \end{subfigure}
    \caption{ESS Ischaemic Stroke Lesion (ISL) distributions (slices: 140, 150, 160, 170, 180)}
    \label{fig:ess_prob_maps_full_gt}
\end{figure}

\clearpage

\section{Statistical tables} \label{appendix C}

The following figures and tables present the pairwise statistical significance ($p<0.05$) results from our bootstrap testing.

\subsection{Main model comparison}

These figures report the statistical comparison results of the metric outcomes for the models evaluated in \textbf{Tables \ref{tab:results_classes}} and \textbf{\ref{tab:results_avg}} of the main text. Critical difference plots are used to visualise multiple model comparisons effectively.

\begin{figure}[htbp]
    \centering
    \includegraphics[width=1\linewidth]{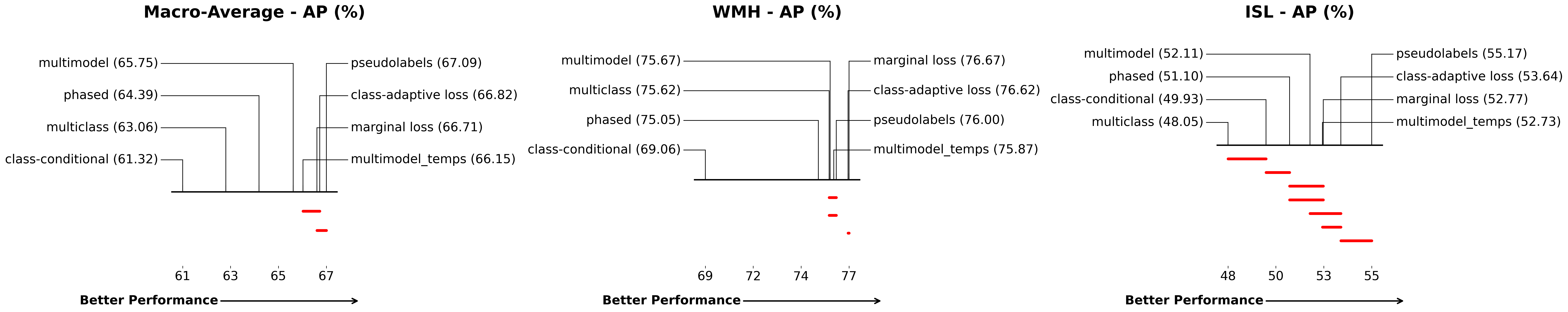}
    \caption{Critical difference (CD) diagrams showing the statistical significance of performance differences between models for Macro-Average, WMH, and ISL classes. The x-axis plots the mean performance on the average precision (AP) metric. Horizontal red lines group models that perform comparably (no statistically significant difference ($p\geq0.05$)), while unlinked models exhibit significant performance differences.}
    \label{fig:main_cd_ap}
\end{figure}

\begin{figure}[htbp]
    \centering
    \includegraphics[width=1\linewidth]{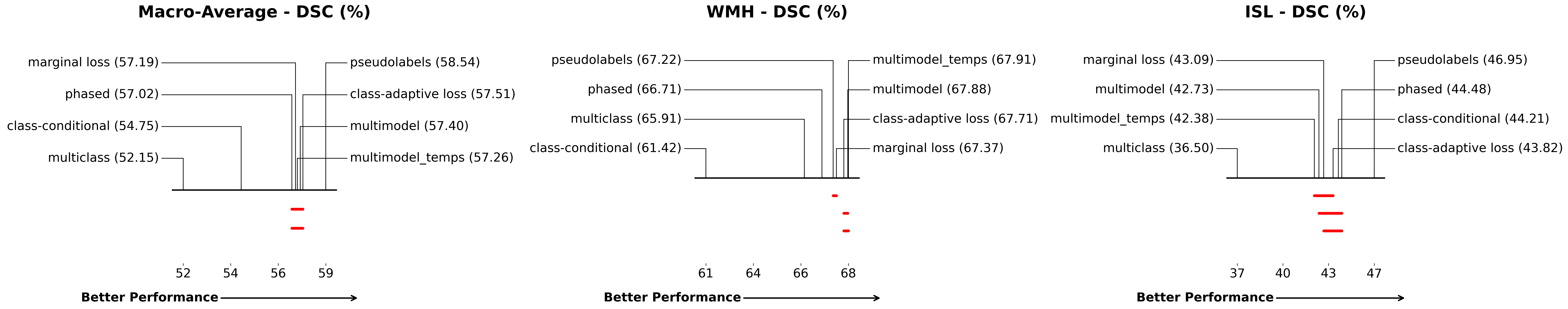}
    \caption{Critical difference (CD) diagrams showing the statistical significance of performance differences between models for Macro-Average, WMH, and ISL classes. The x-axis plots the mean performance on the dice similarity coefficient (DSC) metric. Horizontal red lines group models that perform comparably (no statistically significant difference ($p\geq0.05$)), while unlinked models exhibit significant performance differences.}
    \label{fig:main_cd_dsc}
\end{figure}

\begin{figure}[htbp]
    \centering
    \includegraphics[width=1\linewidth]{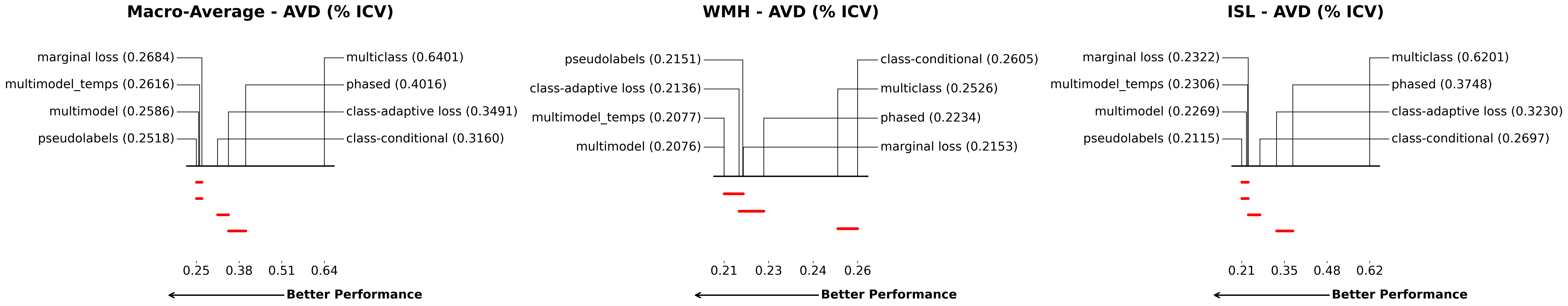}
    \caption{Critical difference (CD) diagrams showing the statistical significance of performance differences between models for Macro-Average, WMH, and ISL classes. The x-axis plots the mean performance on the absolute volume difference (AVD) metric. Horizontal red lines group models that perform comparably (no statistically significant difference ($p\geq0.05$)), while unlinked models exhibit significant performance differences.}
    \label{fig:main_cd_avd}
\end{figure}

\begin{figure}[htbp]
    \centering
    \includegraphics[width=1\linewidth]{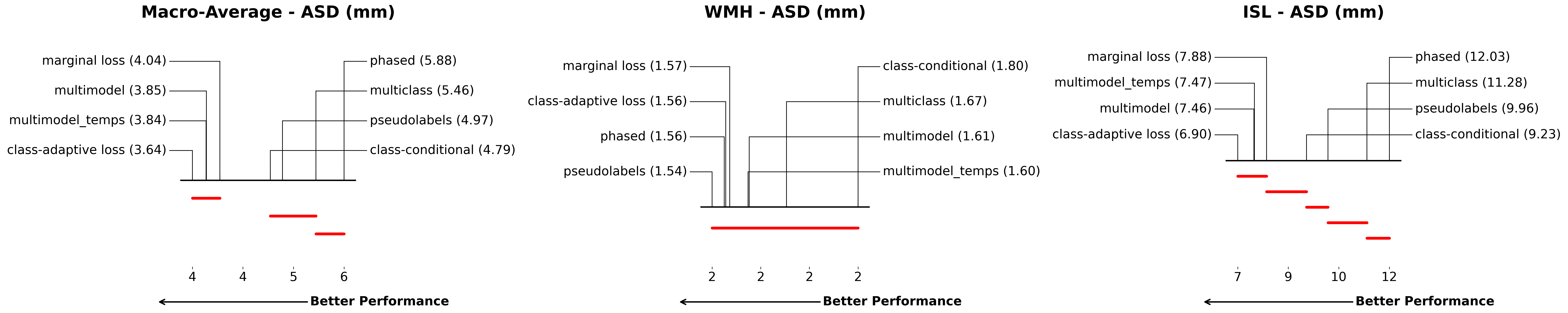}
    \caption{Critical difference (CD) diagrams showing the statistical significance of performance differences between models for Macro-Average, WMH, and ISL classes. The x-axis plots the mean performance on the average surface distance (ASD) metric. Horizontal red lines group models that perform comparably (no statistically significant difference ($p\geq0.05$)), while unlinked models exhibit significant performance differences.}
    \label{fig:main_cd_asd}
\end{figure}

\begin{figure}[htbp]
    \centering
    \includegraphics[width=1\linewidth]{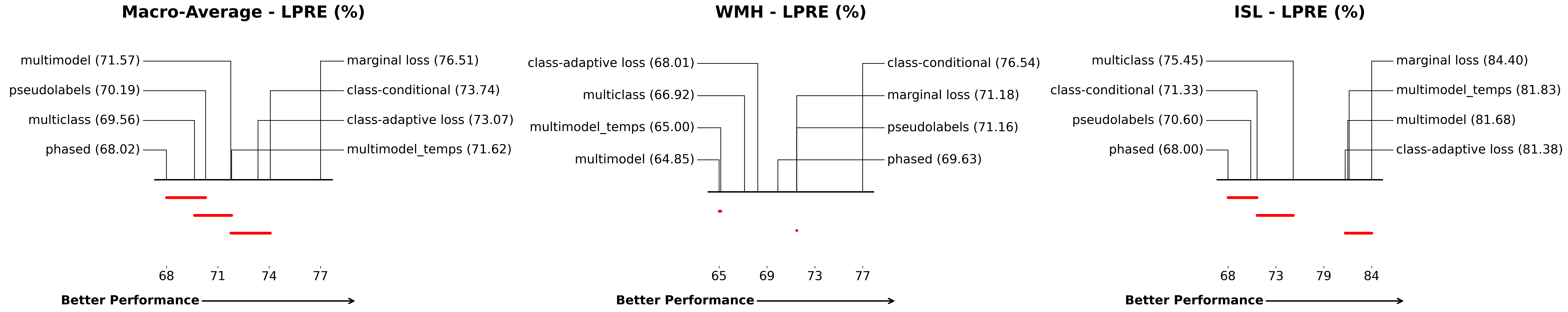}
    \caption{Critical difference (CD) diagrams showing the statistical significance of performance differences between models for Macro-Average, WMH, and ISL classes. The x-axis plots the mean performance on the lesion precision (LPRE) metric. Horizontal red lines group models that perform comparably (no statistically significant difference ($p\geq0.05$)), while unlinked models exhibit significant performance differences.}
    \label{fig:main_cd_lpre}
\end{figure}

\begin{figure}[htbp]
    \centering
    \includegraphics[width=1\linewidth]{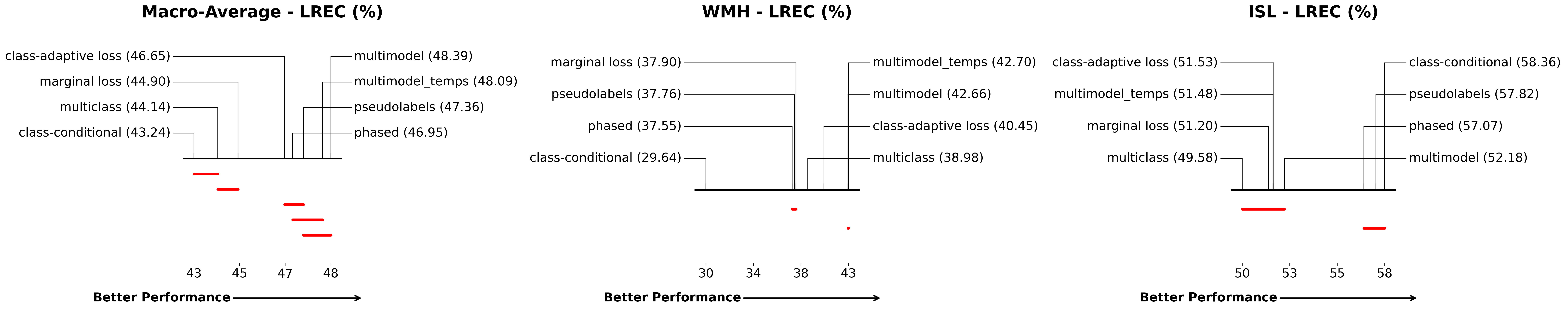}
    \caption{Critical difference (CD) diagrams showing the statistical significance of performance differences between models for Macro-Average, WMH, and ISL classes. The x-axis plots the mean performance on the lesion recall (LREC) metric. Horizontal red lines group models that perform comparably (no statistically significant difference ($p\geq0.05$)), while unlinked models exhibit significant performance differences.}
    \label{fig:main_cd_lrec}
\end{figure}

\begin{figure}[htbp]
    \centering
    \includegraphics[width=1\linewidth]{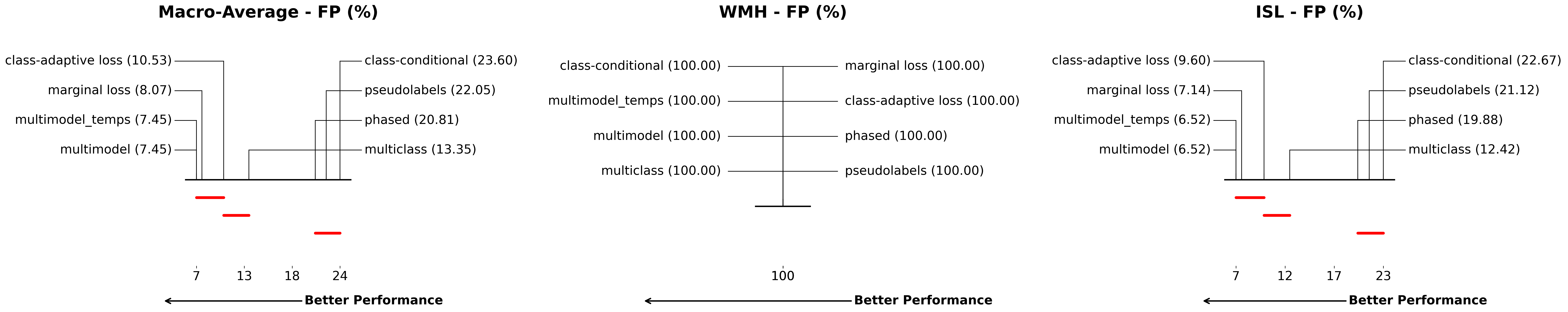}
    \caption{Critical difference (CD) diagrams showing the statistical significance of performance differences between models for Macro-Average, WMH, and ISL classes. The x-axis plots the mean performance on the scan-level false positive (FP) metric. Horizontal red lines group models that perform comparably (no statistically significant difference ($p\geq0.05$)), while unlinked models exhibit significant performance differences.}
    \label{fig:main_cd_fp}
\end{figure}

\clearpage
\subsection{Fully vs. partially labelled binary models}

This table presents the statistical comparisons between binary models trained exclusively on the \acrfull{fls} versus those trained on the \acrfull{pls}, corresponding to the performance metrics detailed in \textbf{Table \ref{tab:extra_data}} of the main text.

\includepdf[pages=-, scale=0.9, offset=0 60, pagecommand={\thispagestyle{plain}}]{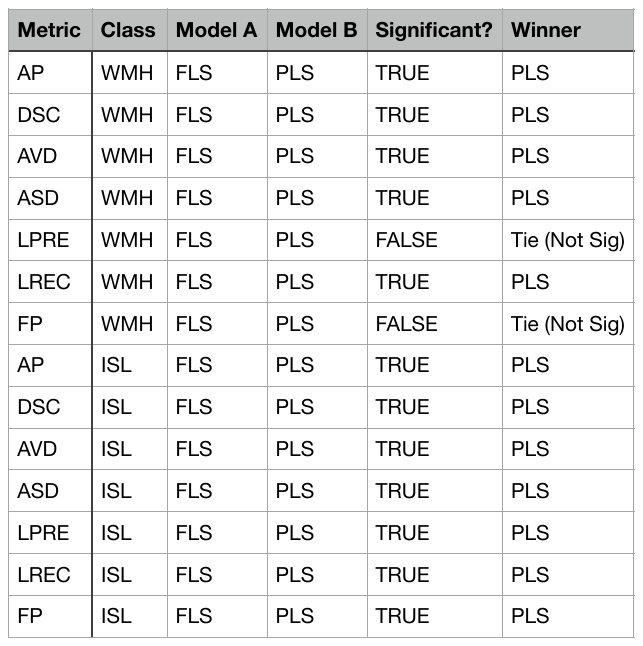}

\subsection{Fully vs. partially labelled binary models: per-dataset change}

This table presents the statistical significance of the per-dataset percentage changes in performance metrics when comparing FLS-trained and PLS-trained binary models, corresponding to the results reported in \textbf{Table \ref{tab:extra_data_2}} of the main text.

\includepdf[pages=-, scale=0.9, offset=0 60, pagecommand={\thispagestyle{plain}}]{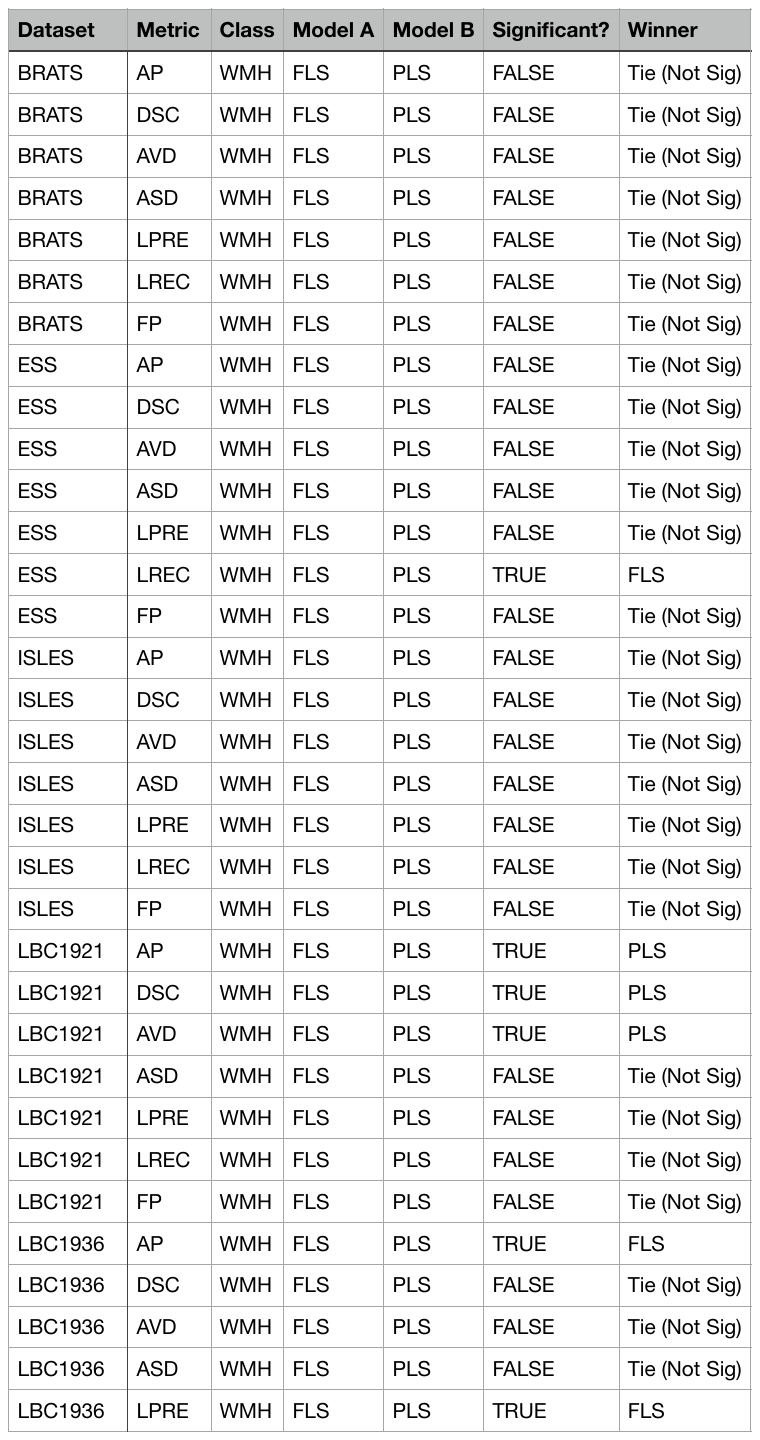}

\subsection{Negative controls on BRATS}

These figures present the statistical comparisons of the false-positive prediction rates within the tumour ground truth masks on the BRATS test set, corresponding to the data presented in \textbf{Table \ref{tab:results_brats}} of the main text. Note that these pairwise comparisons evaluate the difference in the false-positive \textit{rate} (calculated out of the 17 total test cases), rather than the absolute count of false-positives. Critical difference plots are used to visualise multiple model comparisons effectively.

\begin{figure}[htbp]
    \centering
    \includegraphics[width=1\linewidth]{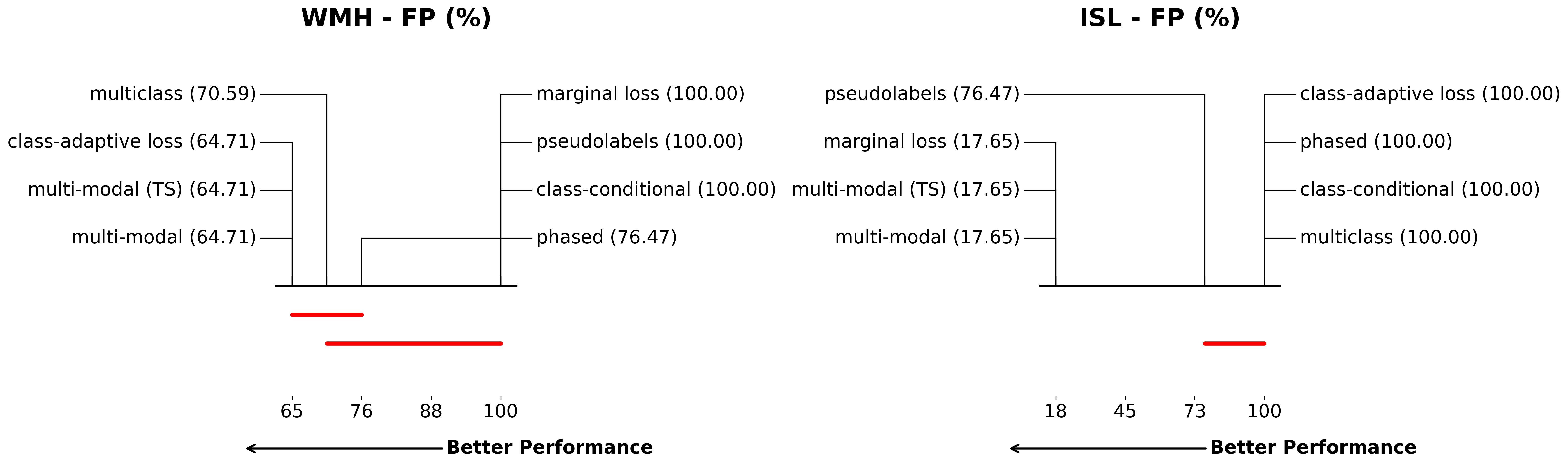}
    \caption{Critical difference (CD) diagrams showing the statistical significance of performance differences between models for WMH and ISL classes on the BRATS dataset. The x-axis plots the mean performance on the scan-level false positive (FP) metric. Horizontal red lines group models that perform comparably (no statistically significant difference ($p\geq0.05$)), while unlinked models exhibit significant performance differences.}
    \label{fig:brats_cd_fp}
\end{figure}

\bibliographystyle{elsarticle-num} 
\bibliography{references}

\end{document}